\documentclass{article}
\usepackage[mathlines]{lineno}

\PassOptionsToPackage{numbers, compress}{natbib}
\usepackage[main, final]{neurips_2026}

\usepackage[utf8]{inputenc} %
\usepackage[T1]{fontenc}    %
\usepackage[colorlinks = True, linkcolor = violet, citecolor = green!60!black]{hyperref}
\usepackage{url}            %
\usepackage{booktabs}       %
\usepackage{multirow}       %
\usepackage{amsfonts}       %
\usepackage{ltablex}        %

\usepackage{array}
\usepackage{tabularx}
\usepackage{threeparttable} %
\usepackage{listings}       %
\usepackage{longtable}
\usepackage{subcaption}     %
\usepackage{verbatim}      %
\usepackage{amsmath}        %
\usepackage{nicefrac}       %
\usepackage{microtype}      %
\usepackage[table]{xcolor}
\usepackage{cleveref}       %
\usepackage{graphicx}
\usepackage{pifont}
\usepackage{colortbl}
\usepackage{enumitem}
\usepackage{algorithm}
\usepackage{algpseudocode}
\usepackage{wrapfig}

\definecolor{tabhead}{RGB}{232,239,247}
\definecolor{tabwarm}{RGB}{252,239,224} %

\newcolumntype{L}[1]{>{\raggedright\arraybackslash}p{#1}}
\newcolumntype{C}[1]{>{\centering\arraybackslash}p{#1}}

\newcommand{\R}{\mathbb{R}}

\newcommand{\x}{\boldsymbol{x}}

\newcommand{\cmark}{\textcolor{green!60!black}{\ding{51}}}
\newcommand{\xmark}{\textcolor{red!70!black}{\ding{55}}}

\definecolor{ficboMix}{RGB}{221,132,82}
\definecolor{ficboAdd}{RGB}{239,54,73}
\definecolor{gpFb}{RGB}{74,112,180}
\definecolor{gpBase}{RGB}{34,150,243}
\definecolor{randomCol}{RGB}{150,123,100}
\definecolor{feedbackCol}{RGB}{93,105,107}
\definecolor{piboUCB}{RGB}{84,164,98}
\definecolor{piboEI}{RGB}{35,220,130}

\newcommand{\legendline}[1]{%
  \raisebox{0.45ex}{\textcolor{#1}{\rule{1em}{1.4pt}}}%
}

\title{In-Context Black-Box Optimization with Unreliable Feedback}

\author{%
  Nicolas S. Blumer \\
  ELLIS Institute Finland\\
  Aalto University\\
  \And
  Julien Martinelli \\
  ELLIS Institute Finland\\
  Aalto University\\
  \And
  Samuel Kaski \\
  ELLIS Institute Finland\\
  Aalto University\\
  University of Manchester\\
  \vspace{-.8cm}
  \And
  \texttt{\{firstname.lastname\}@aalto.fi}
}

\begin{document}
\maketitle

\begin{abstract}
Black-box optimization in science and engineering often comes with side information: experts, simulators, pretrained predictors, or heuristics can suggest promising candidates.
Such feedback can accelerate search, but can also be biased, input-dependent, or misleading. 
Feedback-aware Bayesian optimization methods typically handle tasks independently, limiting adaptation across feedback sources.
In-context optimizers enable cross-task adaptation, but usually rely only on optimization history at test time.
We study \emph{feedback-informed in-context black-box optimization} (FICBO), where a pretrained optimizer conditions on both the observed history and cheap auxiliary feedback for the current candidate set. 
We introduce structured priors over how auxiliary feedback relates to the objective, and use them to pretrain a feedback-aware transformer across regimes ranging from highly informative to misleading.
At test time, the model estimates source reliability in context by comparing observed objective values with auxiliary signals, improving query selection.
On synthetic and real-world tasks, FICBO effectively exploits informative feedback while remaining robust to weak or misleading sources. 
Empirical investigations further illustrate how the model perceives test-time sources, offering insights into its interpretability and decision-making process.
\end{abstract}

\section{Introduction}

Optimizing expensive black-box experiments is central to scientific and engineering applications, including biology~\citep{wu2025generative}, chemistry~\citep{sundin2022human}, and materials science~\citep{liang2021benchmarking}.
In such domains, experts can guide search through manual assessment, hand-crafted heuristics, or surrogate models, but their feedback can be systematically unreliable~\citep{mikkola,fan2024multi}.
Expertise may be reliable only in familiar regions, while proxies can omit relevant mechanisms, for example when a chemical reaction model ignores a competing side reaction that becomes important in part of the design space, or extrapolate poorly beyond their design conditions.
Feedback from a related but distinct task may likewise transfer imperfectly, as when CIFAR-10 accuracy ranks architectures differently from ImageNet16-120.
Such feedback is rarely uniformly good or bad: it may be accurate for some tasks or regions and misleading in others, with its reliability unknown in advance and potentially changing as search progresses.

\begin{figure}[h!]
\centering
    \includegraphics[width=1\linewidth, , trim=0 70 0 0pt, clip]{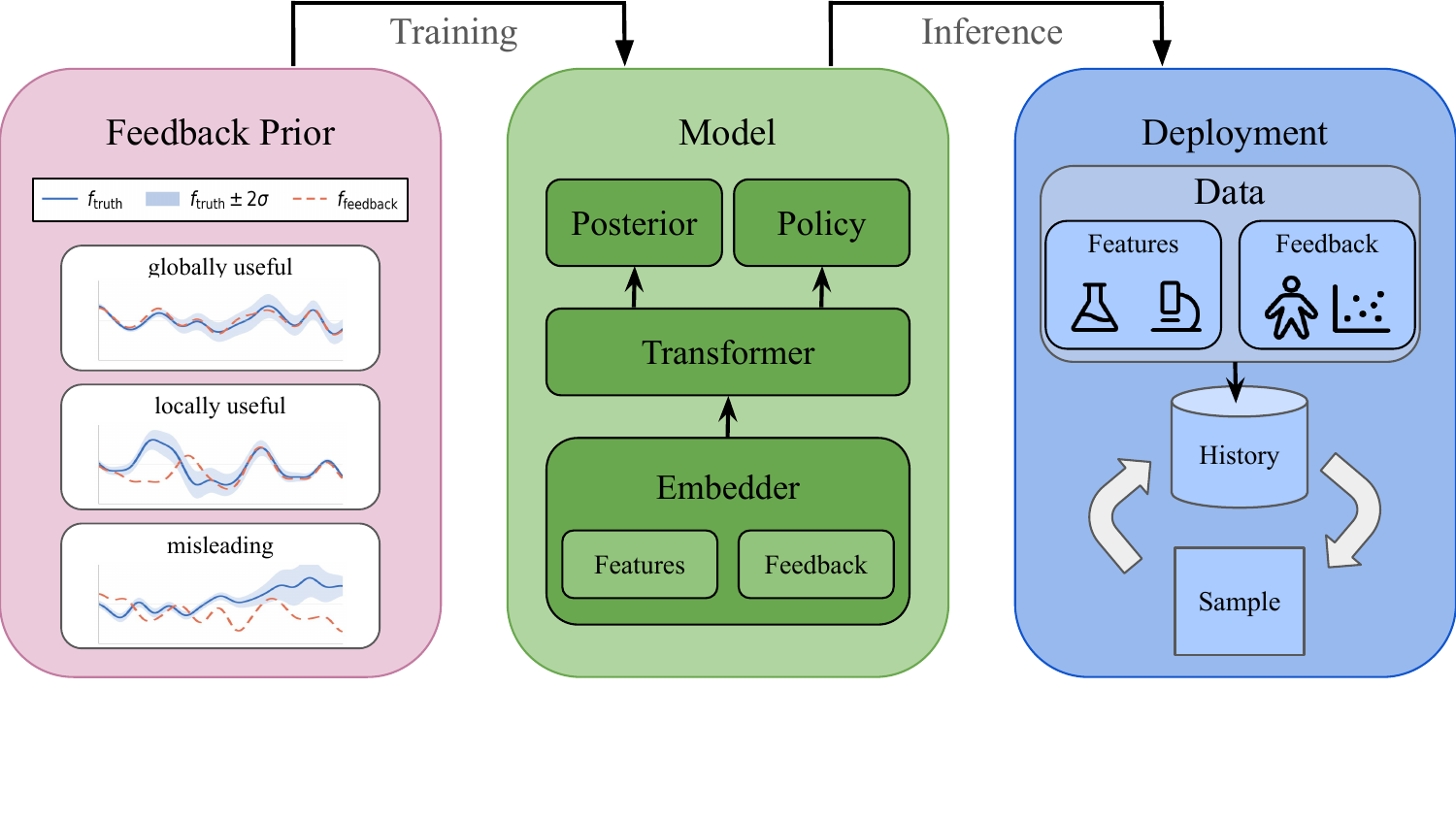}
    \label{fig:gist}

    \centering
    \vspace{-.7cm}
    \small
    \setlength{\tabcolsep}{0pt}
    \renewcommand{\arraystretch}{1.1}
    \setlength{\aboverulesep}{0pt}
    \setlength{\belowrulesep}{0pt}
    \setlength{\extrarowheight}{0.2ex}

\begin{tabularx}{\textwidth}{
    L{4.85cm}
    C{1.65cm}
    C{1.85cm}
    C{2.70cm}
    C{2.90cm}}
\toprule
\rowcolor{tabhead}
\textbf{Methods}
& \shortstack{\textbf{Surrogate}\\\textbf{amortized}}
& \shortstack{\textbf{Query policy}\\\textbf{amortized}}
& \shortstack{\textbf{Auxiliary feedback}\\\textbf{at test time}}
& \shortstack{\textbf{Feedback-aware}\\\textbf{pretraining}}\\
\midrule

Vanilla BO~\cite{garnett2023bayesian}
& \xmark & \xmark & \xmark & N. A. \\

$\pi$BO~\citep{hvarfner2022pibo},
CoExBO~\cite{adachi}
& \xmark & \xmark & \cmark & N. A. \\

PFNs4BO~\cite{muller2023pfns4bo}, GIT-BO~\citep{yu2026gitbo}
& \cmark & \xmark & \xmark & \xmark \\

BoFormer~\citep{hung2025boformer}, $\alpha$-PFN~\cite{rakotoarison2026}
& \xmark & \cmark & \xmark & \xmark \\

NAP~\cite{maraval2023end}, PABBO~\citep{zhang2025pabbo}, TAMO~\citep{zhang2026taskagnostic}
     & \cmark & \cmark & \xmark & \xmark \\

\midrule
\rowcolor{tabwarm}
\textbf{FICBO (This work)}
& \cmark & \cmark & \cmark & \cmark \\
\bottomrule
\end{tabularx}
    \vspace{-.5cm}
    \caption{\textbf{Feedback-informed in-context black-box optimization (FICBO) and related work.} During training, tasks are sampled from a feedback prior with auxiliary sources of varying reliability. A transformer learns both posterior prediction and query selection from features and feedback. At deployment, it conditions on the optimization history and the available feedback to guide search.
    }
    \label{tab:comparison}
    \vspace{-0.2cm}
\end{figure}

A natural framework for optimizing expensive objectives under such imperfect guidance is Bayesian optimization (BO), which models the unknown objective with a probabilistic surrogate, often a Gaussian process~\citep{RasmussenW06}, and selects the next design through an acquisition function that balances exploration and exploitation~\cite{garnett2023bayesian}. 
This probabilistic and sequential structure makes BO particularly well suited to settings with auxiliary feedback, since such information can shape inference and guide design selection. 
Existing feedback-aware BO methods exploit expert knowledge through informative priors or user beliefs over the objective~\citep{sundin2022human, hvarfner2022pibo, mikkola, adachi}. 
Some reweight the acquisition function using a static user prior whose influence is controlled by a user-specified parameter~\citep{hvarfner2022pibo,souza2020prior}.
Yet the appropriate influence depends on source reliability, which is unknown a priori and may vary across the search space. %

Amortized optimization offers a natural way to learn this feedback-adaptive behavior across tasks. By pretraining a neural optimizer on a distribution of synthetic tasks with diverse feedback, one can move part of the optimization burden offline and encode assumptions about the expected problem class directly into the task prior. The resulting model can then adapt \emph{in context} to a new problem from its optimization history alone, without task-specific retraining. Recent work has shown this paradigm to be highly effective in settings such as Bayesian experimental design and BO~\citep{foster,maraval2023end,huang2025aline,zhang2026taskagnostic,nguyen2026flow}.

However, existing amortized optimizers typically take only the standard optimization history as input. They are not designed for settings where an auxiliary source is available at test time and may help on some tasks, be irrelevant on others, or induce negative transfer if overused~\citep{mikkola}. We therefore aim to learn through pretraining how such feedback should modulate posterior inference and query selection.

\textbf{Contributions.} We introduce feedback-informed in-context black-box optimization (FICBO), an
in-context black-box optimization framework for settings with unreliable auxiliary feedback, using expert sources as a primary example (Figure~\ref{tab:comparison}).
Concretely, we contribute: (1) conceptually, a formulation in which an optimizer receives auxiliary feedback at test time and must infer its reliability on-the-fly; (2) methodologically, a feedback-aware amortized black-box optimization framework pretrained under a feedback prior over source reliability%
; and (3) empirically, an evaluation protocol spanning synthetic and real tasks, with ablations over feedback strength and bias type, comparisons against classical and amortized baselines, and empirical investigations that illustrate how the model perceives test-time sources.
\vspace{-10pt}
\section{Background: \emph{in-context} black-box optimization}\label{sec:background}
    \vspace{-.1cm}
We consider the black-box optimization problem
\begin{equation}
    \x^* \in \arg\max_{\x \in \mathcal{X}} f(\x),
\end{equation}
where $\mathcal{X} \subseteq \mathbb{R}^d$ is the design space and $f:\mathcal{X}\to\mathbb{R}$ is unknown and accessible only through expensive evaluations. At round $t$, querying $\x_t$ yields a noisy observation $y_t = f(\x_t) + \varepsilon_t$ with $\varepsilon_t \sim \mathcal{N}(0,\sigma^2)$, and the history after $t$ rounds is $H_t = \{(\x_i,y_i)\}_{i=0}^{t-1}$. 

Classical methods solve each optimization problem independently. In contrast, \emph{in-context} black-box optimization leverages commonalities across related problems by shifting computation offline. This motivates defining a distribution over related tasks, $\tau \sim p(\tau)$, where each task has its own objective $f_\tau$. By pretraining across such tasks, a neural optimizer can adapt at test time to a new problem directly from its observed history, without requiring task-specific retraining.

We focus on a pool-based setting, which is the standard setting in the amortized BO/Bayesian Experimental Design (BED)/Active Learning (AL) literature (e.g., \citep{foster, huang2025aline, zhang2026taskagnostic, zhang2025pabbo, swersky2020amortized}). 
Each task $\tau$ is associated with a finite candidate set $Q^\tau = \{\x_j^\tau\}_{j=1}^{N_\tau}$.
We discuss how our approach extends to the continuous setting in Appendix \ref{app:Continuous}.
At round $t$, the optimizer selects $\x_t^\tau \in Q_t^\tau \subseteq Q^\tau$, observes $y_t^\tau = f_\tau(\x_t^\tau) + \varepsilon_t^\tau$ with $\varepsilon_t^\tau \sim \mathcal{N}(0,\sigma_\tau^2)$, and updates the corresponding history $H_t^\tau = H_0^\tau \cup \{(\x_i^\tau,y_i^\tau)\}_{i=0}^{t-1}$.
The optimizer comprises a predictive component
\begin{equation}
    q_{\phi}(y \mid \x, H_t^\tau),
\end{equation}
which estimates the value of a candidate $\x \in Q_t^\tau$, and a policy that directly selects the next query,
\begin{equation}
    \x_{t}^\tau \sim \pi_{\psi}(\cdot \mid H_t^\tau, Q_t^\tau),
\end{equation}
with $Q_{0}^\tau = Q^\tau$ and $Q_{t+1}^\tau = Q_{t}^\tau \setminus \{\x_{t}^\tau\}$.
The policy is the primary object, as it determines which candidate is evaluated next. The predictive model serves as an auxiliary training signal, providing predictions with uncertainty for interpretability and potentially acting as a faster GP surrogate~\citep{maraval2023end,huang2025aline,zhang2026taskagnostic}.

Training proceeds offline over trajectories sampled from $p(\tau)$, optimizing
\vspace{-.2cm}
\begin{equation}
    \max_{\psi}\;
    \mathbb{E}_{\tau \sim p(\tau)}
    \Big[
        \sum_{t=0}^{T_\tau-1} r_t^\tau \gamma^t
    \Big],
\end{equation}
where $y^*_t = \max\{y : (\x, y) \in H_t^\tau \}$  denotes the best value observed strictly before round $t$, and $r_t^\tau = \max(y_t^\tau - y^*_t,\; 0)$ is the improvement of the $t$-th query over it. $\gamma \in (0,1]$ is the discount factor.
Unlike classical BO, which maximizes an acquisition function at each step, the in-context optimizer is trained end-to-end to maximize cumulative reward over the full trajectory, replacing repeated model fitting and acquisition optimization with a learned procedure~\citep{maraval2023end,huang2025aline,zhang2026taskagnostic}.

\vspace{-.2cm}
\section{Method}\label{sec:method}
\vspace{-.2cm}

We extend the \emph{in-context} black-box optimization setup (Section~\ref{sec:background}) to a feedback-aware setting, where the optimizer also receives candidate-wise auxiliary feedback over the pool,
\begin{equation}
    U^\tau = \{u^\tau(\x) : \x \in Q^\tau\}.
\end{equation}
We assume feedback is available for every point in the query set \footnote{If the source is too expensive to query exhaustively, a cheap surrogate fit to a small number of labeled points can be used instead, as in preference elicitation~\cite{adachi,zhang2025pabbo,ungredda2023elicit}, at the cost of additional bias. The model-based source (an amortized GP fit to source-private data) described in Sec.~\ref{sec:priorfeedback} is designed to let the model learn to discount exactly this kind of bias.}.

The signal $U^\tau$ conveys source-specific information about candidate quality rather than ordinary input features. Its usefulness may vary across tasks and regions of the design space: $u^\tau(\x)$ may be noisy, systematically biased, only locally informative, or misleading. The central challenge is to infer from the observed optimization history how strongly this signal should influence search on the current task.

Rather than a single feedback model, we construct two complementary priors, each representing a different way feedback can go wrong: access to a different, incomplete set of features that only partially explains the objective, or a wrong interpretation of how latent factors of the task should be weighted.
On top of each, we apply additional distortions that further degrade the signal.
Together, they are designed to cover a broad range of feedback, from uninformative to highly reliable.
These priors were designed to be general and serve as a starting point, and practitioners can extend them to reflect their own knowledge and uncertainty about the feedback they expect to encounter.

\vspace{-.1cm}
\subsection{Priors over feedback mechanisms}
\label{sec:priorfeedback}

\begin{figure}[t]
    \centering
    \setlength{\tabcolsep}{2pt}
    \begin{tabular}{ccccc}
        \multicolumn{5}{c}{\includegraphics[width=0.4\linewidth]{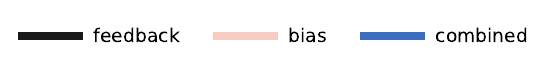}} \\[-12pt]
        \includegraphics[width=0.195\linewidth]{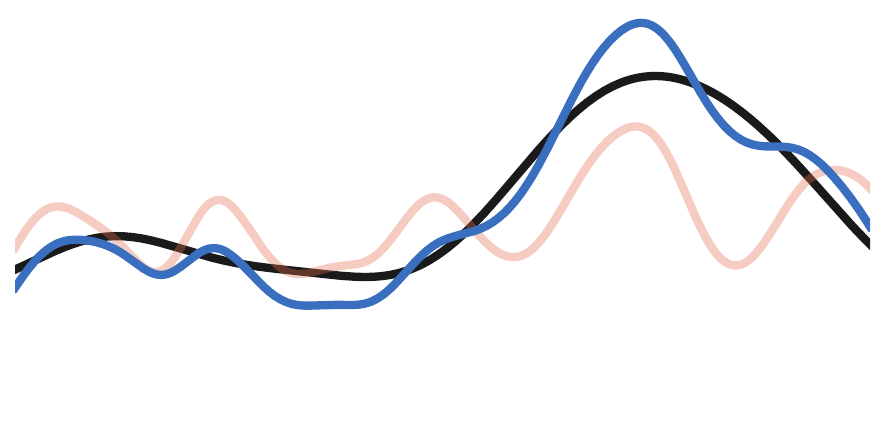} &
        \includegraphics[width=0.195\linewidth]{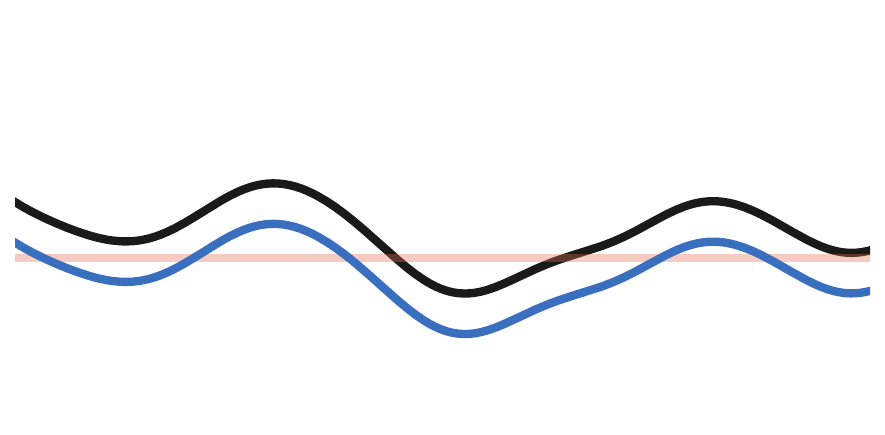} &
        \includegraphics[width=0.195\linewidth]{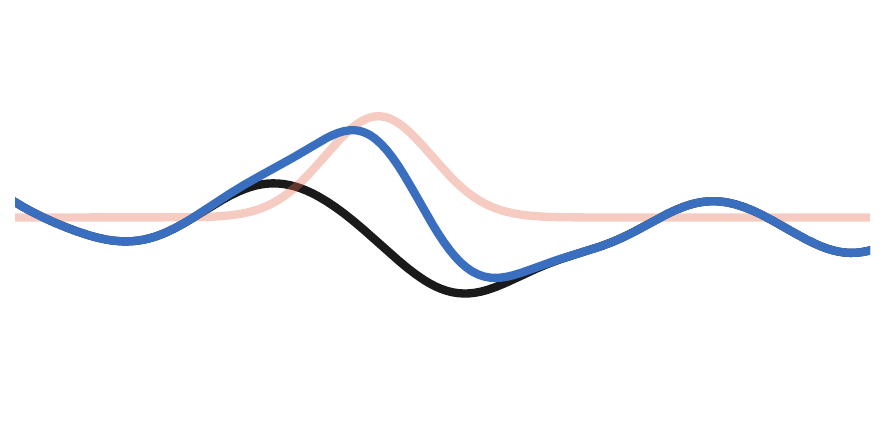} &
        \includegraphics[width=0.195\linewidth]{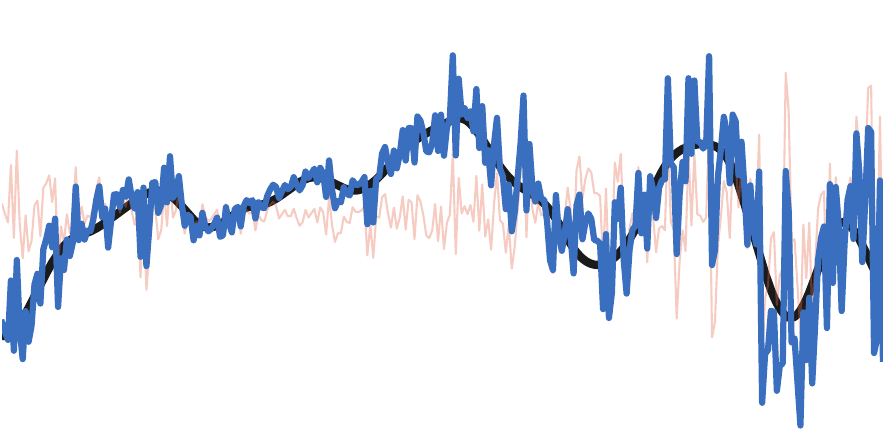} &
        \includegraphics[width=0.155\linewidth]{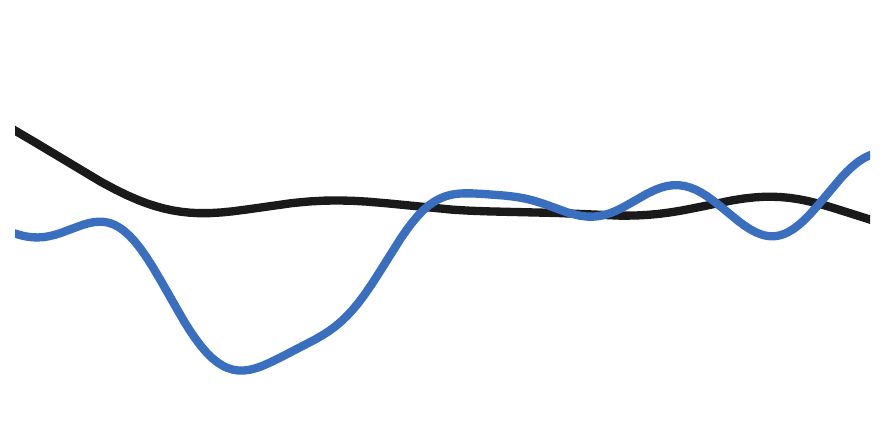}
        \\[-2pt]
        {\small Additive GP Bias} &
        {\small Constant Shift} &
        {\small Local Distortion} &
        {\small Noise} &
        {\small Catastrophic Failure}
    \end{tabular}
    \caption{Illustration of five signal degradation bias types applied to a feedback GP.}
    \label{fig:bias_examples}
\end{figure}

Following the amortized pretraining paradigm~\citep{huang2025aline, zhang2025pabbo, hollmann2023tabpfn, swersky2020amortized},
we pretrain on synthetic optimization episodes sampled from a joint prior over objectives and auxiliary feedback sources. Each task is
\begin{equation}
    \tau
    =
    \bigl(
        Q_{\mathrm{full}}^\tau,\,
        P_{\mathrm{model}}^\tau,\,
        H_0^\tau,\,
        f_\tau,\,
        h_\tau,\,
        T_\tau
    \bigr),
    \label{eq:task_tuple}
\end{equation}
where $Q_{\mathrm{full}}^\tau=\{\x^{\mathrm{full}}_j\}_{j=1}^{N_\tau}$ is a latent candidate pool, $P_{\mathrm{model}}^\tau$ is the projection whose output determines visibility to the optimizer, $H_0^\tau$ is the initial history, $f_\tau$ is the objective, $h_\tau$ is the feedback-generating map, and $T_\tau$ is the horizon. 
The optimizer observes only $Q^\tau$, its feedback values, and the outcomes it collects. 
The features removed by the projection $P_{\mathrm{model}}^\tau$, along with the various latent functions introduced later, are all hidden from the optimizer. The optimizer sees only
\vspace{-.1cm}
\begin{equation}
    \x_j^\tau
    =
    P_{\mathrm{model}}^\tau \x_j^{\mathrm{full}},
    \qquad
    Q^\tau
    =
    \{\x_j^\tau:\x_j^{\mathrm{full}}\in Q_{\mathrm{full}}^\tau\},
    \label{eq:model-visible-pool}
\end{equation}
while objective and feedback values are generated as $y_j^\tau=f_\tau(\x_j^{\mathrm{full}})+\varepsilon_j^\tau$ and $u_j^\tau=h_\tau(\x_j^{\mathrm{full}})$. 
The source parameters and bias parameters are unobserved. We use expert and source interchangeably, as the feedback source is directly or indirectly an expert.

Every task follows the same three stages: (i) a prior generates the
objective $f_\tau$ together with an \emph{ideal} source signal $g_\tau$ which is
informative about $f_\tau$, but usually not identical to it; (ii) a degradation $\mathcal{B}_\tau$ gives $\tilde g_\tau = \mathcal{B}_\tau[g_\tau]$ (Fig.~\ref{fig:bias_examples}); (iii) optionally, $\tilde g_\tau$ is replaced by a surrogate fitted to source-private data of the function. 
The observed feedback is
$u^\tau(\x_j) = h_\tau(\x_j^{\mathrm{full}})$:
\begin{equation}
    \x^{\mathrm{full}}
    \;\longrightarrow\;
    g_\tau
    \;\xrightarrow{\ \mathcal{B}_\tau\ }\;
    \tilde g_\tau
    \;\xrightarrow{\ M_\tau\ (\text{opt.})\ }\;
    u,
    \qquad
    \x = P_{\mathrm{model}}\,\x^{\mathrm{full}}.
    \label{eq:pipeline}
\end{equation}
The two priors below differ in stage (i), i.e., in \emph{why} the source is imperfect: it sees different inputs than the optimizer (additive prior), or the source misinterprets the relationship of latent factors (mixture prior). Stages (ii)--(iii) can be applied to both, but we only applied (iii) to the additive prior.

\textbf{Additive prior.}
For the additive prior, we randomize the visibility pattern by sampling the number of source-visible input dimensions $d_{\mathrm{src}}^\tau$ and an overlap $o^\tau$ between the optimizer's and source's features, giving $d_{\mathrm{full}}^\tau=d+d_{\mathrm{src}}^\tau-o^\tau$ where $d$ is the optimizer input dimension:
\begin{equation}
    \x^{\mathrm{full}}
    =
    \bigl(
        \x_{\mathrm{model}},
        \x_{\mathrm{shared}},
        \x_{\mathrm{src}}
    \bigr),
    P_{\mathrm{model}}\x^{\mathrm{full}}
    =
    (\x_{\mathrm{model}},\x_{\mathrm{shared}}),
    P_{\mathrm{src}}\x^{\mathrm{full}}
    =
    (\x_{\mathrm{shared}},\x_{\mathrm{src}}).
    \label{eq:visibility-split}
\end{equation}
Here $\x_{\mathrm{model}} \in \R^{d - o^\tau}$ is optimizer-only, $\x_{\mathrm{src}} \in \R^{d^\tau_{\mathrm{src}}-o^\tau}$ is source-only,
and $\x_{\mathrm{shared}} \in \R^{o^\tau}$ is shared. 
This generates tasks where feedback may contain information absent from the optimizer's input yet remain incomplete or biased.
$d$ is fixed per trained model (we train one variant per $d \in \{1,2,3\}$), while $d^\tau_{\mathrm{src}}$ and $o^\tau$ are resampled per task: the optimizer sees the first $d$ of $d_{\mathrm{full}}^\tau = d + d^\tau_{\mathrm{src}} - o^\tau$ coordinates, the source the last $d^\tau_{\mathrm{src}}$, so the two views overlap in exactly $o^\tau$ coordinates. 

The objective combines two independent GP components:
one depending on features visible only to the optimizer, and one depending on features visible to the source as described above. 
We sample $f_\tau^{\mathrm{model}}\sim\mathcal{GP}(0,k_\tau^{\mathrm{model}})$ on $\x_{\mathrm{model}}$ and $f_\tau^{\mathrm{src}}\sim\mathcal{GP}(0,k_\tau^{\mathrm{src}})$ on $(\x_{\mathrm{shared}},\x_{\mathrm{src}})$, and set

\begin{equation}
    f_\tau(\x^{\mathrm{full}})
    =
    \frac{
        (1-w_\tau)\,
        f_\tau^{\mathrm{model}}(\x_{\mathrm{model}})
        +
        w_\tau\,
        f_\tau^{\mathrm{src}}(\x_{\mathrm{shared}},\x_{\mathrm{src}})
    }{
        \sqrt{(1-w_\tau)^2+w_\tau^2}
    },
    \qquad
    w_\tau\in[0,1].
    \label{eq:objective-decomp}
\end{equation}
The denominator keeps $\mathrm{Var}[f_\tau]$ between the component variances for every $w_\tau$, so the output scale of $f_\tau$ reveals little about $w_\tau$. 
Relevance must be inferred from the history.
The source's ideal signal is $g_\tau(\x^{\mathrm{full}})=f_\tau^{\mathrm{src}}(\x_{\mathrm{shared}},\x_{\mathrm{src}})$, so $w_\tau$ controls source relevance: large $w_\tau$ makes the source-visible component central to the objective, while small $w_\tau$ makes it accurate but weakly relevant.

The ideal signal is then distorted as $\tilde g_\tau=\mathcal{B}_\tau[g_\tau]$, where $\mathcal{B}_\tau$ may add noise, global shifts, smooth GP-shaped bias, local reliability effects, or rare replacements; examples are shown in Figure~\ref{fig:bias_examples}. 
The distorted signal can be used directly, or, in the model-based branch, passed through a source model (an amortized GP) conditioned on additional source-only observations. 
\begin{equation}
    \mathcal{D}_{\mathrm{src}}^\tau
    =
    \left\{
        \left(
            P_{\mathrm{src}}\x_i^{\mathrm{full}},
            \tilde g_\tau(\x_i^{\mathrm{full}})
        \right)
    \right\}_{i=1}^{n_{\mathrm{src}}},
    \qquad
    M_\tau
    =
    \operatorname{Fit}
    \bigl(\mathcal{D}_{\mathrm{src}}^\tau\bigr).
    \label{eq:source-private-data}
\end{equation}
Here $\operatorname{Fit}$ denotes conditioning a source model on $\mathcal{D}_{\mathrm{src}}^\tau$; in pretraining, this is an amortized GP whose predictive mean gives the feedback.
The feedback map is
\begin{equation}
    h_\tau(\x^{\mathrm{full}})
    =
    \begin{cases}
        \tilde g_\tau(\x^{\mathrm{full}}),
        & \text{direct source}, \\[2pt]
        M_\tau(P_{\mathrm{src}}\x^{\mathrm{full}}),
        & \text{model-based source}.
    \end{cases}
    \label{eq:feedback-pipeline}
\end{equation}
The source model is initialized once from $\mathcal{D}_{\mathrm{src}}^\tau$, separate from the optimizer's data, and held fixed during the BO episode (the test-time XGBoost expert is also fit once, on source-only points, Appendix~\ref{sec:benchmarkfunctions}).

\textbf{Mixture prior.}
The second variant makes reliability a matter of component weighting rather than feature access. Let $K_\tau=K_{\mathrm{real}}+K_{\mathrm{decoy}}^\tau$ and sample independent GP components $z_{\tau,1},\dots,z_{\tau,K_\tau}$ on $\x^{\mathrm{full}}$. Only the first $K_{\mathrm{real}}$ components determine the objective:
\begin{equation}
    f_\tau(\x^{\mathrm{full}})
    =
    \frac{1}{\sqrt{K_{\mathrm{real}}}}
    \sum_{k=1}^{K_{\mathrm{real}}}
    z_{\tau,k}(\x^{\mathrm{full}}).
    \label{eq:reweighted-objective}
\end{equation}
The source forms feedback from a masked and reweighted combination of real and decoy components. With $m_{\tau,1}=1$, $m_{\tau,k}\sim\mathrm{Bernoulli}(p_{\mathrm{src}})$ for $k>1$, and $S_\tau=\{k:m_{\tau,k}=1\}$, we sample
\begin{equation}
    \bar{\boldsymbol{a}}_{\tau,S_\tau}
    \sim
    \mathrm{Dirichlet}
    \bigl(\alpha_{\mathrm{src}}\mathbf{1}_{|S_\tau|}\bigr),
    \qquad
    a_{\tau,k}
    =
    \frac{\bar a_{\tau,k}}{\lVert \bar{\boldsymbol a}_{\tau,S_\tau}\rVert_2}\,\mathbb{I}\{k\in S_\tau\}
    \label{eq:masked-dirichlet-weights}
\end{equation}
Fixing $m_{\tau,1} = 1$ means the source is never entirely uninformative by construction, and the $\ell_2$ normalization gives the source weights unit norm, like the weights $1/\sqrt{K_{\mathrm{real}}}$ of $f_\tau$, so that $g_\tau$ and $f_\tau$ are comparably scaled.
The source signal and feedback are then
\begin{equation}
    g_\tau(\x^{\mathrm{full}})
    =
    \sum_{k=1}^{K_\tau}
    a_{\tau,k}\,
    z_{\tau,k}(\x^{\mathrm{full}}),
    \qquad
    h_\tau
    =
    \mathcal{B}_\tau[g_\tau],
    \label{eq:reweighted-feedback-map}
\end{equation}
where $\mathcal{B}_\tau$ may again be the identity or one of the degradation operators in Figure~\ref{fig:bias_examples}. This prior produces coherent but unreliable sources: the source may ignore real components, overemphasize some of them, or attend to decoys that do not affect the objective.

\textbf{Inference problem induced by the prior.}
In both priors, the learner receives the pool feedback $U^\tau$ and the evaluated history $H_t^\tau$, but not $w_\tau$, hidden coordinates, latent components, source weights, bias parameters, or source-private data.
Here $u_j^\tau$ denotes the
feedback attached to the projected candidate $\x_j^\tau$.
Source reliability must therefore be inferred in context, from how feedback values agree with observed objective evaluations. 
Source-data sampling scheme details are deferred to Appendix~\ref{app:synthgen}.

\subsection{Feedback-aware model}

For a task $\tau$ at round $t < T_\tau$, each point $\x \in Q^\tau$ carries both its features and auxiliary feedback $u^\tau(\x)$. 
Evaluated points additionally carry an observed outcome $y$, giving an augmented history $\widetilde{H}_t^\tau = \widetilde{H}_0^\tau \cup \{(\x_i^\tau, u_i^\tau, y_i^\tau)\}_{i=0}^{t-1}$, where $\widetilde{H}_0^\tau$ is $H_0^\tau$ with feedback attached.
We embed each point into a token via learned embedders $e_x$, $e_u$, $e_y$,
\begin{equation}
    \boldsymbol{E}^{\mathrm{ctx}} = \mathrm{Concat}(e_x(\x), e_u(u)) + e_y(y), \qquad
    \boldsymbol{E}^{\mathrm{qry}} = \mathrm{Concat}(e_x(\x), e_u(u)),
\end{equation}
where context tokens use $(\x, u, y)$ from $\widetilde{H}_t^\tau$, and query tokens use $(\x, u)$ for $\x \in Q_t^\tau$ with $u = u^\tau(\x)$.
A shared transformer backbone maps them to contextualized representations $\widehat{\boldsymbol{E}}^{\mathrm{ctx}}$ and $\widehat{\boldsymbol{E}}^{\mathrm{qry}}$, with query tokens attending to both the context and the full query set. This allows each contextualized candidate representation $\widehat{\boldsymbol{E}}^{\mathrm{qry}}$ to depend on both the optimization history and the pool-wide pattern of auxiliary feedback.

Two heads operate on top of the backbone. The \emph{policy head / acquisition head} maps each contextualized query token to a scalar logit, inducing a categorical distribution over $Q_t^\tau$,
\begin{equation}
    \pi_\psi(\x_j^\tau \mid \widetilde{H}_t^\tau, U^\tau)
    \propto
    \exp\bigl(\mathrm{MLP}_\psi(\widehat{\boldsymbol{E}}_j^{\mathrm{qry}})\bigr).
\end{equation}
The \emph{inference head} maps each query token to an $L$-component Gaussian mixture,
\begin{equation}\label{eq:gmm-head}
    q_\phi(y \mid \x, \widetilde{H}_t^\tau, U^\tau)
    =
    \sum_{\ell=1}^{L} \omega_\ell(\x)\,\mathcal{N}(y;\mu_\ell(\x),\sigma_\ell^2(\x)),\end{equation}
and can be used in place of the policy head by pairing it with a standard acquisition function $\alpha$, giving $\x_{t}^\tau = \arg\max_{\x \in Q_t^\tau} \alpha(\x; q_\phi, \widetilde{H}_t^\tau, U^\tau)$.

\textbf{Training.} Both heads are trained jointly after a warm-up on synthetic episodes sampled from a task prior.
The inference head minimizes a step-wise negative log-likelihood over held-out targets $\mathcal{T}^\tau$, whose labels are never included in $\widetilde H_t^\tau$.
\begin{equation}
\label{eq:Lpred}
    \mathcal{L}_{\mathrm{pred}}(\phi)
    =
    -\mathbb{E}_{\tau}
    \left[
        \frac{1}{T_\tau}\sum_{t=0}^{T_\tau-1}
         \frac{1}{|\mathcal{T}^\tau|}
        \sum_{(\x,y)\in \mathcal{T}^\tau}
        \log q_\phi(y \mid \x, \widetilde{H}_t^\tau, U^\tau)
    \right].
\end{equation}
$\mathcal{L}_{\mathrm{pred}}$ also contains, with equal weight, the NLL of $d+1$ latent GP hyperparameters (Additive prior: the optimizer-visible lengthscales and the output scale of $f_\tau^{\mathrm{model}}$; Mixture prior: the visible lengthscales and the output scale of $z_{\tau,1}$).

We cast policy learning as a finite-horizon MDP where the state at step $t$ is the augmented history and remaining pool, and optimize with REINFORCE using the improvement reward $r_t^\tau = \max(y_t^\tau - y^*_t,\; 0)$.
Further details can be found in~\cref{app:training}.
The policy loss is then
\begin{equation}
    \mathcal{L}_{\mathrm{pol}}(\psi)
    =
    -\mathbb{E}_{\tau,\pi_\psi}
    \left[
        \frac{1}{T_\tau}\sum_{t=0}^{T_\tau-1}
        \log \pi_\psi(\x_{t}^\tau \mid \widetilde{H}_t^\tau, U^\tau)
        \tilde{R}_t^\tau
    \right],
\label{eq:lpol}
\end{equation}
where $\tilde{R}_t^\tau$ is the discounted return $G_t^\tau = \sum_{t'\geq t}\gamma^{t'-t} r_{t'}^\tau$, normalized across the batch at each timestep to reduce gradient variance (full definition in Appendix~\ref{app:training}). 
The code is available on \href{https://github.com/Thahit/FICBO}{GitHub}.

\section{Related work}
\begin{table}[t]
\centering
\small
\setlength{\tabcolsep}{4pt}
\renewcommand{\arraystretch}{1.05}
\begin{tabularx}{\textwidth}{@{}l c X X l@{}}
\toprule
Task & $d$ & Objective & Feedback & Alignment \\
\midrule
Synthetic, marginal expert & 2 & 13 analytic functions & objective marginalized over source-hidden dims & incomplete \\
Synthetic, XGBoost expert  & 2 & 13 analytic functions & shallow regressor on the source dim & biased \\
Branin (multi-fidelity)    & 2 & high-fidelity Branin & low-fidelity variant & tunable ($\rho$)  \\
\midrule
Airfoil Small              & 3 & $C_L/C_D$, large NeuralFoil & $C_L/C_D$, small NeuralFoil & aligned, noisy \\
Airfoil Lift               & 3 & $C_L/C_D$, large NeuralFoil & $C_L$ only & structurally biased \\
Neural Architecture Search      & 3 & ImageNet16-120 accuracy & CIFAR-10 accuracy & optimum differs \\
Reactor                    & 2 & yield of intermediate $[B]$ & conversion of reactant $A$ & misleading late \\
\bottomrule
\end{tabularx}
\caption{\textbf{Evaluation tasks.} $d$ is the input dimensionality seen by the optimizer (discrete for NAS). Synthetic tasks use a full dimensionality of 2 or 3: the optimizer sees the first two dimensions and the source only the last, which gives one shared dimension at full dimensionality 2 and none at 3. The pool size is 300 (123 for NAS) and the horizon is $T=30$ for all tasks.}
\label{tab:tasks}
\end{table}

\textbf{Expert knowledge and feedback in Bayesian optimization and sequential design.}
A substantial literature studies how external expertise can improve BO and Bayesian experimental design, whether through informative priors, interactive feedback, or auxiliary predictive sources~\cite{mikkola2024humans}. Prior-guided approaches such as $\pi$BO incorporate user beliefs about promising regions or the optimum location directly into the optimization procedure \cite{hvarfner2022pibo}, while other methods use interactive preference elicitation or corrective human feedback during the search \cite{adachi}.
Closely related, robust multi-fidelity formulations show that a human expert can act as a cheap but imperfect signal alongside expensive ground-truth evaluations~\citep{mikkola}. In our setting, however, the auxiliary signal is available over the candidate pool; the challenge is to amortize how its reliability is interpreted across tasks.
Expert-informed BO has already found concrete applications in scientific discovery, for instance, in drug-design settings where chemist knowledge or domain feedback is incorporated into the optimization loop \cite{nahal2024human,sundin2022human}. Overall, these works illustrate the value of expert guidance, but they remain task-specific: they do not learn, across a distribution of tasks, how to interpret and calibrate a potentially biased feedback channel.

\textbf{\emph{In-context} black-box optimization.}
A growing line of work amortizes parts of the BO loop using neural in-context models.
Several works, for instance, replace the standard GP surrogate with transformer-based models that infer task-specific predictive distributions in a single forward pass~\cite{hollmann2023tabpfn,muller2023pfns4bo,yao2025fomemo,yu2026gitbo}. Other approaches instead amortize the acquisition mechanism itself, learning to propose queries directly from optimization histories rather than optimizing a handcrafted acquisition function at test time~\cite{hung2025boformer,rakotoarison2026}.
Beyond such partial amortization, recent work targets end-to-end amortized optimization, from early neural acquisition processes~\cite{maraval2023end} to preferential optimization with pairwise feedback~\cite{zhang2025pabbo} and multi-objective optimization~\cite{zhang2026taskagnostic}. Other advances leverage representations from pretrained tabular foundation models to learn acquisition policies, reusing their predictive capabilities rather than training the entire optimizer from scratch~\cite{huang2026polar}. Finally, while many of these methods rely on discrete candidate grids, recent work uses flow matching to generate queries directly in continuous search spaces, paving the way to higher dimensional problems~\cite{nguyen2026flow}.
Yet these methods typically assume a single reliable information channel. Here, by contrast, we consider an additional expert-feedback source whose usefulness cannot be taken for granted.

\section{Experiments}
\label{sec:experiments}

We evaluate whether feedback-aware pretraining lets an in-context optimizer exploit useful auxiliary
signals while remaining robust to misleading ones. Our evaluation spans analytic benchmark functions
with controlled feedback, and real-world tasks where the auxiliary signal is a cheap but imperfect proxy for the expensive objective. 
All methods are run in the same pool-based setting. 
We run each experiment on 100 independent replicas (one fixed seed, a batch of 100 environments), resampling candidate pools, initial observations, and observation noise (0.1 std.) while keeping them shared across methods within each replica.
Each step's simple regret is divided by the gap between the best value in the candidate pool and the best initial observation, which gives a common $[0,1]$ scale; cumulative regret sums this normalized simple regret over the $T$ steps.

\textbf{Baselines.}
We compare the two variants of our method,
 FICBO (\legendline{ficboAdd} Additive Prior; \legendline{ficboMix} Mixture Prior), against three families of in-context baselines.
First, a feedback-agnostic baseline uses the same in-context
optimization architecture, but does not receive the feedback channel
(\legendline{gpBase} GP-prior ICBO  (in-context black-box optimizer)).
Second, a naive feedback-feature baseline
concatenates $u(\x)$ to the input features, testing whether standard amortization
can exploit feedback without source-aware pretraining
(\legendline{gpFb} GP-prior ICBO + feedback feature).
Third,
$\pi$BO-like baselines reweight EI or UCB scores from the GP-prior ICBO surrogate using feedback $u(\x)$
(\legendline{piboUCB} $\pi$BO-UCB; \legendline{piboEI} $\pi$BO-EI~\citep{hvarfner2022pibo}).
Overall, baselines differ only in how the feedback channel is used.
We also include a greedy feedback baseline (\legendline{feedbackCol} Feedback), evaluating candidates in decreasing feedback score, and random search (\legendline{randomCol} Random). Details in Appendix~\ref{app:baseline-details}, including additional baselines from transfer learning.

\captionsetup{skip=0pt}

\begin{figure}[b]
    \centering

    \begin{subfigure}[b]{0.02\textwidth}
        \raisebox{0.1cm}{\rotatebox{90}{\parbox{1.6cm}{\centering\footnotesize Best found ($\uparrow$)}}}
    \end{subfigure}%
    \begin{subfigure}[b]{0.186\textwidth}
        \includegraphics[width=\textwidth]{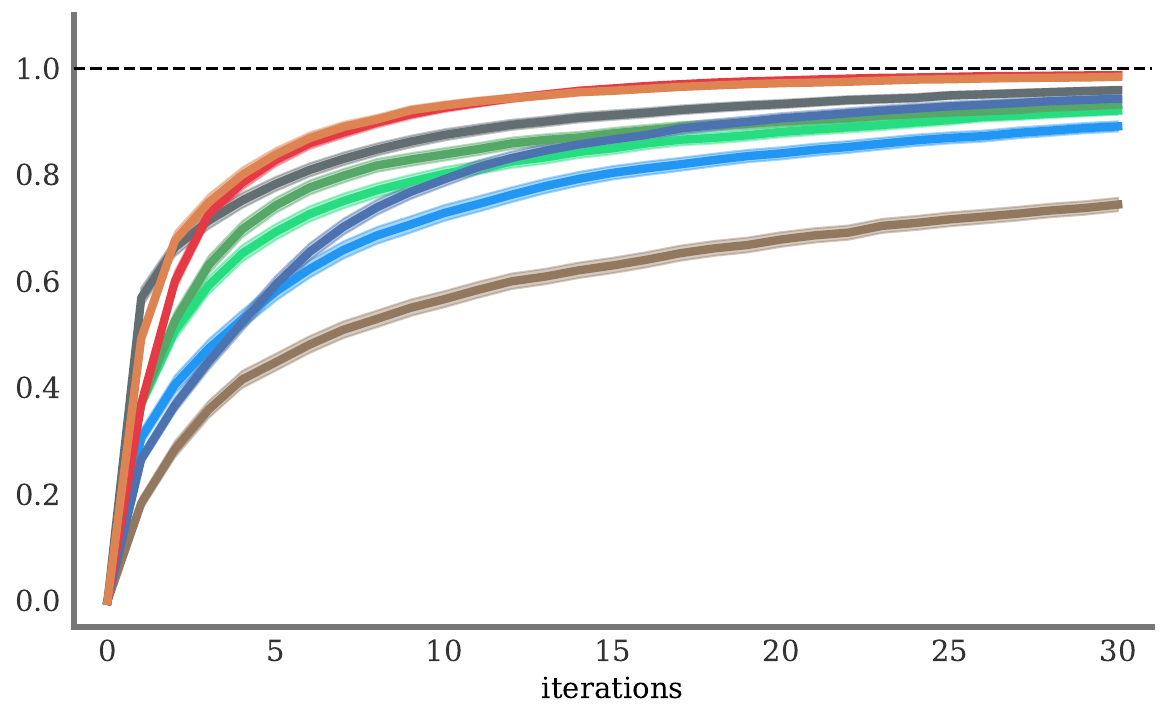}
    \end{subfigure}%
    \hspace{0.002\textwidth}%
    \begin{subfigure}[b]{0.19\textwidth}
        \includegraphics[width=\textwidth]{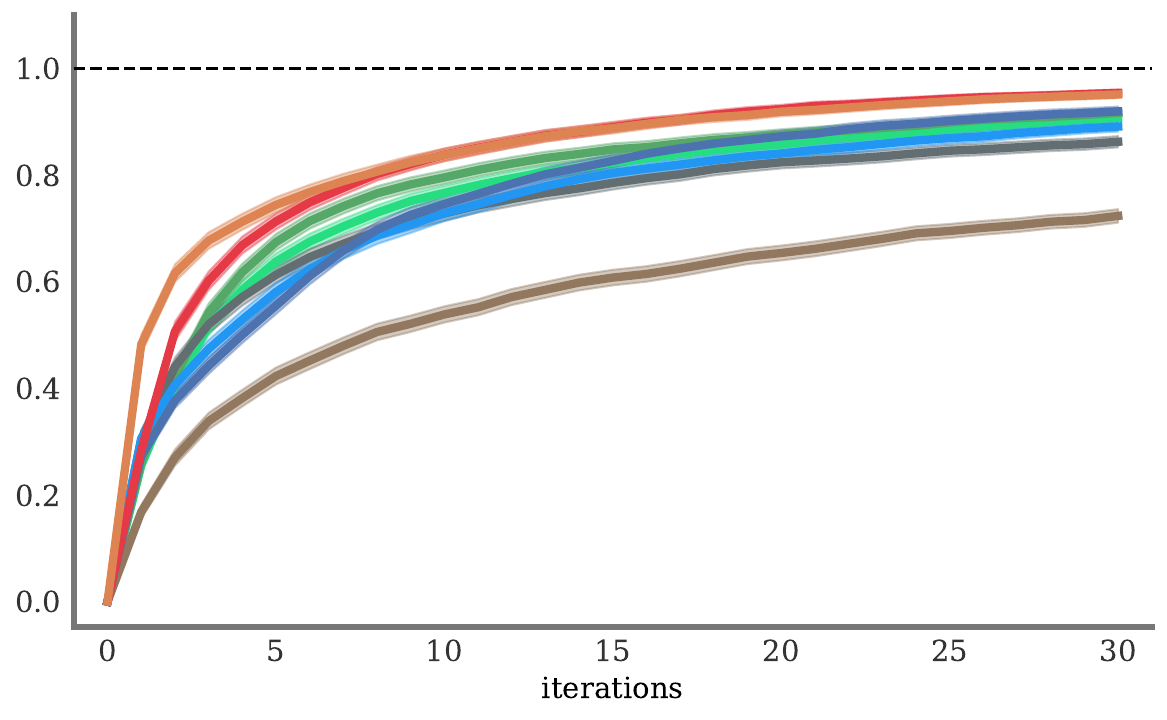}
    \end{subfigure}%
    \hspace{0.002\textwidth}%
    \begin{subfigure}[b]{0.19\textwidth}
        \includegraphics[width=\textwidth]{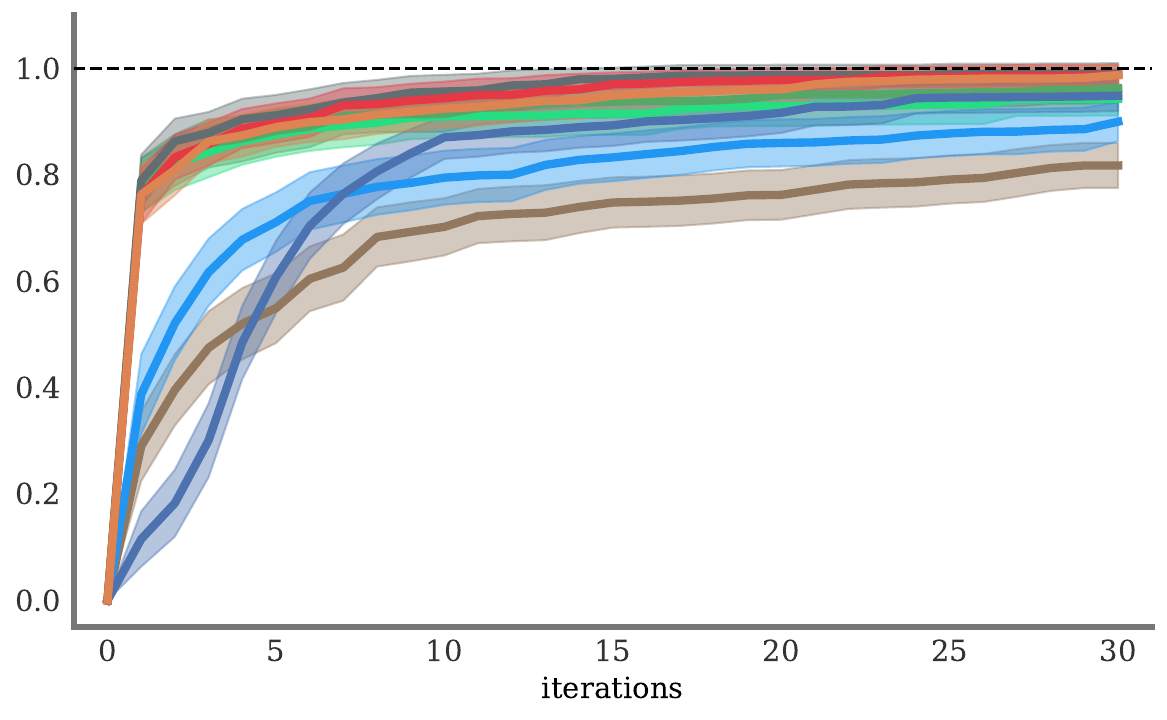}
    \end{subfigure}%
    \hspace{0.002\textwidth}%
    \begin{subfigure}[b]{0.19\textwidth}
        \includegraphics[width=\textwidth]{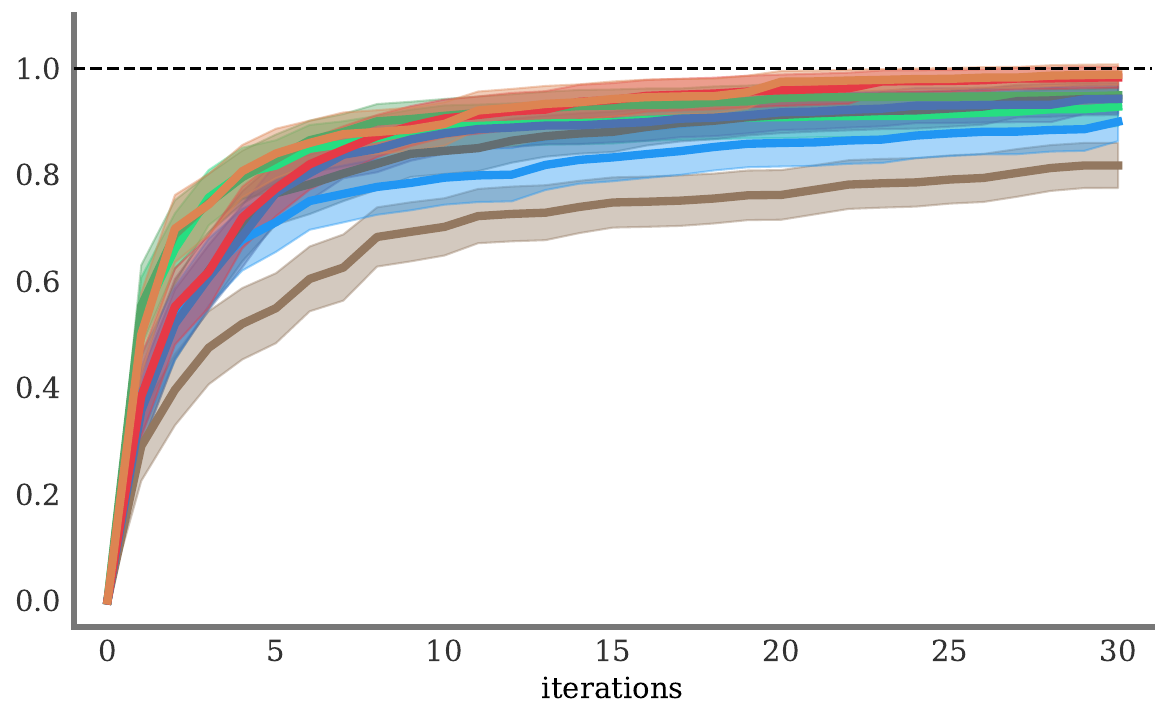}
    \end{subfigure}%
    \hspace{0.002\textwidth}%
    \begin{subfigure}[b]{0.19\textwidth}
        \includegraphics[width=\textwidth]{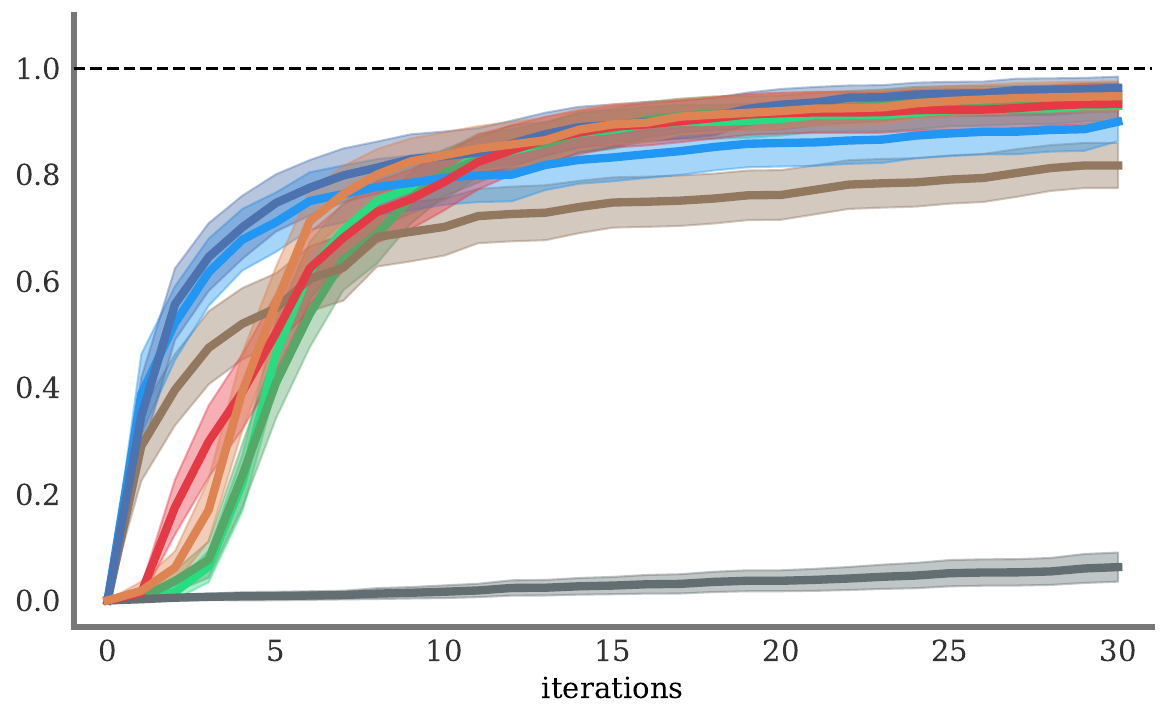}
    \end{subfigure}

    \vspace{-0.5em}

    \begin{subfigure}[b]{0.02\textwidth}
        \raisebox{0.4cm}{\rotatebox{90}{\parbox{1.9cm}{\centering\footnotesize Cum. regret ($\downarrow$)}}}
    \end{subfigure}%
    \begin{subfigure}[b]{0.19\textwidth}
        \includegraphics[width=\textwidth]{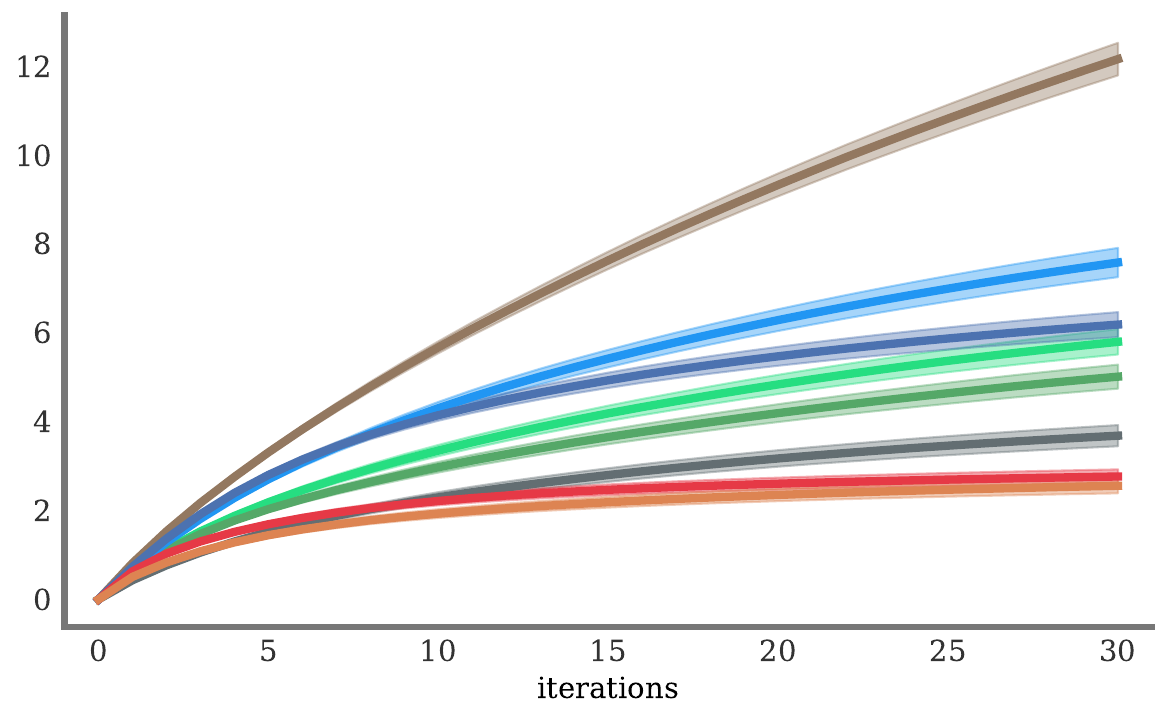}
        \caption{Marginal expert}
        \label{fig:marginal_cum}
    \end{subfigure}%
    \hspace{0.002\textwidth}%
    \begin{subfigure}[b]{0.19\textwidth}
        \includegraphics[width=\textwidth]{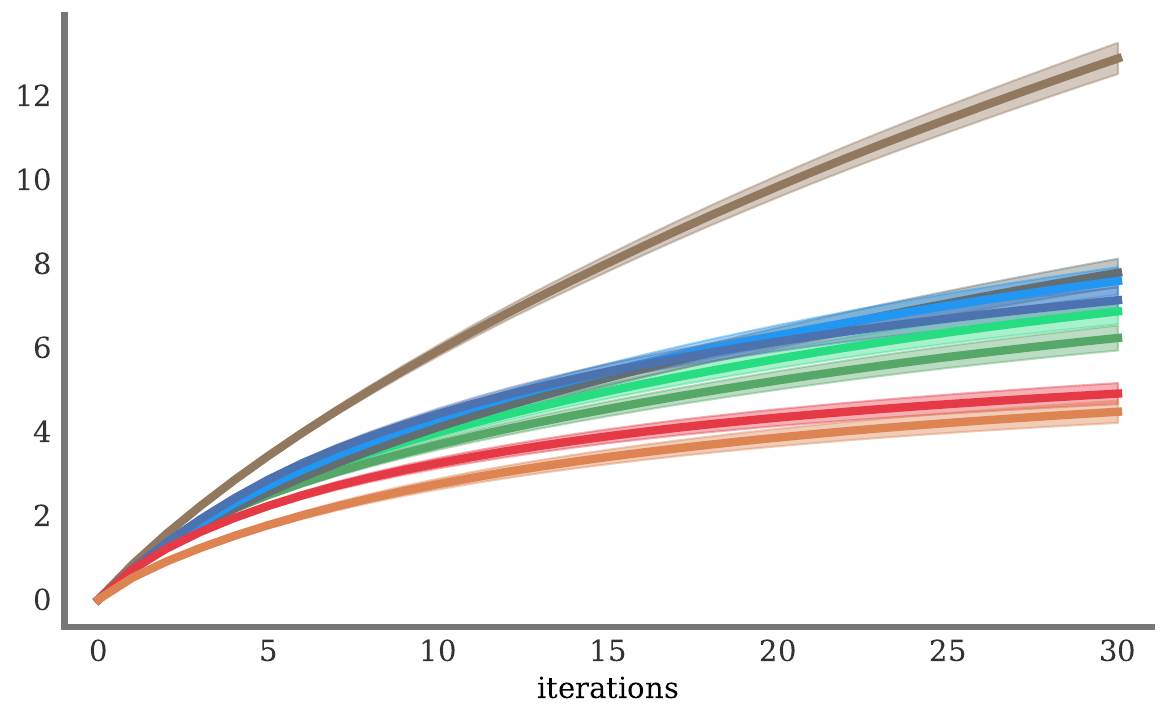}
        \caption{XGBoost expert}
        \label{fig:xgb_cum}
    \end{subfigure}%
    \hspace{0.002\textwidth}%
    \begin{subfigure}[b]{0.19\textwidth}
        \includegraphics[width=\textwidth]{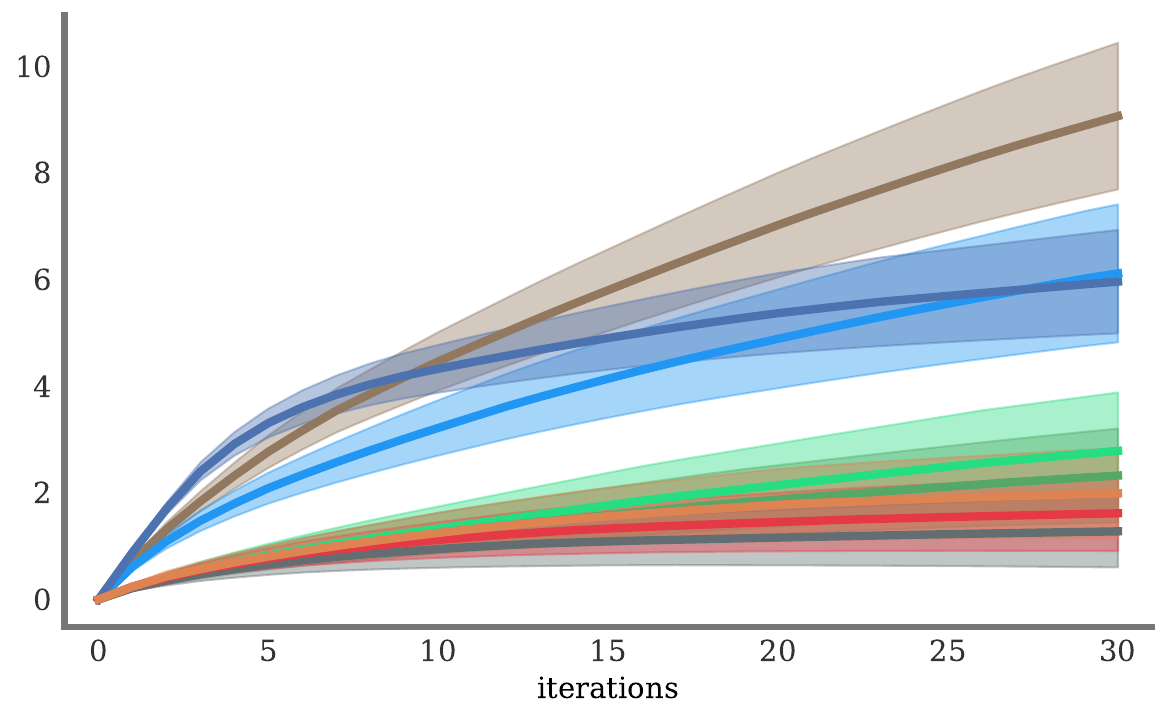}
        \caption{Branin, $\rho=0$}
        \label{fig:branin0_cum}
    \end{subfigure}%
    \hspace{0.002\textwidth}%
    \begin{subfigure}[b]{0.19\textwidth}
        \includegraphics[width=\textwidth]{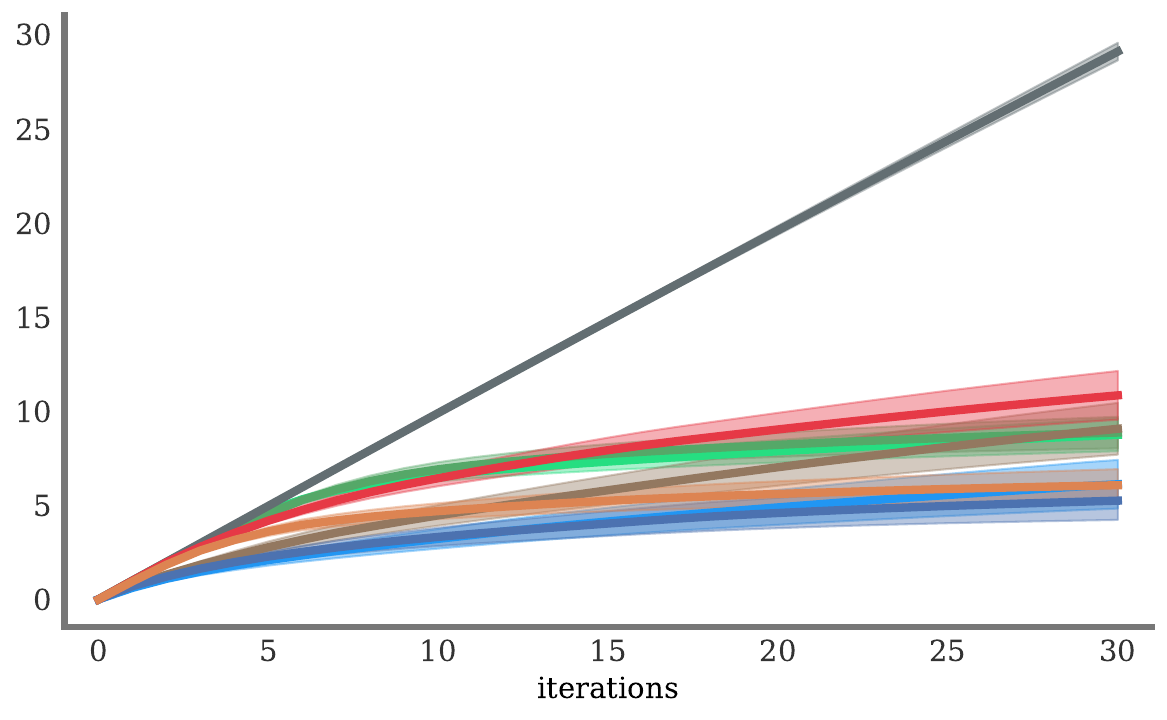}
        \caption{Branin, $\rho=0.5$}
        \label{fig:branin05_cum}
    \end{subfigure}%
    \hspace{0.002\textwidth}%
    \begin{subfigure}[b]{0.19\textwidth}
        \includegraphics[width=\textwidth]{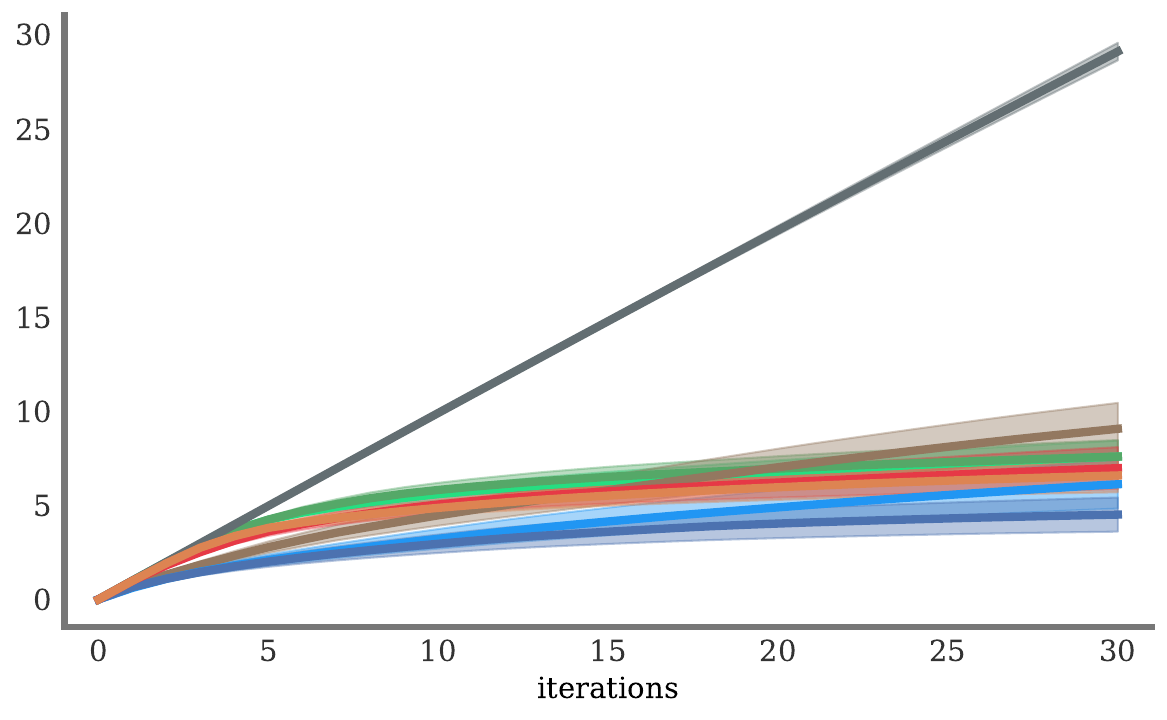}
        \caption{Branin, $\rho=0.75$}
        \label{fig:branin075_cum}
    \end{subfigure}

    \vspace{-0.2em}

    \begin{subfigure}[b]{\textwidth}
        \centering
        \includegraphics[width=0.99\textwidth]{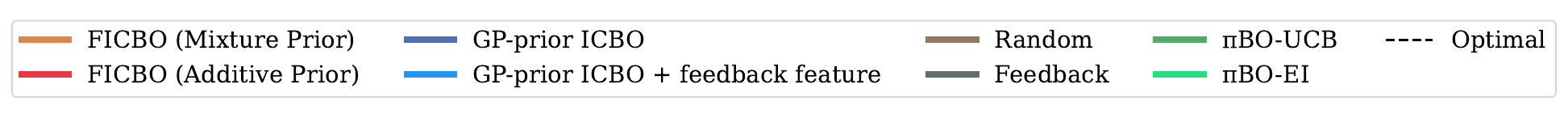}
    \end{subfigure}
    \caption{
        \textbf{Performance across synthetic benchmarks and multi-fidelity Branin tasks.}
        \textbf{Top row:} normalized best objective found so far.
        \textbf{Bottom row:} normalized cumulative regret.
        \textbf{(a--b)} Results averaged over 13 benchmark functions
        (Sphere, Ackley, Rastrigin, Rosenbrock, Schwefel, Levy, 2D and 3D versions, and Hartmann 3D) under two feedback regimes:
        marginal expert (observes only the last input dimension) and XGBoost expert (data-driven but informationally limited).
        \textbf{(c--e)} Results for multi-fidelity Branin function under three discrepancy levels ($\rho \in \{0,0.5,0.75\}$), where larger $\rho$ increases the mismatch between feedback and objective.
        Mean $\pm$ 95\% bootstrap confidence~intervals.
    }
        \vspace{-10pt}
\label{fig:combined_results}
\end{figure}

\vspace{-.2cm}
\subsection{Results}
\vspace{-.1cm}
\begin{figure}[t]
    \captionsetup[subfigure]{labelformat=empty}
    \centering

    \begin{subfigure}[b]{0.02\textwidth}
        \raisebox{0.15cm}{\rotatebox{90}{\parbox{1.8cm}{\centering\small Best found ($\uparrow$)}}}
    \end{subfigure}%
    \begin{subfigure}[b]{0.235\textwidth}
        \includegraphics[width=\textwidth]{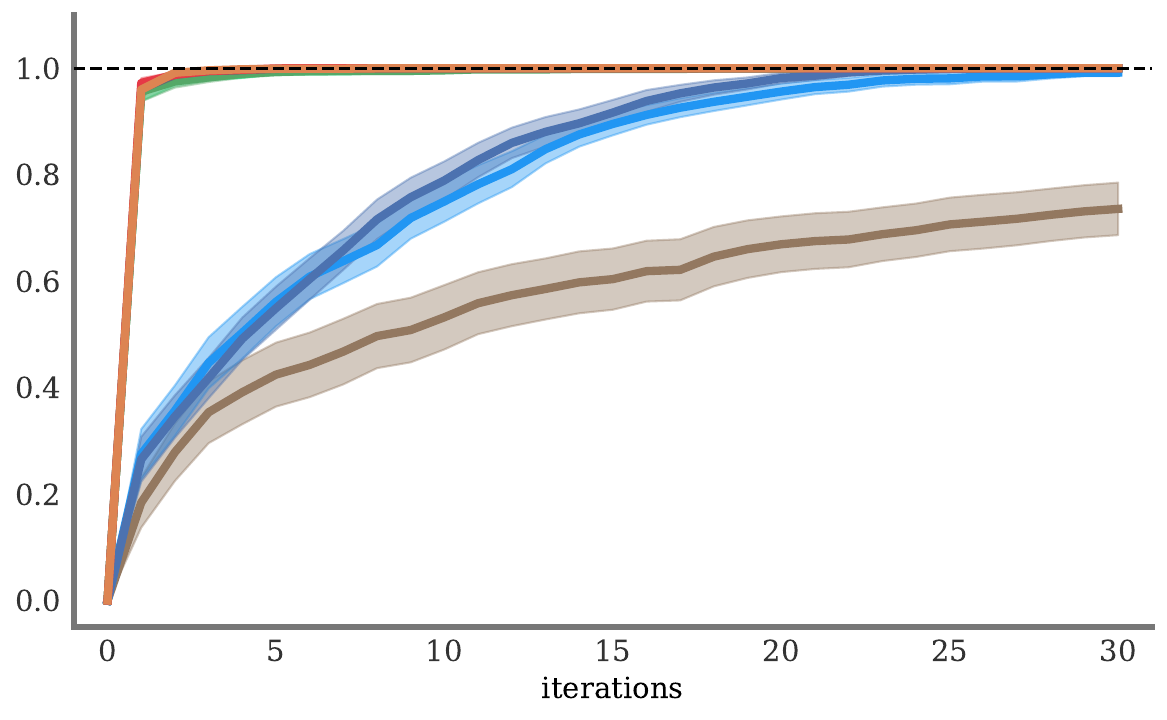}
    \end{subfigure}%
    \hspace{0.006\textwidth}%
    \begin{subfigure}[b]{0.235\textwidth}
        \includegraphics[width=\textwidth]{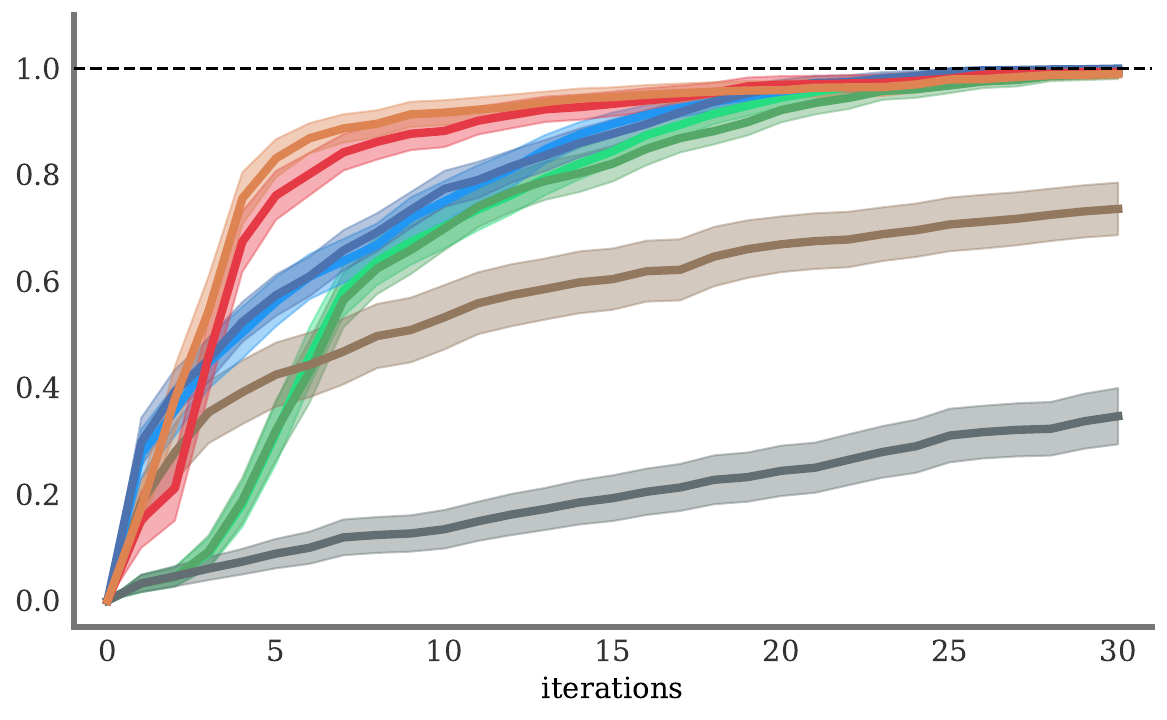}
    \end{subfigure}%
    \hspace{0.006\textwidth}%
    \begin{subfigure}[b]{0.235\textwidth}
        \includegraphics[width=\textwidth]{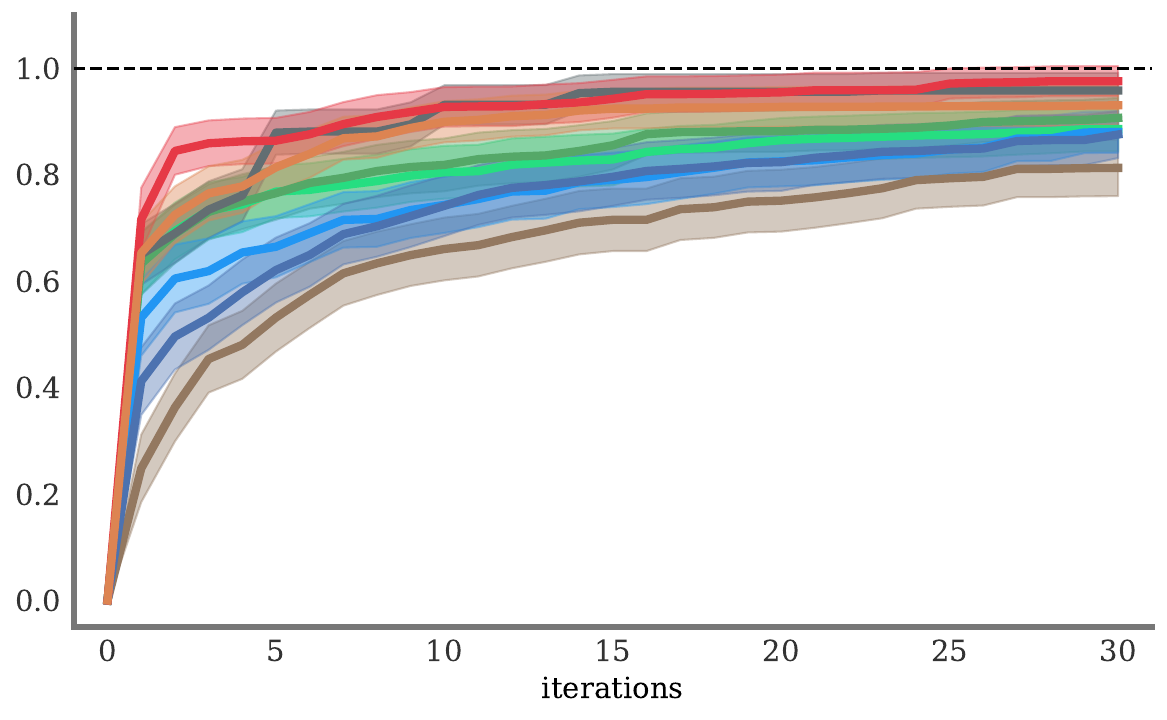}
    \end{subfigure}
    \hspace{0.006\textwidth}%
    \begin{subfigure}[b]{0.235\textwidth}
        \includegraphics[width=\textwidth]{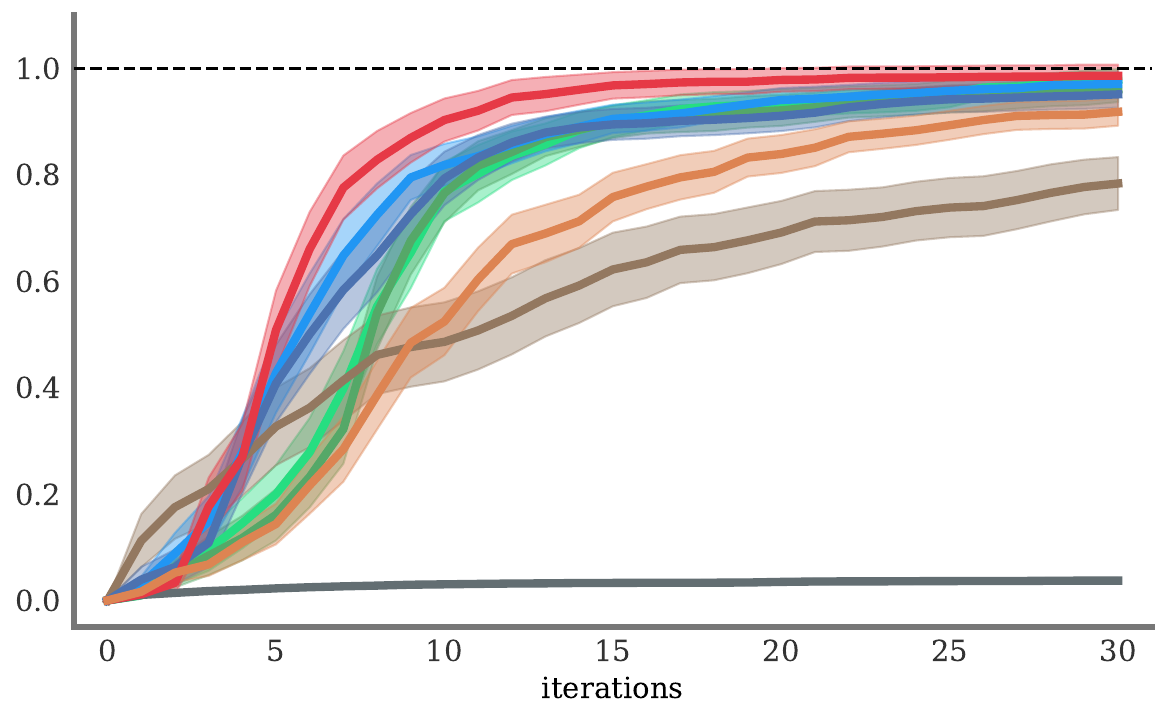}
    \end{subfigure}%

    \vspace{0.3em}

    \begin{subfigure}[b]{0.02\textwidth}
        \raisebox{0.9cm}{\rotatebox{90}{\parbox{2cm}{\centering\small Cum. regret ($\downarrow$)}}}
    \end{subfigure}%
    \begin{subfigure}[b]{0.235\textwidth}
        \includegraphics[width=\textwidth]{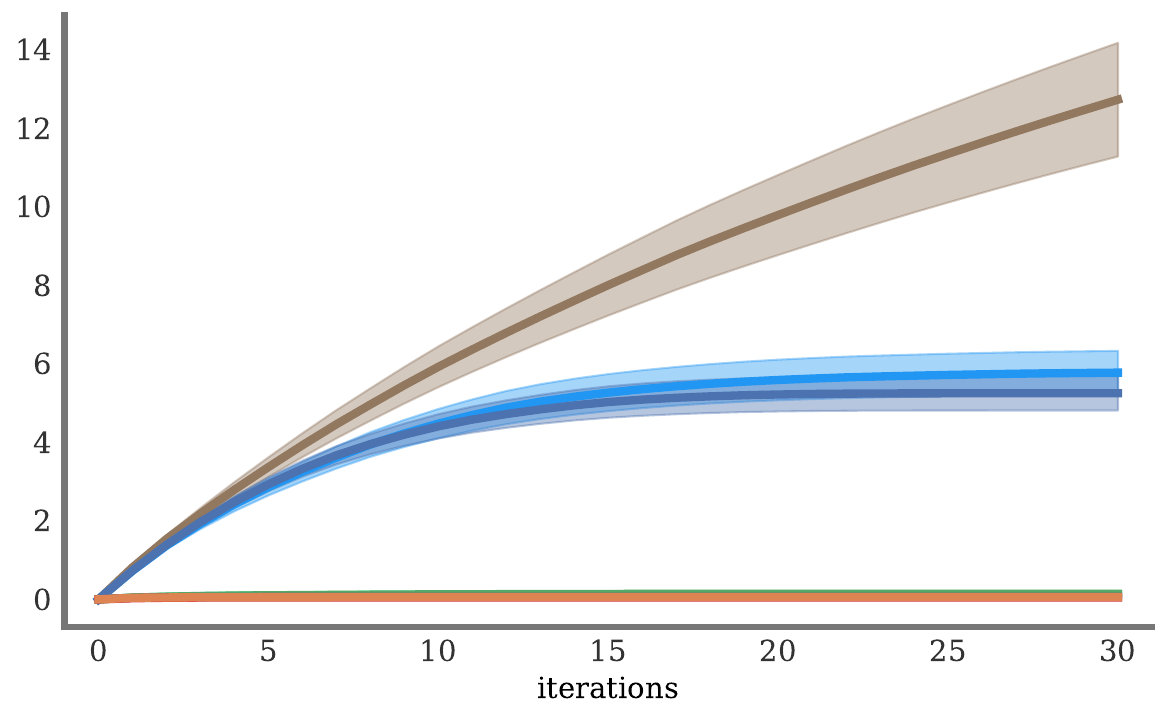}
\caption{\shortstack[c]{(a) Airfoil Small\\[-1pt] \footnotesize well-aligned feedback}}
        \label{fig:airfoil_small_cum}
    \end{subfigure}%
    \hspace{0.006\textwidth}%
    \begin{subfigure}[b]{0.235\textwidth}
        \includegraphics[width=\textwidth]{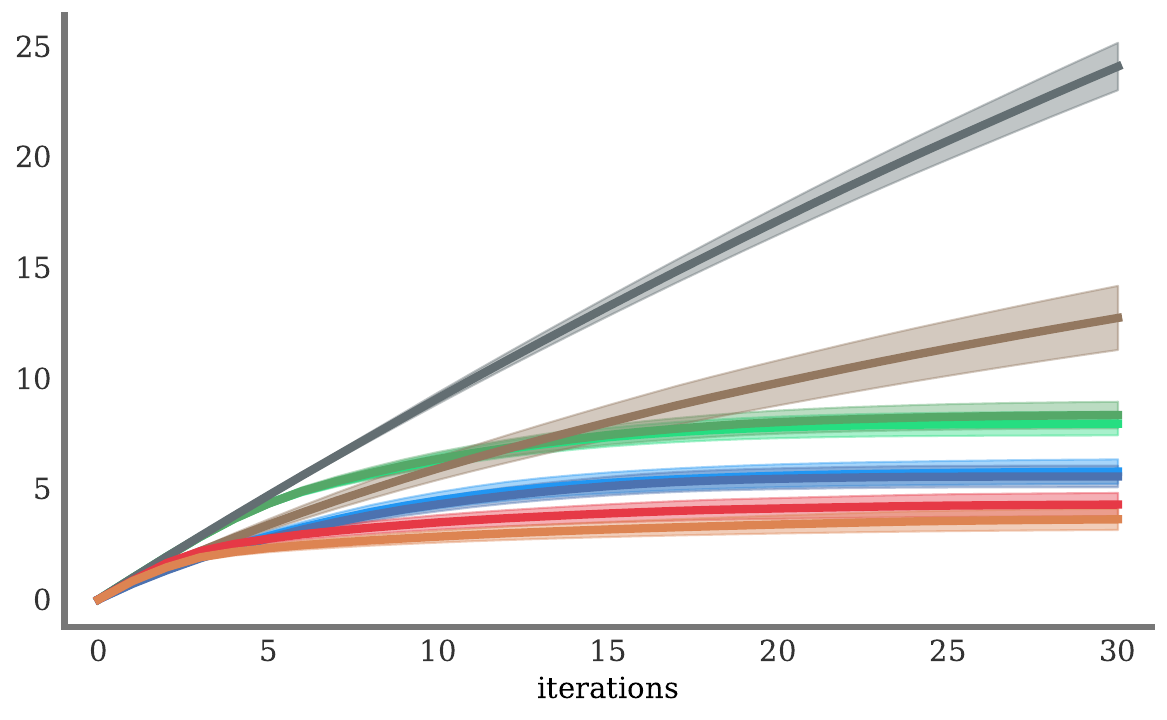}
\caption{\shortstack[c]{(b) Airfoil Lift\\[-1pt] \footnotesize structurally biased feedback}}
        \label{fig:airfoil_lift_cum}
    \end{subfigure}%
    \hspace{0.006\textwidth}%
    \begin{subfigure}[b]{0.235\textwidth}
        \includegraphics[width=\textwidth]{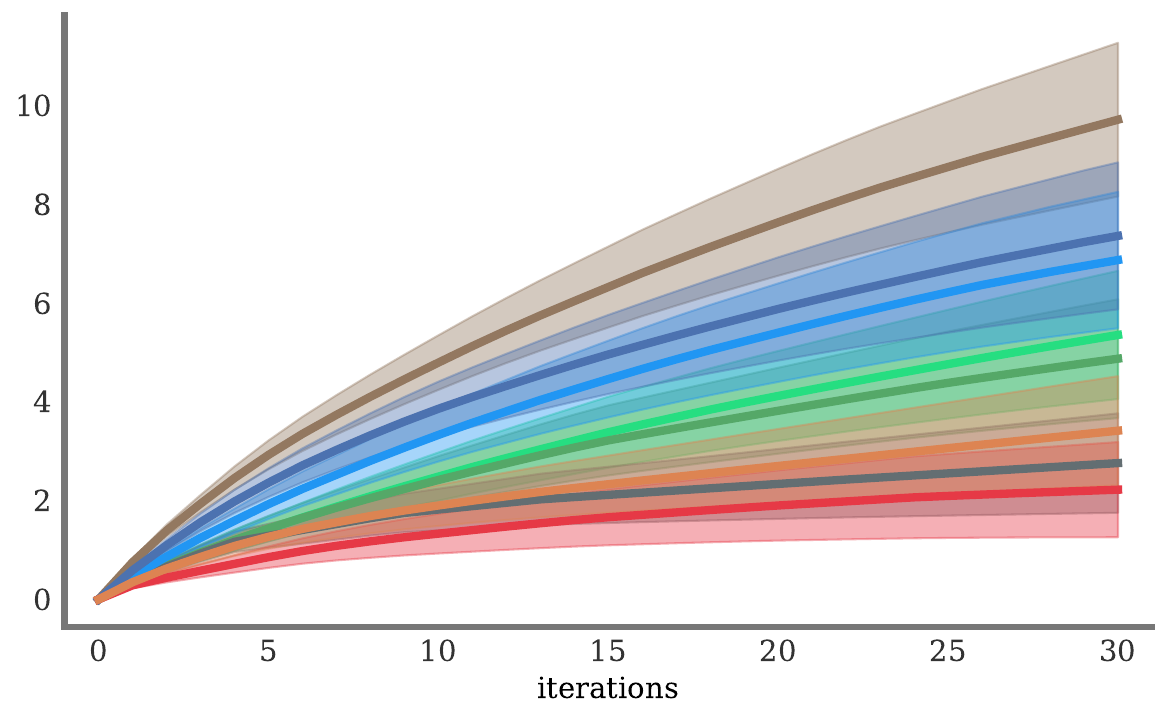}
\caption{\shortstack[c]{(c) Neural Architecture Search\\[-1pt] \footnotesize informative feedback}}
        \label{fig:mlhp_cum}
    \end{subfigure}
    \hspace{0.006\textwidth}%
    \begin{subfigure}[b]{0.235\textwidth}
        \includegraphics[width=\textwidth]{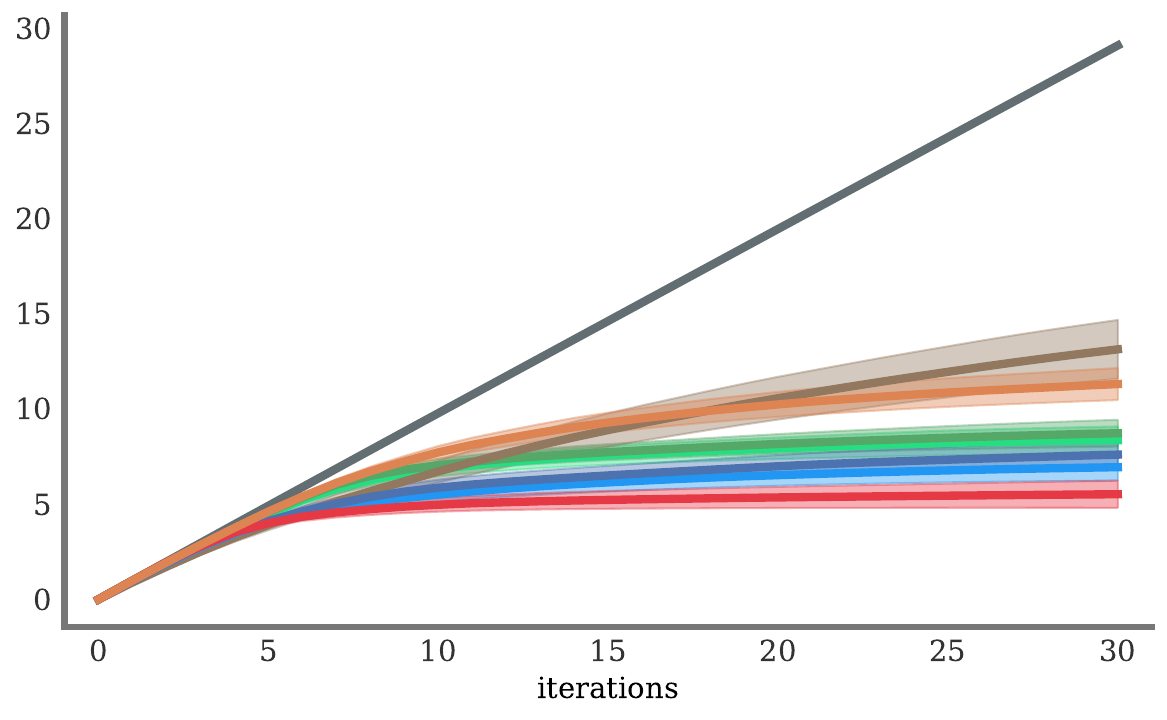}
\caption{\shortstack[c]{(d) Reactor\\[-1pt] \footnotesize misleading feedback}}        \label{fig:reactor_cum}
    \end{subfigure}%

    \vspace{-0.2em}
    \begin{subfigure}[b]{\textwidth}
        \centering
        \includegraphics[width=0.99\textwidth, trim=0 0 0 10pt, clip]{img/legends/Legend.pdf}
    \end{subfigure}

    \caption{
        \textbf{Real-world optimization tasks.}
        \textbf{Top row:} normalized best value found.
        \textbf{Bottom row:} normalized cumulative regret.
        \textbf{(a)} Airfoil optimization with a cheaper neural surrogate
        as feedback (Airfoil Small); \textbf{(b)} airfoil optimization with lift-only feedback (Airfoil Lift); \textbf{(c)}
        Neural Architecture Search (NAS) with a cheap auxiliary performance signal; and \textbf{(d)} reactor optimization with conversion-based feedback.
        Mean $\pm$ 95\% bootstrap confidence
        intervals. \textbf{FICBO exploits aligned feedback, but shows prior-dependent robustness to structurally biased proxies in the Reactor example.} 
        The curves in the Airfoil Small panel overlap: FICBO and Feedback find the best solution within the first steps because the feedback is well aligned; $\pi$BO is slightly slower (Table~\ref{tab:ablation}).
    }
    \vspace{-5pt}
\label{fig:realworld_results}
\end{figure}

\subsubsection{Synthetic benchmarks}
\vspace{-.1cm}

Figure~\ref{fig:combined_results}\textbf{a-b} first evaluates out-of-distribution analytic functions under the two feedback regimes of Table~\ref{tab:tasks}. The \emph{marginal expert} gives a signal that is directionally useful but incomplete, while the \emph{XGBoost expert} is additionally limited by its training data and model class (\cref{sec:benchmarkfunctions}).

The results show a hierarchy in how feedback is used. First, in both regimes, the naive feedback-feature baseline improves over the feedback-agnostic amortized model (\legendline{gpFb} $>$ \legendline{gpBase}), indicating that the auxiliary signal contains exploitable information even when treated as an ordinary input coordinate.
Second, the $\pi$BO variants
(\legendline{piboUCB} $\pi$BO-UCB; \legendline{piboEI} $\pi$BO-EI) improve over the generic amortized prior by using feedback $u(\x)$ as a factor on the acquisition function, suggesting it is a stronger inductive bias than simple feature concatenation (GP-prior ICBO + feedback feature) when feedback is informative. Finally, FICBO
(\legendline{ficboAdd} Additive; \legendline{ficboMix} Mixture) achieves the lowest regret in both regimes. 
Unlike $\pi$BO, whose feedback influence follows a fixed schedule, FICBO is pretrained under a prior over source usefulness and can infer from the observed history how much the current source should be trusted. 
Comparisons to more baselines can be found in \cref{app:additional_Baselines}.

The XGBoost regime is harder because feedback errors include extrapolation and model misspecification rather than only marginalization, but the same qualitative ordering remains, suggesting that the learned trust calibration transfers beyond the exact synthetic prior.

Lastly, Figure~\ref{fig:combined_results}\textbf{c--e} shows settings where feedback is a smooth low-fidelity variant with increasing discrepancy $\rho$ (Appendix~\ref{sec:branin}).  As Figure~\ref{fig:branin} shows, larger $\rho$ mainly corrupts the \emph{ranking} induced by feedback, moving from near-aligned at $\rho=0$ to rank-misaligned at $\rho=0.75$. This explains the performance transition: feedback-aware methods benefit strongly at $\rho=0$, FICBO remains competitive at the weakly informative intermediate setting $\rho=0.5$, while at $\rho=0.75$, methods that rely heavily on feedback, most noticeably the naive feedback baseline, are pulled toward misleading regions. This case suggests an out-of-prior regime with a smooth but globally misaligned proxy.

\vspace{-.2cm}
\subsubsection{Real-world tasks}
\vspace{-.2cm}
The real-world tasks complement the analytical benchmarks with practical optimization objectives and cheap proxies that arise naturally in each setting.
The airfoil experiments optimize the lift-to-drag ratio $C_L/C_D$ of a NACA airfoil over angle of attack, thickness, and camber.
We use a higher-accuracy NeuralFoil configuration as the benchmark oracle, standing in for a more expensive aerodynamic evaluation (Appendix~\ref{sec:aero}).
On Airfoil Small (Figure~\ref{fig:realworld_results}\textbf{a}), the errors of the proxy reflect only its approximation quality. 
The greedy feedback baseline is already strong, as expected, and FICBO remains competitive, which shows that robustness to unreliable sources does not prevent the model from exploiting genuinely informative feedback.

Airfoil Lift (Figure~\ref{fig:realworld_results}\textbf{b}) represents a structurally biased feedback regime. 
The proxy uses lift alone: it captures the numerator $C_L$ of the objective but ignores the drag penalty $C_D$. 
This is misleading because the true ratio $C_L/C_D$ peaks before stall, where lift is high but drag remains moderate; at larger angles of attack, drag rises sharply and the ratio deteriorates.
A source based only on lift can therefore keep recommending high-lift designs past the true optimum, so the feedback-only baseline accumulates large regret, whereas FICBO uses
early objective evaluations to detect this mismatch and discount the proxy.
This is the setting in which source-aware pretraining is most useful: the signal is a partially correct heuristic whose failure mode must be inferred in context.
Next, on the Neural Architecture Search task (Figure~\ref{fig:realworld_results}\textbf{c}), CIFAR-10 accuracy transfers usefully but does not rank the best architectures correctly (Appendix~\ref{sec:nas}). 
On the two-dimensional reactor task (Figure~\ref{fig:realworld_results}\textbf{d}), conversion of reactant A is a cheap proxy for intermediate-product yield, but becomes misleading once high conversion favors the final product instead (Appendix~\ref{sec:reaction}). 
In this harder regime, the additive prior is more robust than the mixture prior, suggesting that the choice of feedback prior matters when the proxy is weakly informative or structurally biased.

\vspace{-.1cm}
\subsubsection{Interpreting feedback use}\label{sec:interpret}
\vspace{-.1cm}

We use two post-hoc diagnostics to inspect how FICBO relates to the auxiliary source at test time.
The first measures whether the learned policy continues to select candidates favored by the source as BO observations accumulate; the second visualizes how the feedback-error profiles of test tasks compare to the pretraining prior. Neither diagnostic is used during pretraining or deployment.

\begin{figure}[h]
\centering
\includegraphics[width=.55\linewidth]{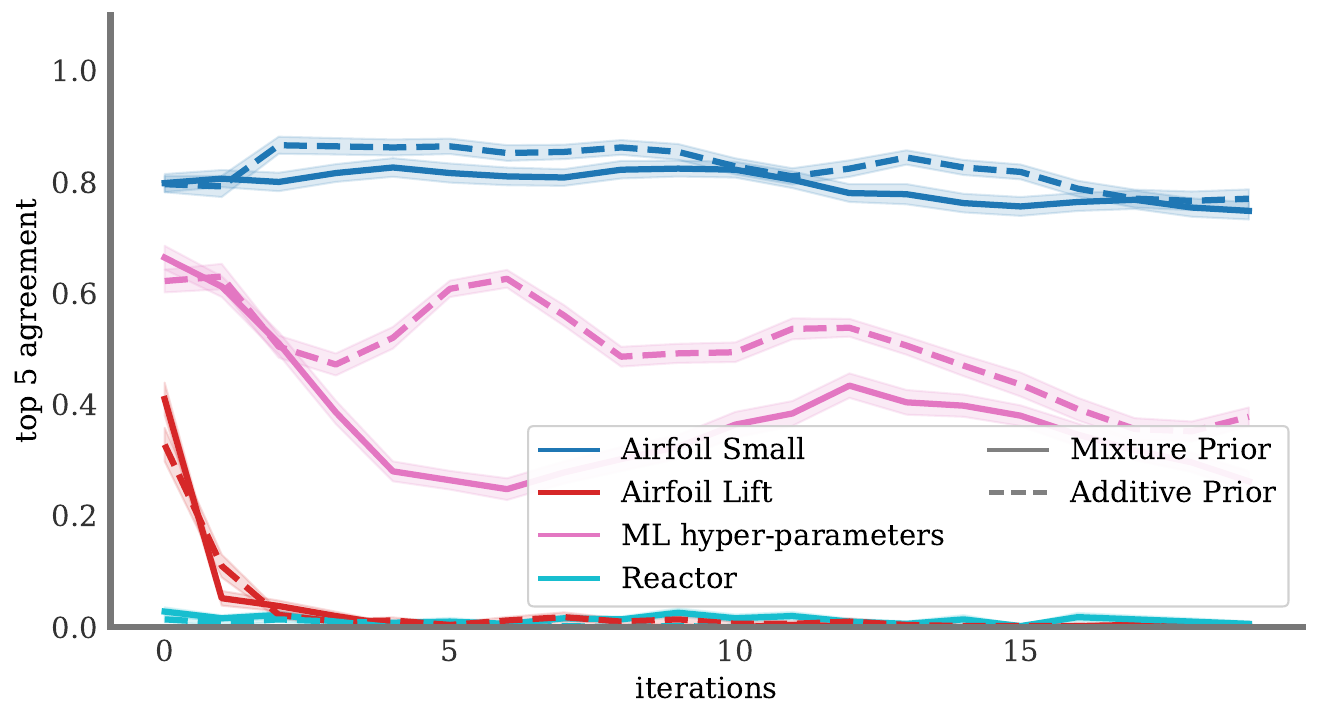}
\includegraphics[width=.44\linewidth]{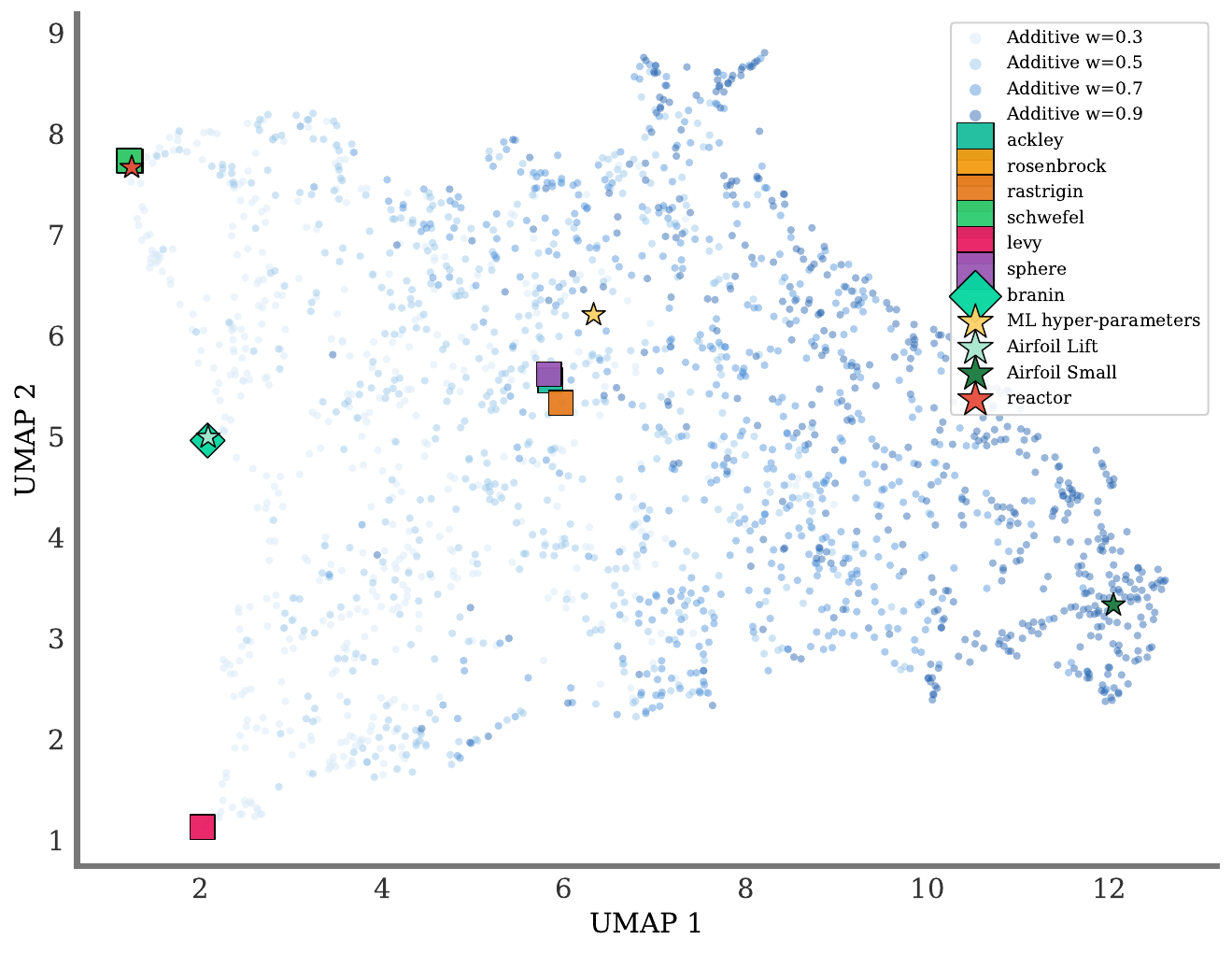}
\caption{
\textbf{Post-hoc diagnostics of feedback use.}
\textbf{Left:} top-$5$ policy--feedback agreement over BO iterations for FICBO. Mean $\pm$ std computed over 100 replicas.
\textbf{Right:} UMAP embedding of feedback-error profiles for synthetic prior samples and test tasks. Synthetic points are generated without additional bias operators; benchmark points use the 3D marginal-expert setting, where the expert observes the last dimension and the remaining dimensions are marginalized out.
}
    \vspace{-10pt}
\label{fig:umap}
\end{figure}

\textbf{Policy--feedback agreement.}
The left panel of Figure~\ref{fig:umap} reports a top-5 policy--feedback agreement diagnostic for both FICBO priors. At each BO step, we record the overlap between the policy's 5 highest-rated query points and the 5 highest auxiliary feedback points.
This asks whether the policy's preferred designs are also strongly favored by the source.

On Airfoil Small, where feedback is a lower-accuracy evaluator of the same objective, agreement stays close to one for both priors, consistent with sustained use of a reliable proxy. 
Airfoil Lift shows the opposite pattern: agreement starts non-negligible but collapses within a few BO steps as observations reveal that lift $C_L$ alone is misleading for $C_L/C_D$. 
The reactor task has near-zero agreement almost from the start, indicating that the policy largely avoids the conversion proxy. 
The Neural Architecture Search task lies between these extremes: CIFAR-10 accuracy is useful for transfer, but agreement decreases over time, because CIFAR-10 accuracy does not fully determine the ImageNet16-120 ranking, and the rate of decrease differs between priors. 
Overall, the curves show task- and prior-dependent source reliance rather than fixed decay.
$\pi$BO imposes decay via a hand-tuned schedule~\citep{hvarfner2022pibo,adachi}; FICBO revises reliance from observed evidence.

\textbf{Feedback-profile embedding.}
The right panel provides a complementary static view of the source itself. 
Each task is represented by the binned signed and absolute errors $d_i = u_i - y_i$, with points sorted by feedback value, and then projected with UMAP (Appendix~\ref{sec:umap}).
This is a qualitative coverage diagnostic, suggesting which test sources resemble high-alignment samples from the feedback prior and which occupy different regions of feedback--objective disagreement. 
For example, Airfoil Small lies near highly aligned synthetic sources, whereas biased or less directly useful proxies, such as Airfoil Lift and the reactor feedback, appear away from this regime.  
Finally, Appendix~\ref{app:rollout} provides a qualitative rollout illustrating how FICBO updates its posterior and policy distribution as new observations revise its reliance on the feedback signal.

\vspace{-.2cm}
\subsection{Ablation study}
\vspace{-.2cm}

\begin{wraptable}[15]{r}{0.5\textwidth}
    \vspace{-0.9\baselineskip}
    \centering
    \scriptsize
    \setlength{\tabcolsep}{3pt}
    \renewcommand{\arraystretch}{0.87}
    \begin{threeparttable}
    \begin{tabular}{@{}lcc@{}}
    \toprule
    Method & Airfoil Small & Airfoil Lift \\
    \midrule
    FICBO (Mixture) & $0.020_{\pm 0.068}$ & $3.617_{\pm 2.401}$ \\
    Mixture, no bias operators & $0.029_{\pm 0.102}$ & $10.386_{\pm 6.063}$ \\
    FICBO (Additive) & $0.014_{\pm 0.062}$ & $4.295_{\pm 2.841}$ \\
    Additive, no bias operators & $0.014_{\pm 0.060}$ & $4.999_{\pm 2.494}$ \\
    \midrule
    \multicolumn{3}{@{}l}{\textit{Additive-prior signal variants (no source-model fit\tnote{a})}} \\
    $g_\tau$-based feedback (standard) & $0.038_{\pm 0.079}$ & $5.206_{\pm 2.975}$ \\
    $g_\tau$-based, no catastrophic & $0.032_{\pm 0.078}$ & $5.747_{\pm 3.742}$ \\
    $g_\tau$-based, no shift & $0.016_{\pm 0.062}$ & $5.924_{\pm 3.452}$ \\
    $g_\tau$-based, no biases & $0.032_{\pm 0.093}$ & $7.026_{\pm 3.542}$ \\
    $f_\tau$-based, standard biases & $0.021_{\pm 0.080}$ & $24.649_{\pm 5.341}$ \\
    \midrule
    $\pi$BO-EI & $0.097_{\pm 0.388}$ & $7.807_{\pm 2.486}$ \\
    $\pi$BO-UCB & $0.113_{\pm 0.422}$ & $8.167_{\pm 2.834}$ \\
    Feedback & $0.020_{\pm 0.079}$ & $24.179_{\pm 5.486}$ \\
    Random & $12.692_{\pm 7.489}$ & $12.692_{\pm 7.489}$ \\
    \bottomrule
    \end{tabular}
    \begin{tablenotes}
    \footnotesize
    \item[a] The direct-source branch of Eq.~\eqref{eq:feedback-pipeline} is used.
    \end{tablenotes}
    \end{threeparttable}
    \caption{
    Final cumulative regret ($\downarrow$) on Airfoil Small and Airfoil Lift. Mean $\pm$ std over 100 seeds.
    }
    \label{tab:ablation}
\end{wraptable}
Table~\ref{tab:ablation} contrasts well-aligned feedback (Airfoil Small) with a structurally biased proxy (Airfoil Lift).
Removing bias operators barely affects Airfoil Small but clearly hurts Airfoil Lift, especially for the Mixture prior, suggesting that bias augmentation mainly improves robustness to misleading sources.
The clearest ablation comes from the signal-variant rows.
Using $g_\tau$ directly, without source-model fitting, degrades Airfoil Lift performance relative to the full Additive-prior model, suggesting that the fitting step helps.
By contrast, generating pretraining feedback from the full objective $f_\tau$ makes the source overly informative:
performance remains strong on Airfoil Small but falls to the feedback-only baseline on Airfoil Lift, suggesting that overly reliable pretraining feedback can induce over-trust in structurally biased~proxies.

The bias ablations support this conclusion. 
Removing all bias operators worsens performance on Airfoil Lift, indicating that exposure to distorted sources helps the model learn when to discount feedback. 
The smaller effect of removing individual operators suggests that the benefit comes from diverse feedback--objective relationships rather than any single distortion type. 
The bias family thus broadens the range of source reliabilities seen offline, so the model learns to calibrate trust in context instead of assuming feedback is a faithful proxy for the~objective.

\vspace{-.2cm}
\section{Discussion}
\vspace{-.2cm}

We introduced FICBO, an in-context optimization framework for settings with cheap but unreliable auxiliary feedback.
It treats feedback as a stochastic information source and pretrains both posterior prediction and query selection under a prior over feedback access, relevance, and distortion.
Across synthetic benchmarks and real-world tasks, the resulting optimizer learns to exploit aligned feedback while discounting unreliable proxies from in-context evidence.

\textbf{Limitations.}
A central challenge is coverage of the feedback prior: transfer is strongest when the prior spans the main ways in which deployment feedback can agree or disagree with the objective. The feedback-profile embedding in Figure~\ref{fig:umap} provides one diagnostic for this coverage, while the Branin stress test illustrates a difficult regime: smooth but globally rank-misaligned feedback. Second, we assume feedback is available over the candidate pool, which is natural for cheap simulators or pretrained proxies, but expensive feedback would require an additional acquisition mechanism. 
Third, the feedback source is fixed during the BO episode and does not update after new objective evaluations, as opposed to changing feedback sources, as when dealing with human feedback. 
Limiting experiments to low dimensions leaves open the question of whether the approach scales to high-dimensional search spaces, where surrogate modelling and feedback alignment may behave differently.
Finally, pretraining requires a horizon distribution, so very different test-time budgets are formally out-of-distribution; in practice, models trained with $T_\tau\in [10, 20]$ were successfully evaluated for $T=30$.

\textbf{Perspectives.}
A natural next step is to combine feedback-aware pretraining with richer objective priors, either problem-specific or broadly diversified in the spirit of tabular foundation models~\citep{qu2026tabiclv2}. 
For instance, human-in-the-loop simulators or cognitive-agent models could provide partially reliable feedback priors~\citep{zhu2026more}.
The goal would be to cover not only a wide range of objective functions, but also a wide range of feedback--objective relationships. Another direction is to handle multiple test-time feedback sources, learning which cheap but imperfect signals to combine for each task.

\begin{ack}
This work was supported by the Research Council of Finland (Flagship programme: Finnish Center
for Artificial Intelligence FCAI) and the EU funding ERC ODD-ML 101201120. NB and SK were supported by the Research Council of Finland
(RCF-NSF FinBioFAB, grant agreement 365982). 
SK was also supported by UKRI Turing AI World-Leading Researcher Fellowship (EP/W002973/1).
JM was supported by the Research Council of Finland (grant agreement 377145).
The authors acknowledge the computational resources provided by the Aalto Science-IT Project from
Computer Science IT.
Views and opinions expressed are however those of the author(s) only and do not necessarily reflect those of the European Union or the European
Research Council Executive Agency. 
Neither the European Union nor the granting authority can be held responsible for them.
\vspace{1em} %

\noindent
\begin{minipage}{\textwidth}
  \centering
  \includegraphics[height=1.5cm]{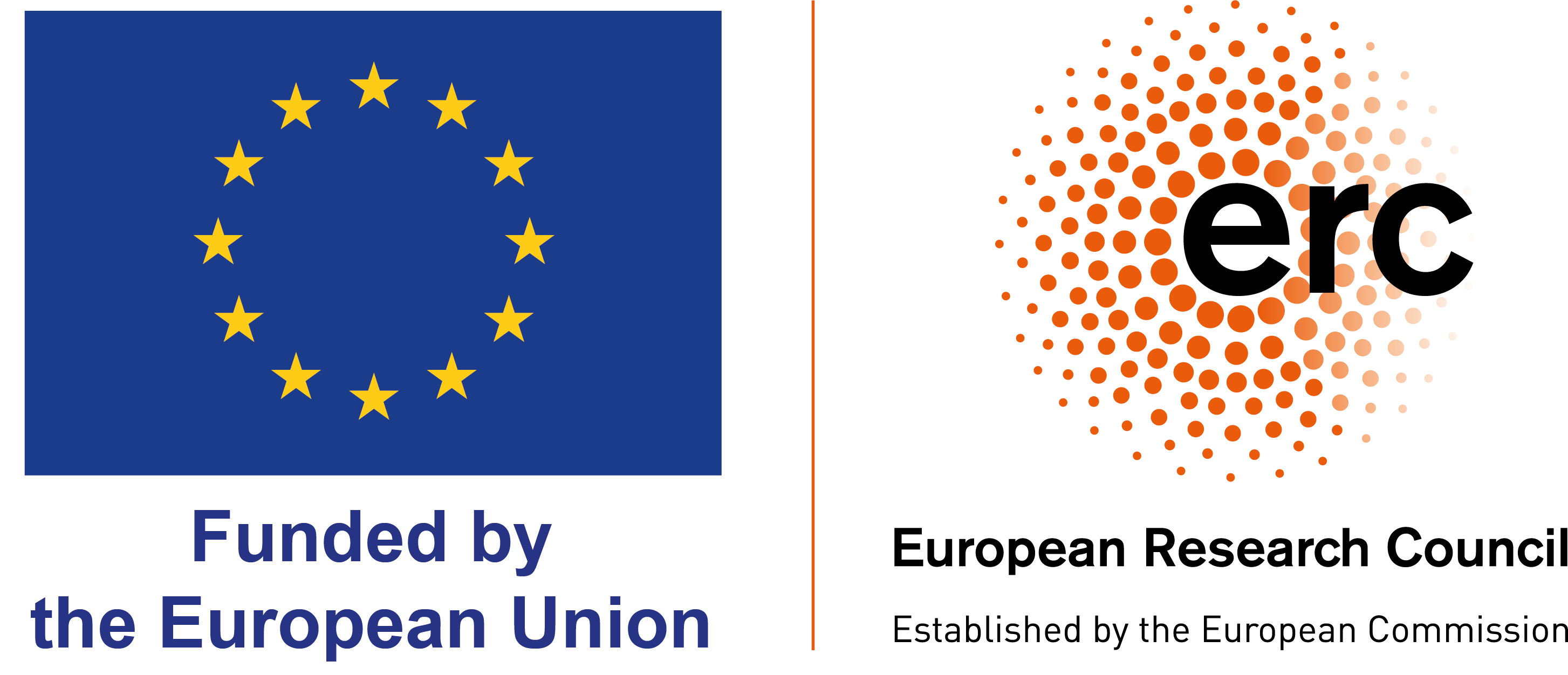}
  \hfill
\end{minipage}
\end{ack}

\bibliographystyle{unsrtnat}
\bibliography{references}
\newpage
\appendix

\setcounter{figure}{0}
\setcounter{table}{0}
\setcounter{equation}{0} %
\renewcommand{\thefigure}{S\arabic{figure}}
\renewcommand{\thetable}{S\arabic{table}}
\renewcommand{\theequation}{S\arabic{equation}}

The appendix is organized as follows:
\begin{itemize}
    \item \textbf{Appendix~\ref{app:addexp}} presents additional experiments, including classical-GP $\pi$BO comparisons and a qualitative policy rollout.
    \item \textbf{Appendix~\ref{app:synthgen}} details synthetic pretraining task generation, including GP draws, feedback priors, and structured bias operators.
    \item \textbf{Appendix~\ref{app:details_vis}} provides baseline and benchmark details, including synthetic and real-world tasks and the feedback-profile embedding diagnostic.
    \item \textbf{Appendix~\ref{app:algs}} describes the MDP formulation, training objective, deployment modes, and pretraining algorithm.
    \item \textbf{Appendix~\ref{app:detailed_results}} reports detailed per-benchmark results across feedback configurations.
    \item \textbf{Appendix~\ref{app:Continuous}} investigates continuous optimization through gradient-based query refinement and compares it with pool-based selection.
    \item \textbf{Appendix~\ref{app:architecture}} specifies the model architecture, computational resources, and hyperparameters.
\end{itemize}

\section{Additional experiments}\label{app:addexp}
\subsection{Classical-GP baselines}

\definecolor{piboUCBGP}{RGB}{160,80,220}  %
\definecolor{piboEIGP}{RGB}{255,105,180}  %

The main text compares only amortized in-context baselines, so that differences in performance can be attributed to how the feedback channel is integrated. In particular, our $\pi$BO-style baselines use the GP-prior ICBO surrogate: the inference head outputs a Gaussian-mixture posterior, from which we extract moments to form EI or UCB scores before applying the feedback reweighting. 
This isolates the role of feedback integration, but it is not necessarily the strongest possible implementation of $\pi$BO.

\begin{figure}[H]
    \centering
    \newcommand{\rowlabel}[1]{%
        \rotatebox{90}{\parbox{3.5cm}{\centering\small #1}}}

    \begin{subfigure}[b]{0.46\textwidth}
        \includegraphics[width=\textwidth]{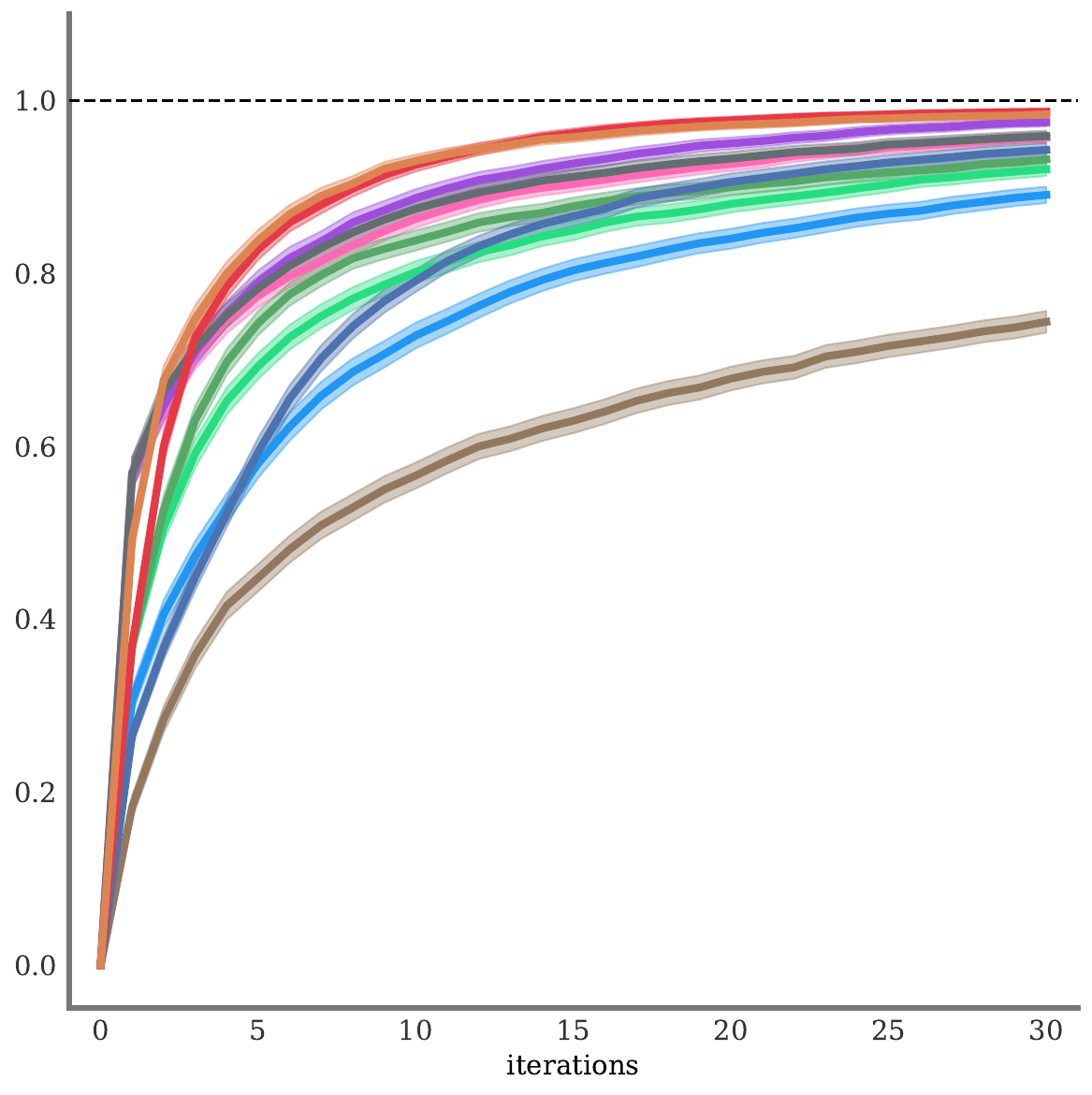}
        \caption{Best-so-far ($\uparrow$)}
    \end{subfigure}
    \hfill
    \begin{subfigure}[b]{0.46\textwidth}
        \includegraphics[width=\textwidth]{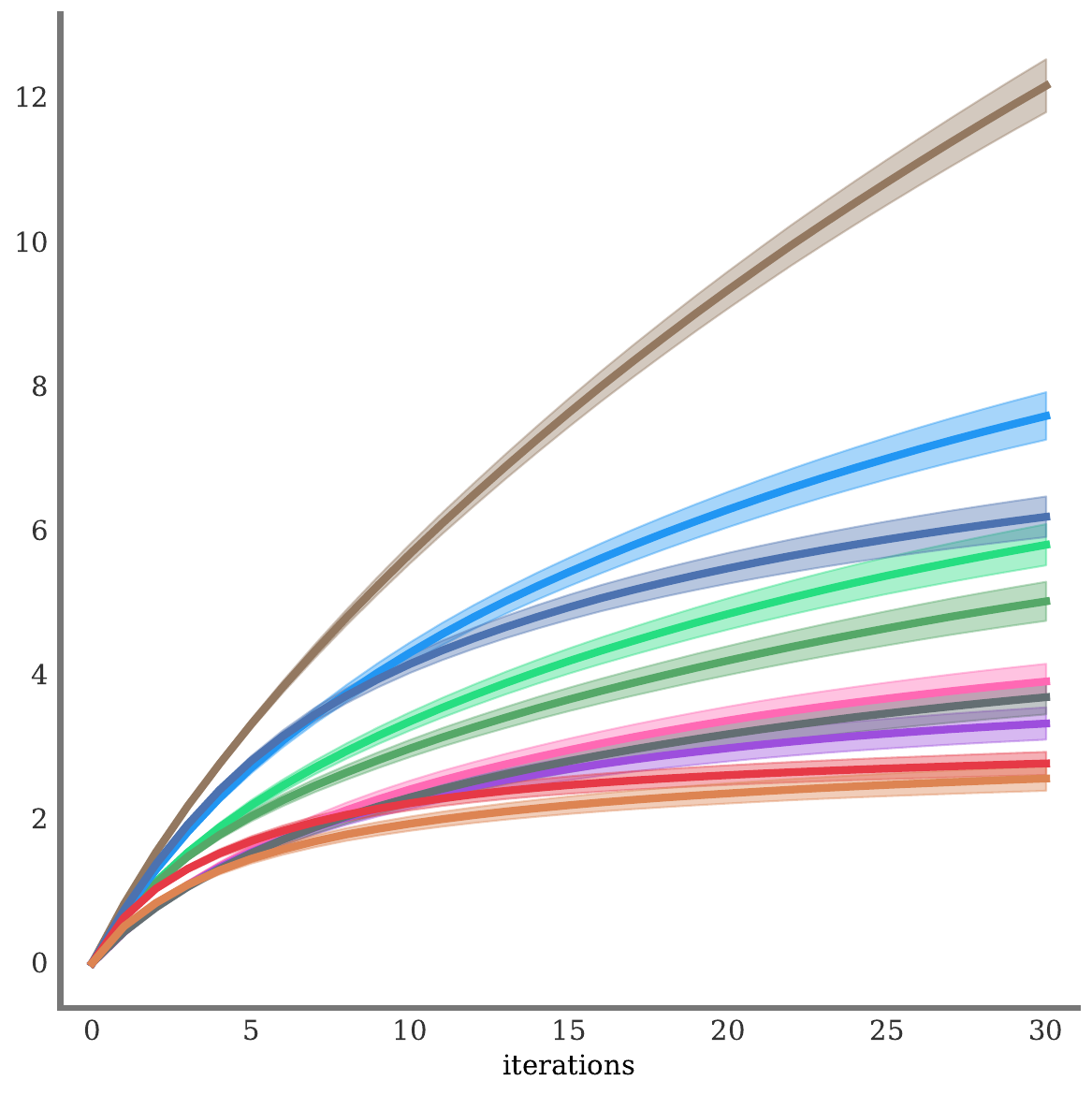}
        \caption{Cumulative regret ($\downarrow$)}
    \end{subfigure}

    \begin{subfigure}[b]{0.96\textwidth}
        \centering
        \includegraphics[width=0.99\textwidth]{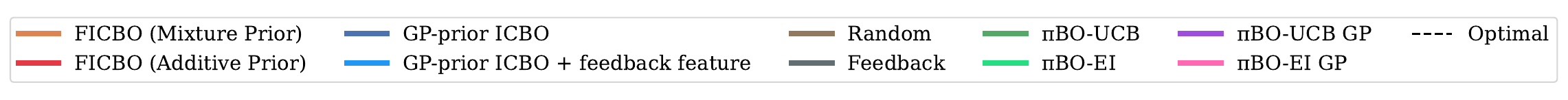}
    \end{subfigure}
    \caption{
    \textbf{Comparison with classical-GP $\pi$BO baselines on marginalized benchmark functions.}
    The main-text $\pi$BO baselines use the GP-prior ICBO surrogate.
    Here, $\pi$BO-UCB GP and $\pi$BO-EI GP instead use a classical GPyTorch GP surrogate with closed-form EI and UCB, refit online at each step.
    Mean $\pm$ 95\% bootstrap confidence intervals.
    \textbf{GP-based $\pi$BO improves over the in-context
    $\pi$BO variants, but FICBO remains best in both best-so-far value and cumulative regret.}
    }
    \label{fig:comp_BoTorch}
\end{figure}

Ultimately, however, black-box optimization is judged by the quality of the designs it finds, regardless of whether the optimizer is amortized or fit online.
For completeness, we also evaluate $\pi$BO with a classical GP surrogate (batched GPyTorch exact GP, ARD Matérn-3/2 kernel, 100 Adam steps per BO step, warm-started from the previous step) and closed-form EI and UCB, refit online at every step.
This gives a stronger non-amortized comparison: it is slower, but provides a standard GP posterior that is better matched to EI and UCB. 
Figure~\ref{fig:comp_BoTorch} shows that, on the marginal expert, this replacement improves the $\pi$BO baselines: the GP-surrogate variants (\legendline{piboUCBGP} $\pi$BO-UCB GP; \legendline{piboEIGP} $\pi$BO-EI GP) outperform their in-context-surrogate counterparts (\legendline{piboUCB} $\pi$BO-UCB; \legendline{piboEI} $\pi$BO-EI).
However, FICBO remains the strongest method in both best-so-far value and cumulative regret, although the gap is reduced. 
This suggests that source-aware pretraining still provides a benefit over classical feedback reweighting.

\subsection{Rollout Analysis}\label{app:rollout}
\begin{figure}[h]
    \centering
    \includegraphics[width=\textwidth]{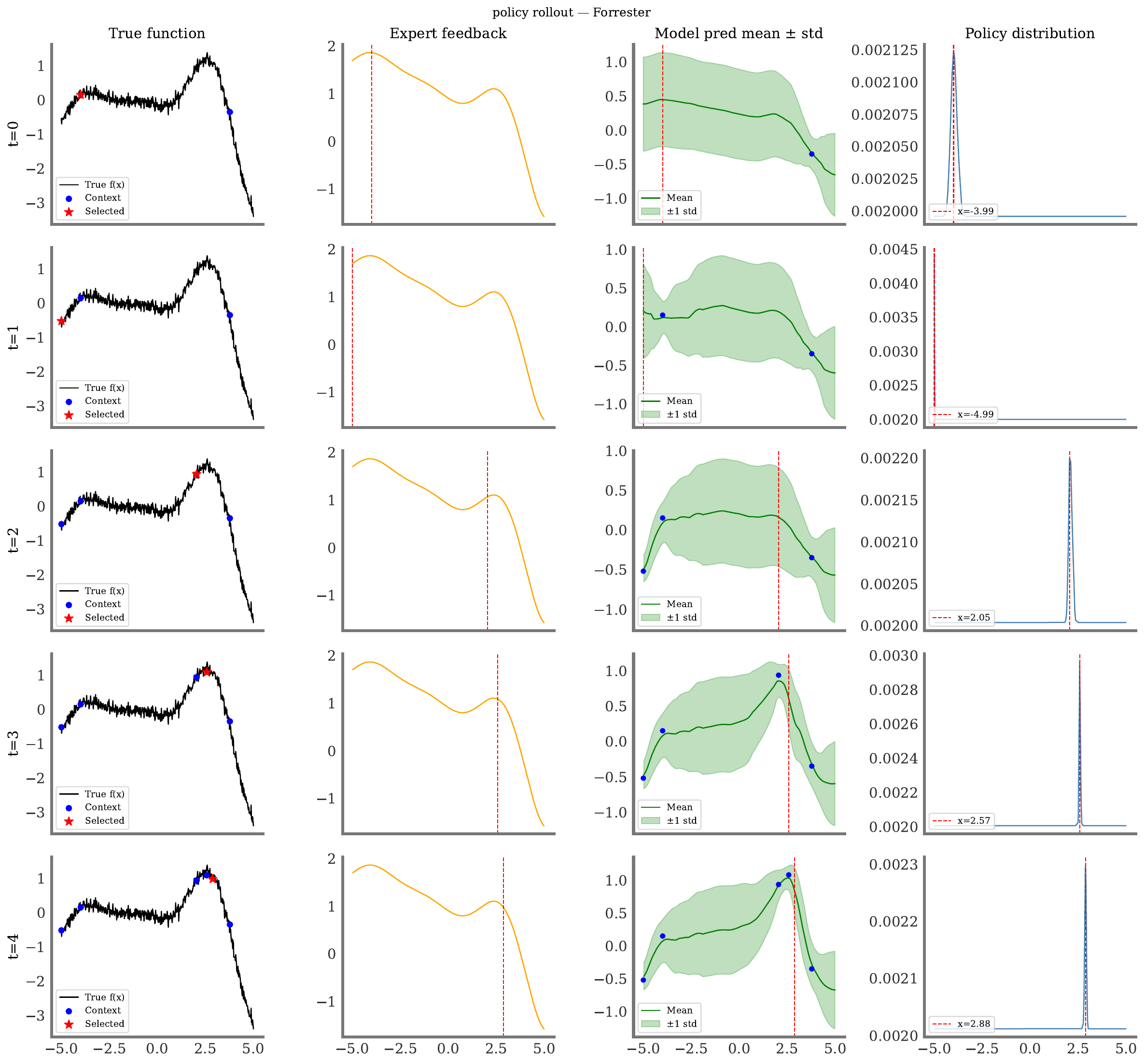}
    \caption{Policy rollout of the amortized policy acquisition function on the Forrester function over $T=5$ steps. 
    Each row corresponds to one acquisition step. 
    From left to right: (1) the true objective function with context points (blue) and the selected candidate (red star), 
    (2) the expert feedback signal over the query set, 
    (3) the model posterior mean (green) with $\pm 1$ standard deviation (shaded), 
    and (4) the learned policy distribution over candidate points. 
    The red dashed vertical line indicates the selected point across all panels.}
    \label{fig:aae_policy_rollout}
\end{figure}

To provide qualitative insight into the behaviour of the learned policy, Figure~\ref{fig:aae_policy_rollout} 
shows a single rollout on the Forrester function, using FICBO (mixture prior). At each acquisition step, the model observes the current context in blue and a set of candidate query points, and must select the most promising candidate. 
The expert feedback signal (second column) provides a biased proxy for the  hidden objective, encoding the expert's preference over the query set. 
In this case, this is the classic low-fidelity variant of the function \citep{forrester2007multi}.
The model posterior (third column) reflects the model's current belief about the objective, which sharpens as more points are observed. 
The policy distribution (fourth column) shows the learned acquisition scores over the query set, which ideally concentrates mass in regions of high objective value.
The model initially agrees with the expert to some extent, but as new information arrives, the posterior is updated.

\section{Generative process of synthetic pretraining dataset}\label{app:synthgen}

\subsection{GP draws}
We construct each synthetic BO task based on a procedure first described by~\cite{huang2025aline}:

\paragraph{Input space and dimensionality.}
We sample the expert input dimension $d_{\mathrm{src}}$ and the number of overlapping dimensions $o$ shared between the main model and the expert; the total input dimension is $d_{\mathrm{full}} = d + d_{\mathrm{src}} - o$. 
The main model observes inputs $\x \in \mathbb{R}^{d}$ (the first $d$ coordinates), while the expert operates over a different and potentially partially overlapping slice of the full $d_{\mathrm{full}}$-dimensional space; the last $d_{\mathrm{src}}$ dimensions.
A model has a fixed number of dimensions; for our experiments, we trained a variant for 1-, 2-, and 3-dimensional inputs.
This decoupling allows us to model realistic settings where the expert has access to features invisible to the main model, and vice versa.

\paragraph{GP hyperparameter prior.}
All GP components, regardless of prior type, share the same hyperparameter prior: 
Matérn family with $\nu \in \{3/2, 5/2, \infty\}$ (each with probability $1/3$), length-scales $\mathcal{U}(0.1\sqrt{d}, 2.0\sqrt{d})$, output scale $\mathcal{U}(0.1, 1.0)$, 
and are isotropic with probability 0.5; in the mixture prior, an isotropic draw gives all real components one shared lengthscale and all decoy components another shared lengthscale.
The expert feedback signal can be further distorted by the structured bias model described in \cref{appendix:desr_of_biases}.
The query, target, and context points are sampled uniformly in $[-5,5]^{d_{\mathrm{full}}}$; the additional expert points are sampled as described in \cref{app:synthgen:additive}. 
These numbers, as well as sections of the code, are based on \cite{huang2025aline}.

\subsection{Additive prior}\label{app:synthgen:additive}

\paragraph{Function generation.}
We draw two independent GPs: a main GP over the main-only dimensions and 
a hidden GP over the expert dimensions. The observed signal is a 
combination of both:
\begin{equation}
    f_\tau = \frac{(1 - w)\, f_\tau^{\mathrm{model}} + w\, f_\tau^{\mathrm{src}}}
    {\sqrt{(1-w)^2 + w^2}},
\end{equation}

where $w_\tau \sim \mathcal{U}(w_{\min}, w_{\max})$ controls the relative contribution of the hidden component, and the denominator normalizes the variance. 
When the main model has no exclusive dimensions ($d = o$), $f_\tau^{\mathrm{model}}$ is set to zero and $w=1$ (only for the 1D model using our hyper-parameters).

The expert's raw signal is $g_\tau$ (or optionally $f_\tau$): the output of the hidden GP evaluated at all points. This signal is optionally perturbed by the structured bias model described in \cref{appendix:desr_of_biases}, yielding a biased feedback signal $\tilde g_\tau$.
Depending on the expert model, the biased signal is further processed. 
For the mixture prior, feedback is returned directly without further changes. 
For the additive prior, we fit a model (amortized GP) to some points generated by the previously constructed function. 
This is to simulate that the additional source may only have information in some area of the design space.
Of these points, 25\% are sampled uniformly and 75\% around 1-5 random cluster centers (std $\mathcal{U}(0.5,1.5)$) in the expert input space, which creates an additional bias.
Its predictive mean at the context, query, and target points is the feedback signal seen by the main model.
The output of the expert model is $u$.
In both cases, the main model only ever observes noisy evaluations of $f_\tau$ as the ground truth signal and $\x_{1:d}$ as inputs, as well as the final feedback $u$. 
The expert's extra observations are never directly available to the main model.

\subsection{Mixture prior}\label{app:synthgen:mixture}
As an alternative that moves the source error from feature access to component weighting, we also consider a latent-component mixture prior. 
It keeps the visibility split: the components are GPs on all $d_{\mathrm{full}}$ coordinates, the optimizer sees the first $d$, and the source signal uses all of them.
For each task, we sample a multi-output Gaussian process
\begin{equation}
    \boldsymbol{z}_\tau(\x)
    =
    \big(z_{\tau,1}(\x),\dots,z_{\tau,K_\tau}(\x)\big),
\end{equation}
whose components represent latent factors underlying the task. The true objective is defined by a scaled sum over a subset of relevant components,
\begin{equation}
    f_\tau(\x)
    =
    \frac{1}{\sqrt{K_{\mathrm{real}}}}
    \sum_{k=1}^{K_{\mathrm{real}}}
    z_{\tau,k}(\x)
\end{equation}
where $\{1,\ldots,K_{\mathrm{real}}\} \subseteq \{1,\dots,K_\tau\}$ indexes the components that actually matter for optimization. Components outside $\{1,\ldots,K_{\mathrm{real}}\}$ act as decoys.

The auxiliary source instead forms its latent signal from a possibly different subset $S_\tau$, constructed by masking each component independently. With $m_{\tau,1}=1$ and $m_{\tau,k}\sim\mathrm{Bernoulli}(p_{\mathrm{src}})$ for $k>1$, the active set is $S_\tau = \{k : m_{\tau,k}=1\}$; equivalently, $|S_\tau|-1 \sim \mathrm{Binomial}(K_\tau-1,\,p_{\mathrm{src}})$. We draw source weights from a symmetric Dirichlet prior on $S_\tau$,
\begin{equation}
    \bar{\boldsymbol{a}}_{\tau,S_\tau}
    \sim
    \mathrm{Dirichlet}\bigl(\alpha_{\mathrm{src}}\mathbf{1}_{|S_\tau|}\bigr),
\end{equation}
and normalize them to unit $\ell_2$ norm, like the weights $1/\sqrt{K_{\mathrm{real}}}$ of $f_\tau$,
\begin{equation} 
    a_{\tau,k}
    =
    \frac{\bar a_{\tau,k}}{\lVert \bar{\boldsymbol a}_{\tau,S_\tau}\rVert_2}\,\mathbb{I}\{k\in S_\tau\}.
\end{equation}
The resulting latent source signal is
\begin{equation}
    g_\tau(\x)
    =
    \sum_{k=1}^{K_\tau}
    a_{\tau,k}\, z_{\tau,k}(\x),
\end{equation}
and the observed feedback again takes the form
\begin{equation}
    u^\tau = \mathcal{B}_\tau[g_\tau].
\end{equation}

This alternative prior treats auxiliary information as a biased combination of latent task factors. Its informativeness depends on the overlap between $S_\tau$ and $\{1,\ldots,K_{\mathrm{real}}\}$, while $\mathcal{B}_\tau$ captures further distortions. This connects naturally to latent-factor and coregionalization views, where different signals share hidden components but weight them differently~\citep{alvarez2012kernels}.

\subsection{Combined Prior}
We do not force a single choice, since both priors have sensible interpretations. 
One can train a single model on a 50/50 mixture of both priors; we call this variant FICBO (Combined).
As an alternative that does not require training an additional model, the resulting policies can be combined by averaging their softmax probabilities, now called FICBO (Ensemble). 
Unsurprisingly, the results of the benchmarks are in between what the individual priors achieve as can be seen in \cref{app:tab:ensembles}, or in more detail in \cref{app:detailed_results}.

\begin{table}[h]
\centering
\caption{Cumulative regret for FICBO combination variants across marginal and XGBoost expert regimes (lower is better)}
\label{app:tab:ensembles}
\begin{tabular}{lcccc}
\toprule
\multirow{2}{*}{Method} & \multicolumn{2}{c}{Marginal expert} & \multicolumn{2}{c}{XGBoost expert} \\
\cmidrule(lr){2-3} \cmidrule(lr){4-5}
 & Mean & Std & Mean & Std \\
\midrule
FICBO (Mixture prior)   & 2.548 & 3.033 & 4.329 & 4.635 \\
FICBO (Ensemble)        & 2.750 & 3.176 & 4.683 & 4.638 \\
FICBO (Combined) (= train on both priors)   & 2.858 & 3.103 & 4.932 & 4.605 \\
FICBO (Additive Prior)  & 3.259 & 3.319 & 5.181 & 4.670 \\
\bottomrule
\end{tabular}
\end{table}

\subsection{Description of Biases}\label{appendix:desr_of_biases}
In our framework, feedback can be perturbed by a set of structured biases. Each bias can be activated independently and is controlled by one or more parameters drawn from predefined priors. Inactive biases are effectively set to zero. 
Biases are applied sequentially to the feedback, and the order may matter for some spatially-dependent transformations. 

\subsubsection{Noise}
\textbf{Description:} Experts may make random, unstructured errors in their feedback across the input space. Homoscedastic noise models this uniform uncertainty.  

\textbf{Implementation:} Add Gaussian noise with constant variance to all feedback points.  

\textbf{Parameters and Priors:}
\begin{itemize}
    \item $\sigma$: Standard deviation of the noise, drawn from a uniform prior over a specified lower and upper bound.
\end{itemize}

\paragraph{Knowledge Center Extension} \mbox{}\\
\textbf{Description:} Instead of the overly simplistic homoscedastic noise alone, we designed a second component. 
Experts and models tend to be reliable near regions they are familiar with, and increasingly uncertain further away. This bias models a single center of competence: feedback remains accurate close to this center, while noise grows with distance from it.

\textbf{Implementation:} Select one random point in the input space as the knowledge center. Compute distances from all feedback points to this center, normalize by the maximum distance, and add Gaussian noise whose standard deviation scales linearly with this normalized distance.

\textbf{Parameters and Priors:}
\begin{itemize}
    \item \texttt{noise\_radiation\_scale}: Maximum noise level at the furthest point from the center, drawn from a uniform prior within specified bounds.
\end{itemize}

For the Mixture prior, the noise we apply throughout training is the combination of both mechanisms and is applied 100\% of the time. For the Additive prior we leave the knowledge center off: its source already only knows part of the space, since it sees a subset of the features and is fit to source-private points that are partly clustered (\cref{app:synthgen:additive}). This produces a similar effect, accurate where the source has data and uncertain elsewhere, so a knowledge center would be redundant.

\subsubsection{Constant Shift}
\textbf{Description:} Experts may exhibit a systematic over- or under-estimation of the true function across the entire input space. 
A constant shift models this global bias in feedback.
\textbf{Implementation:} Add a constant offset to all feedback points.
\textbf{Parameters and Priors:}
\begin{itemize}
    \item $\delta \sim \mathcal{U}(\delta_{\min}, \delta_{\max})$: The shift applied to all feedback points.
\end{itemize}

\subsubsection{Additive GP Bias}
\textbf{Description:} By adding a sampled GP with a smooth kernel to the signal, we create correlated noise. It introduces smooth, structured deviations resembling latent functions that the expert might have learned but misrepresented.  

\textbf{Implementation:} Sample a Gaussian process with RBF kernel using lengthscale and scale parameters. Add the GP output to feedback.  

\textbf{Parameters and Priors:}
\begin{itemize}
    \item lengthscale: Controls smoothness of the GP, sampled uniformly from prior bounds.
    \item scale: Magnitude of GP deviations, drawn uniformly from prior bounds.
    (determines the shape of the fluctuations before normalization. A larger scale will give more contrast between points in the GP sample.)
    \item max\_std: Maximum standard deviation for normalization, drawn uniformly from prior bounds.
    (caps the overall magnitude)
\end{itemize}

\subsubsection{Local Distortion}
\textbf{Description:} Models localized deviations in expert feedback due to over/underestimations in certain regions.  

\textbf{Implementation:} Randomly select $n_\text{centers}$ points, assign a distortion magnitude, and decay this effect with distance from each center.  

\textbf{Parameters and Priors:}
\begin{itemize}
    \item $n_\text{centers}$: Number of distortion centers, sampled uniformly from a specified integer range.
    \item mag: Magnitude of local distortion, drawn from a uniform prior, with a random sign per center.
    \item decay: Spatial decay factor, sampled uniformly from prior bounds.
\end{itemize}

\subsubsection{Catastrophic Failure}
\textbf{Description:} Models rare but severe expert failures where the feedback has no meaningful relationship to the true function. 
For instance, due to a fundamental misunderstanding of the task or a complete breakdown in the expert's model.

\textbf{Implementation:} Replace all feedback values with an independently sampled RBF Gaussian process (lengthscale $\mathcal{U}(0.1,2)$), rescaled to the standard deviation of the signal it replaces.

\textbf{Parameters and Priors:}
\begin{itemize}
    \item No continuous parameters; the bias is either fully active or inactive, indicated by a binary flag.
\end{itemize}

\subsubsection{Application order and activation}
Biases are applied sequentially to the feedback signal, and each bias is activated independently with its own probability for each training batch; all tasks in a batch share the same active biases.
The application order is: homoscedastic noise, catastrophic failure, additive GP bias, local distortion, knowledge-center noise, and finally constant shift.
When a bias is inactive, it contributes no perturbation. 
For more details, we refer to \cref{tab:hp-sketch}.

\section{Experiment Details} \label{app:details_vis}

\subsection{Baselines}%
\label{app:baseline-details}
All baselines are evaluated in the same pool-based setting as our method: at each step $t$, a candidate is selected from a fixed query pool $Q_t^\tau$ ($|Q_0^\tau| = 300$ unless stated otherwise), observed, and added to the context. No gradient-based acquisition optimization is performed; all methods score each pool candidate and select the argmax. For a fair comparison, $Q_t^\tau$ is shared between all baselines.

\paragraph{Surrogate models.}
Methods that score candidates with an acquisition function use one of three surrogates:
\begin{itemize}
    \item \textbf{GP-prior ICBO}: the same pretrained transformer model used by our method, except for the feedback encoder, producing a GMM predictive posterior over each query point.
    The model has been trained on generic GP data following \citep{huang2025aline}.
    When run on its own, GP-prior ICBO selects candidates with its policy head.
    \item \textbf{FICBO inference head}: the GMM predictive posterior $q_\phi$ of the trained FICBO model (Mixture Prior), which also conditions on the feedback (semi-amortized mode, \cref{app:algs}).
    \item \textbf{Classical GP}: a batched GPyTorch exact GP with ARD Matérn-3/2 kernel, fit with 100 Adam steps per BO step, warm-started from the previous step.
\end{itemize}
For both GMM posteriors, the weighted mean and variance are extracted as $\mu(\x) = \sum_\ell \omega_\ell \mu_\ell(\x)$ and $\sigma^2(\x) = \sum_\ell \omega_\ell \sigma_\ell^2(\x)$.
The main-text $\pi$BO baselines use the GP-prior ICBO surrogate. 
This is done to isolate the effect of the feedback integration strategy from the surrogate quality.

\paragraph{Acquisition functions.}
We use the following closed-form acquisition functions, which can be combined with any surrogate:
\begin{itemize}
    \item \textbf{UCB}: $\alpha(\x) = \mu(\x) + \kappa \sigma(\x)$. $\kappa$ has been set to 1.
    \item \textbf{EI}: $\alpha(\x) = (\mu(\x) - y^+)\Xi(z) + \sigma(\x)\xi(z)$ where $z = (\mu(\x) - y^+)/\sigma(\x)$, $\Xi$ and $\xi$ the standard normal CDF and PDF, and $y^+$ the best observed value.
\end{itemize}
\paragraph{Semi-amortized FICBO.}
One can also replace the policy head of FICBO with its inference head and an acquisition function (\cref{app:algs}). We use UCB and EI, reweighted with the $\pi$BO feedback method, described below, and call these variants FICBO-UCB and FICBO-EI.
\cref{app:tab:regret_combined_acquOurPrior} shows the results are comparable to the end-to-end policy with the same prior.
For a more detailed comparison, see \cref{app:detailed_results}.

\begin{table}[h]
\centering
\caption{Cumulative regret across marginal and XGBoost expert regimes (lower is better), comparing end-to-end and semi-amortized FICBO with the baselines.}
\label{app:tab:regret_combined_acquOurPrior}
\begin{tabular}{lcccc}
\toprule
\multirow{2}{*}{Method} & \multicolumn{2}{c}{Marginal expert} & \multicolumn{2}{c}{XGBoost expert} \\
\cmidrule(lr){2-3} \cmidrule(lr){4-5}
 & Mean & Std & Mean & Std \\
\midrule
FICBO (Mixture Prior), end-to-end policy & 2.548 & 3.033 & 4.329 & 4.635 \\
FICBO-UCB (Mixture Prior), semi-amortized       & 2.657 & 3.204 & 4.897 & 4.613 \\
FICBO-EI (Mixture Prior), semi-amortized      & 2.771 & 3.388 & 4.880 & 4.688 \\
FICBO (Additive Prior), end-to-end policy & 3.259 & 3.319 & 5.181 & 4.670 \\
Feedback                                 & 3.683 & 4.392 & 7.774 & 6.178 \\
$\pi$BO-UCB                             & 4.966 & 4.845 & 6.199 & 5.355 \\
$\pi$BO-EI                              & 5.466 & 4.988 & 6.477 & 5.513 \\
GP-prior ICBO + feedback feature         & 6.000 & 4.820 & 7.129 & 5.524 \\
GP-prior ICBO                            & 8.462 & 5.939 & 8.462 & 5.939 \\
Random                                   & 12.158 & 6.753 & 12.879 & 6.836 \\
\bottomrule
\end{tabular}
\end{table}

\paragraph{Incorporating expert feedback.}
For baselines that augment the acquisition with feedback ($\pi$BO-style), the feedback signal scores are first normalized before combination. Concretely, the feedback is converted to a policy via a softmax function: $\tilde{u}(\x) = \text{softmax}(u(\x_1), \ldots, u(\x_{|Q_t|}))$
The adjusted score is then:
\begin{equation}
    \alpha_{\text{adj}}(\x) = \alpha(\x) \cdot \tilde{u}(\x)^{\beta / (t+1)}
\end{equation}
where the decaying exponent $\beta/(t+1)$ follows \citet{hvarfner2022pibo}, ensuring the feedback influence vanishes as more objective observations accumulate and the surrogate becomes more reliable. We set $\beta=2.0$ following the original paper. %

\paragraph{GP-prior ICBO with Feedback Feature.}
The concatenation baseline appends the feedback value $u(\x)$ directly to the input features before passing to the amortized model. The model receives $(\x, u(\x))$ as input rather than $\x$ alone. This tests whether the model can implicitly learn to use feedback when it is treated as just another input feature, without any explicit feedback-aware architecture.

\paragraph{Classical-GP controls.}
GP-UCB uses the classical GP surrogate with UCB and no feedback. GP-UCB + feedback feature fits the same GP on $(\x, u(\x))$ instead of $\x$. They are the classical-GP counterparts of GP-prior ICBO and GP-prior ICBO + feedback feature, and show how much of the performance comes from the surrogate and how much from the feedback.

\subsubsection{Additional Baselines} \label{app:additional_Baselines}
\paragraph{Notation.}
For the three additional baselines below we treat the pool-wide expert feedback as a second \emph{information source} $u(\x)$ that is informative about, but distinct from, the sparse real objective $f(\x)$. We reuse $\mu_i(\x), \sigma_i^2(\x)$ for the mean/variance of the $i$-th component model, $Q_t^\tau$ for the pool at step $t$, and index sources with $j \in \{f, u\}$ (real objective and feedback, respectively).

\paragraph{iMFBO~\citep{fan2024multi}.}
Multi-fidelity BO with multiple information sources treats every source as a noisy view of one shared latent objective. Our feedback is not on the scale of the objective, thus we add an affine calibration per environment:
\begin{equation}
    y(\x) = f(\x) + \sigma_f\,\varepsilon, \qquad u(\x) = a\,f(\x) + b + \sigma_u(\x)\,\varepsilon, \qquad \varepsilon \sim \mathcal{N}(0,1),\ a > 0,
\end{equation}
so that the real objective and the feedback are jointly modeled by a single GP over $f$ whose noise diagonal is input- and source-dependent:
\begin{equation}
    \hat{K} = S K S + \mathrm{diag}\big(\sigma_{j_i}(\x_i)^2\big), \qquad S = \mathrm{diag}(s_i),\ s_i = 1 \text{ for objective rows},\ s_i = a \text{ for feedback rows}.
\end{equation}
The objective noise $\sigma_f$ is a learned constant. 
The fidelity function $\sigma_u(\x)$, i.e. how much noise the feedback row contributes at $\x$, is learned jointly with $a$, $b$ and the GP hyperparameters by MAP. 
iMFBO parameterizes $\sigma_u(\x)$ with a small MLP. iMFBO-GP uses one free log-noise value per feedback row, with a GP prior over these values (the MAP estimate of \citet{fan2024multi}, Eq.~12).
Candidates are scored with NVUCB, $\alpha(\x) = \mu(\x) + \kappa\,\gamma(\x)\,\sigma(\x)$ with $\gamma(\x) = \sigma(\x)/\sqrt{\sigma(\x)^2 + \sigma_f^2}$, which decreases the exploration bonus where a query is expected to give little information. 
The iMFBO-UCB and iMFBO-GP-UCB variants use plain UCB on the same model. In all variants, $\kappa = 1$.

\paragraph{RGPE~\citep{feurer2018scalable}.}
The rank-weighted GP ensemble treats the dense pool-wide feedback as a \emph{related} (base) task and the sparse real-objective observations as the \emph{target} task, and combines a GP fit on each: $\mathrm{GP}_u$ trained on all $(\x, u(\x))$ pairs in $Q_t^\tau$, and $\mathrm{GP}_f$ trained on the observed $(\x, f(\x))$ pairs so far. The ensemble posterior at a candidate $\x$ is a weighted combination
\begin{equation}
    \mu_{\text{RGPE}}(\x) = \sum_{i \in \{u, f\}} w_i\, \mu_i(\x), \qquad
    \sigma_{\text{RGPE}}^2(\x) = \sum_{i \in \{u, f\}} w_i^2\, \sigma_i^2(\x),
\end{equation}
where the weights $w_i$ are estimated from a leave-one-out ranking-loss comparison between each base model's predicted pairwise ordering on the target task and the true observed ordering, so that $\mathrm{GP}_u$ is down-weighted automatically as it disagrees with the accumulating real-objective observations. 
We use $\mu_{\text{RGPE}}, \sigma_{\text{RGPE}}$ in place of $\mu, \sigma$ in the UCB/EI acquisition functions defined above.

\paragraph{Multi-Task GP (ICM).}
As a transfer-learning baseline, we fit a single multi-task GP jointly over the real-objective task and the feedback task using a batched re-implementation of BoTorch's \texttt{MultiTaskGP} with one added free outputscale, with an intrinsic co-regionalization (ICM) kernel
\begin{equation}
    K\big((\x, j), (\x', j')\big) = K_{\x}(\x, \x') \cdot B_{j,j'}, \qquad j, j' \in \{f, u\},
\end{equation}
where $K_{\x}$ is a scaled RBF kernel with BoTorch's dimension-scaled lengthscale prior and $B \in \mathbb{R}^{2\times 2}$ is a learned full-rank, non-negative task covariance matrix coupling the real objective and the feedback. 
All available feedback in $Q_t^\tau$ and the real-objective observations collected so far are used to fit $B$ and the GP hyperparameters jointly, refit at every step $t$ with 30 Adam steps, warm-started from the hyperparameters of the previous step. 
Candidates are scored via UCB on the task-$f$ posterior slice, $\alpha(\x) = \mu_f(\x) + \kappa\, \sigma_f(\x)$, with $\kappa = 1$ as above.

All mentioned baselines get outperformed by FICBO and are considerably more computationally expensive. 
RGPE and iMFBO-GP follow (iMFBO-GP is better on the marginal expert, RGPE on the XGBoost expert), then iMFBO and the multi-task GP (\cref{app:tab:extra_baselines}). 
Detailed results can be found in \cref{app:detailed_results}.
\begin{table}[h]
\centering
\caption{Final cumulative regret across marginal and XGBoost expert regimes (lower is better) for additional baselines}
\label{app:tab:extra_baselines}
\begin{tabular}{lcccc}
\toprule
\multirow{2}{*}{Method} & \multicolumn{2}{c}{Marginal} & \multicolumn{2}{c}{XGBoost} \\
\cmidrule(lr){2-3} \cmidrule(lr){4-5}
 & Mean & Std & Mean & Std \\
\midrule
FICBO (Mixture Prior)                        & 2.548  & 3.033 & 4.329  & 4.635 \\
FICBO (Additive Prior)                       & 3.259  & 3.319 & 5.181  & 4.670 \\
Feedback                                     & 3.683  & 4.392 & 7.774  & 6.178 \\
RGPE-UCB                                     & 5.743  & 5.354 & 6.849  & 5.670 \\
GP-prior ICBO + feedback feature       & 6.000  & 4.820 & 7.129  & 5.524 \\
iMFBO-NVUCB                                  & 7.203  & 5.894 & 8.628  & 6.142 \\
iMFBO-GP-NVUCB                               & 4.840  & 4.414    & 8.200  & 6.153   \\
iMFBO-UCB                                    & 7.452  & 5.998 & 8.767  & 6.115 \\
iMFBO-GP-UCB                                 & 5.090  & 4.461    & 8.330  & 	6.193    \\ 
MTGP-UCB (ICM)                               & 8.007  & 6.026 & 9.104  & 6.373 \\
GP-prior ICBO                             & 8.462  & 5.939 & 8.462  & 5.939 \\
Random                                       & 12.158 & 6.753 & 12.879 & 6.836 \\
\bottomrule
\end{tabular}
\end{table}

\subsection{Benchmark functions}\label{sec:benchmarkfunctions}
\paragraph{Sampling.}
We sample $n$ input points $\x \in \mathbb{R}^{d_{\mathrm{full}}}$ uniformly in the normalized space $[-s, s]^{d_{\mathrm{full}}}$, where $s$ is a fixed design scale (set to 5 to match the training distribution used). Each point is mapped to the natural domain of the chosen function before evaluation, and independent Gaussian noise $\varepsilon \sim \mathcal{N}(0, \sigma^2)$ is added to the output. The dataset is split into context, query, and target sets.

\paragraph{Expert feedback.}
For functions with a registered low-fidelity variant, the expert signal $u$ is the output of that variant evaluated at the same inputs, without further noise. For functions without a low-fidelity variant, $u$ is the Monte Carlo marginal of $f$ over the dimensions not observed by the expert (the expert sees only the last dimension, $d_{\mathrm{src}}=1$), %
\begin{equation}
    u(\x_{\mathrm{src}}) \approx \frac{1}{L_{\mathrm{MC}}}\sum_{l=1}^{L_{\mathrm{MC}}} f\!\left(\x_{\mathrm{model}}^{(l)},\, \x_{\mathrm{src}}\right), \quad \x_{\mathrm{model}}^{(l)} \sim \mathcal{U}([-s, s]^{d_{\mathrm{full}}-d_{\mathrm{src}}}),
\end{equation}
with $L_{\mathrm{MC}} = 2000$ samples over the non-expert dimensions (the point's own coordinates plus 1999 uniform draws shared across points). This is to synthetically make the feedback independent of the dimensions that the expert has no knowledge of.
The main model observes noisy evaluations of $f$, $u$, and the main features (similar to the training procedure \cref{app:synthgen}) only; the expert's inputs are never directly available.

\paragraph{XGBoost bias}
To simulate a systematically biased expert, we fit a shallow gradient-boosted tree (XGBoost, 10 estimators, max depth 3, learning rate 0.15, subsample and column sample 0.8) to the noise-free objective values at available context and expert-only points (50 uniformly sampled points), using only the $d_{\mathrm{src}}$ dimensions visible to the expert.
The model is trained once in the beginning, and its predictions serve as $u$ at all context, query, and target locations.
Because the surrogate is deliberately under-specified (low capacity, partial inputs), it produces a \emph{biased} but correlated signal, mimicking a domain expert whose knowledge is both limited and biased (via the limited model class).

\begin{figure}[t]
    \centering
    \begin{subfigure}[b]{0.02\textwidth}
    \end{subfigure}%
    \begin{subfigure}[b]{0.46\textwidth}
        \includegraphics[width=\textwidth]{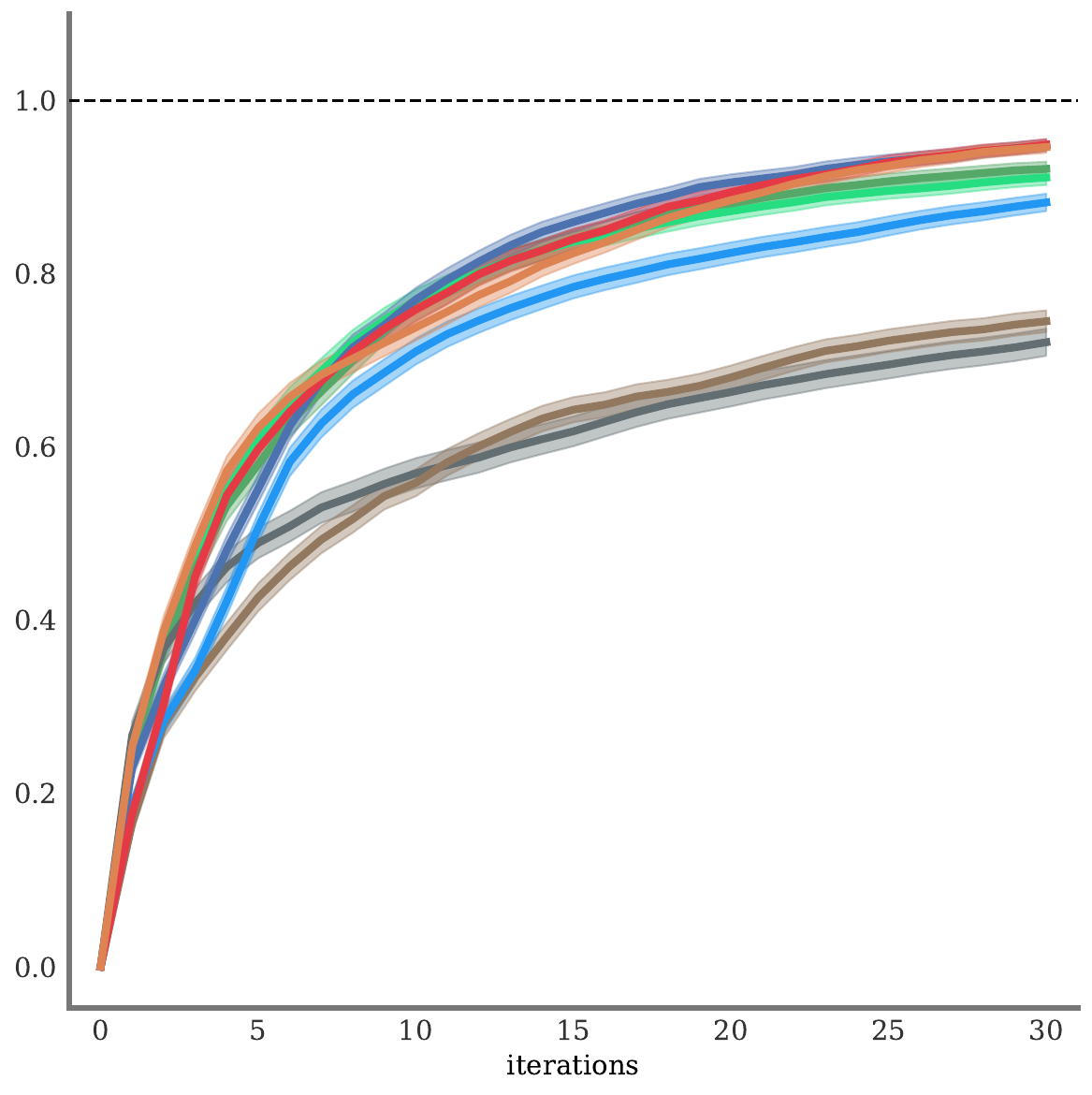}
        \caption{Best found ($\uparrow$)}
    \end{subfigure}%
    \hspace{0.02\textwidth}%
    \begin{subfigure}[b]{0.46\textwidth}
        \includegraphics[width=\textwidth]{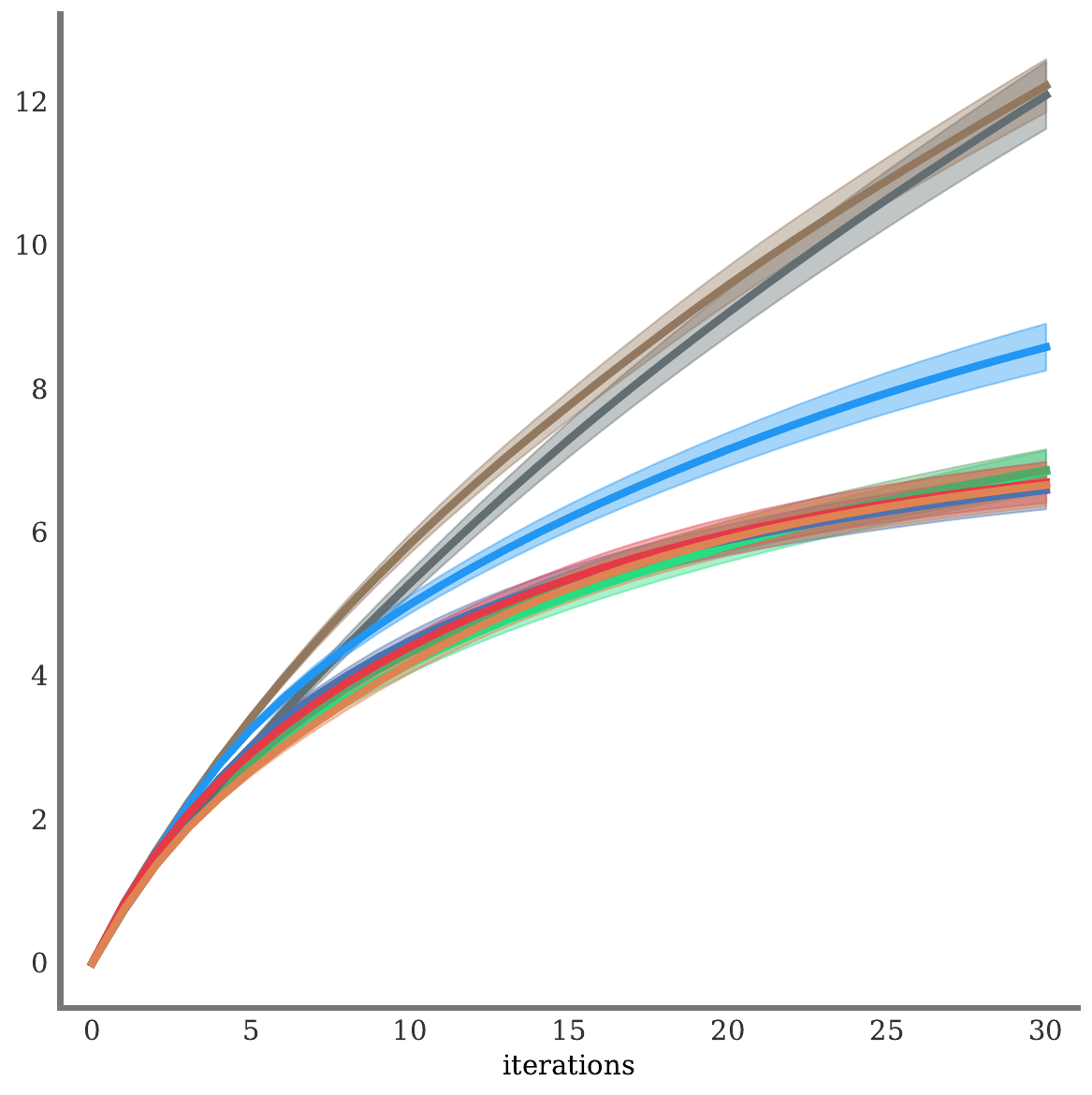}
        \caption{Cumulative regret ($\downarrow$)}
    \end{subfigure}%
    \vspace{0.3em}
    \begin{subfigure}[b]{\textwidth}
        \centering
        \includegraphics[
            width=0.99\textwidth,
            trim=0 0 0 10pt,
            clip
        ]{img/legends/Legend.pdf}
    \end{subfigure}
    \caption{
        \textbf{Marginalized Feedback with Fake peak.}
        \textbf{Left:} normalized best value found.
        \textbf{Right:} normalized cumulative regret.
        Mean $\pm$ 95\% bootstrap confidence intervals.
    }
    \label{app:fig:fake_peak}
\end{figure}

\paragraph{Fake peak}
To test whether the model blindly follows the expert, we corrupt the expert signal with a single fake optimum. 
Per task instance we sample one center $\mathbf{c}$ uniformly in the expert-visible dimensions, 
$\mathbf{c} \sim \mathcal{U}([-s, s]^{d_{\mathrm{src}}})$, and add an exponential bump to the marginalized feedback,
\begin{equation}
    \tilde{u}(\mathbf{x}_{\mathrm{src}}) = u(\mathbf{x}_{\mathrm{src}}) + 1.1\, y^{\star} \exp\!\big(-\lambda \,\lVert \mathbf{x}_{\mathrm{src}} - \mathbf{c} \rVert_2\big),
\end{equation}
where $y^{\star}$ is, per task instance, the largest observed (noisy, normalized) objective value over the sampled context, query, and target points, and $\lambda = 0.5$ is fixed.
The bump height at $\mathbf{c}$ is $1.1\,y^{\star}$, so the expert's apparent best region is usually well above the real optimum. 
Since $\mathbf{c}$ depends only on the expert dimensions and is independent of $f$, the peak carries no information about the true optimum. 
A model that trusts the feedback uncritically will be pulled to $\mathbf{c}$, whereas one that has learned to cross-check $u$ against observed $f_\tau$ values should discount it.
Results are shown in \cref{app:fig:fake_peak}. 
In this example FICBO and $\pi BO$ variants have very similar results, with FICBO reaching slightly better final solutions.

\subsubsection{Function descriptions}
 
The following functions are included in the experiments. 
All functions below are shown in their standard minimisation form. Every objective is then standardized with 10,000 uniform samples.
In the implementation, all are negated so that larger values are better.
The functions are commonly used for BO, BED, or AL benchmarking \citep{huang2025aline}.

\paragraph{Ackley ($d \geq 1$).}
A multimodal function with a single global optimum surrounded by a nearly flat outer region and many local optima,

\begin{equation}
    f(\x) = -20\exp\!\left(-0.2\sqrt{\tfrac{1}{d}\textstyle\sum_i x_i^2}\right) - \exp\!\left(\tfrac{1}{d}\textstyle\sum_i \cos 2\pi x_i\right) + 20 + e,
\end{equation}
defined over $[-5, 5]^d$.

\paragraph{Rastrigin ($d \geq 1$).}
A highly multimodal function with a large number of regularly spaced local optima,
\begin{equation}
    f(\x) = 10d + \sum_{i=1}^d \left(x_i^2 - 10\cos 2\pi x_i\right),
\end{equation}
defined over $[-5.12, 5.12]^d$.

\paragraph{Levy ($d \geq 1$).}
A multimodal function defined via the transformation $w_i = 1 + (x_i - 1)/4$,
\begin{equation}
    f(\x) = \sin^2(\pi w_1) + \sum_{i=1}^{d-1}(w_i-1)^2\bigl[1 + 10\sin^2(\pi w_i + 1)\bigr] + (w_d - 1)^2\bigl[1 + \sin^2(2\pi w_d)\bigr],
\end{equation}
over $[-10, 10]^d$.

\paragraph{Sphere ($d \geq 1$).}
The simplest unimodal baseline $f(\x) = \sum_i x_i^2$, over $[-5.12, 5.12]^d$.

\paragraph{Schwefel ($d \geq 1$).}
A deceptive function whose global optimum is geometrically distant from the next-best local optima,
\begin{equation}
    f(\x) = 418.9829\,d - \sum_{i=1}^d x_i \sin\!\left(\sqrt{|x_i|}\right),
\end{equation}
over $[-500, 500]^d$.

\paragraph{Rosenbrock ($d \geq 2$).}
A classic unimodal-but-narrow-valley function whose global minimum lies inside a long, curved, flat valley, making convergence hard even though the function is unimodal,
\begin{equation}
    f(\x) = \frac{1}{3}\left[\log\!\left(1 + \sum_{i=1}^{d-1}\left(100(x_{i+1}-x_i^2)^2 + (x_i-1)^2\right)\right) - 3\right],
\end{equation}
defined over $[-2, 2]^d$.\footnote{We apply a $\log(1+\cdot)$ compression and affine rescaling to the raw Rosenbrock sum for numerical stability and to bring the output range in line with the other benchmark functions, since the raw sum otherwise grows extremely large in magnitude.}

\paragraph{Hartmann ($d = 3$).}
A smooth multimodal function with four local minima, of which one is global, commonly used as a moderate-dimensional BO benchmark,
\begin{equation}
    f(\x) = -\frac{1}{4}\sum_{i=1}^4 \alpha_i \exp\!\left(-\sum_{j=1}^3 A_{ij}(x_j - P_{ij})^2\right),
\end{equation}
with $\boldsymbol\alpha = (1.0, 1.2, 3.0, 3.2)$ and coefficient matrices
\begin{equation}
    A = \begin{pmatrix} 3.0 & 10.0 & 30.0 \\ 0.1 & 10.0 & 35.0 \\ 3.0 & 10.0 & 30.0 \\ 0.1 & 10.0 & 35.0 \end{pmatrix},
    \qquad
    P = \begin{pmatrix} 0.3689 & 0.1170 & 0.2673 \\ 0.4699 & 0.4387 & 0.7470 \\ 0.1091 & 0.8732 & 0.5547 \\ 0.03815 & 0.5743 & 0.8828 \end{pmatrix},
\end{equation}
defined over $[0,1]^3$, with global minimum $f(\x^*)\approx -0.966$ at $\x^*\approx(0.1146, 0.5556, 0.8526)$.\footnote{We use a $1/4$-scaled version of the canonical Hartmann-3 function (whose unscaled global minimum is $\approx -3.863$) to keep its output range comparable to the other benchmark functions. As with the other functions in this list, this minimization form is negated in the implementation to produce the maximization objective used in experiments.}

\subsection{Branin multi-fidelity function  ($d=2$)}\label{sec:branin}

A standard two-dimensional test function; with the input mapping below, two of its three global minima lie in the domain.
\begin{equation}
f_h(\x) = \left(x_2' - \frac{5.1}{4\pi^2}x_1'^2 + \frac{5}{\pi}x_1' - 6\right)^2 + 10\left(1 - \frac{1}{8\pi}\right)\cos(x_1') + 10
\end{equation}
where $x_i' = 15x_i$, over $[0,1]^2$. The objective is standardized to zero mean and unit variance over the domain (estimated from 10{,}000 uniform samples). 
We use the correlation-adjustable low-fidelity variant of~\citep{toal2015some}:
\begin{equation}
f_l(\x) = f_h(\x) - (\rho + 0.5)\left(x_2' - \frac{5.1}{4\pi^2}x_1'^2 + \frac{5}{\pi}x_1' - 6\right)^2
\end{equation}
where $\rho \in [0,1]$ controls the HF--LF (High Fidelity--Low Fidelity) correlation: the squared term enters $f_l$ with weight $0.5-\rho$; it is halved at $\rho=0$, removed at $\rho=0.5$ (then $f_l$ depends on $x_1$ only), and inverted for $\rho>0.5$, which moves the low-fidelity optimum away from that of $f_h$. The feedback is standardized with mean 0.4484 and std 13.52 (the values for $\rho=0.75$) for every $\rho$.

\begin{figure}[H]
    \centering
    \includegraphics[width=1\linewidth]{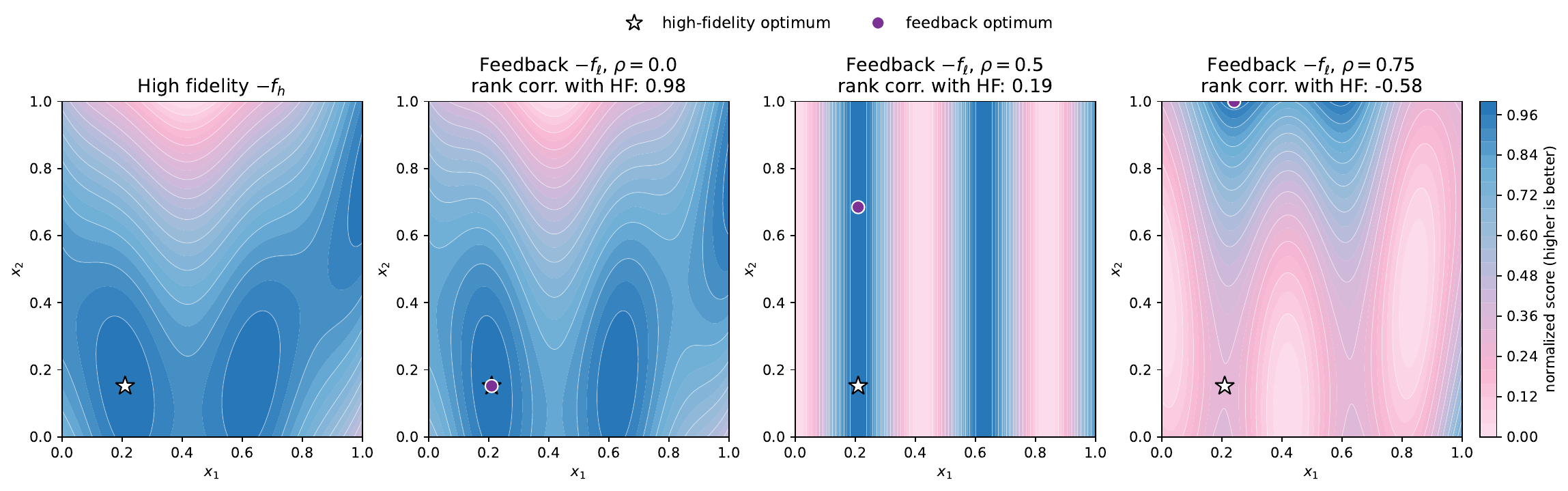}
\caption{
\textbf{Correlation-adjustable Branin feedback surfaces.}
We plot normalized maximization scores for the high-fidelity objective $-f_h$ and the low-fidelity feedback $-f_l$. White stars mark the high-fidelity optimum, and purple circles mark the feedback optimum. As the discrepancy parameter $\rho$ increases, the feedback surface becomes less rank-aligned with the true objective, as measured by Spearman rank correlation.
}
\label{fig:branin}
\end{figure}

\subsection{Sequential Reaction with Auxiliary Signal}\label{sec:reaction}

We consider a standard sequential reaction system:
\[
A \xrightarrow{k_1} B \xrightarrow{k_2} C,
\]
a classical setting in chemical reaction engineering for studying the trade-off between conversion and selectivity (e.g., \cite{fogler1999elements, levenspiel1998chemical}). The objective is the concentration of the intermediate $B$, which we denote as $[B](t_{\mathrm{res}})$, as a function of residence time $t_{\mathrm{res}}$ and temperature $T_{\mathrm{rx}}$. 
The reaction follows first-order kinetics with Arrhenius temperature dependence:
\begin{align}
k_1(T_{\mathrm{rx}}) &= A_1 \exp\left(-\frac{E_{a1}}{R T_{\mathrm{rx}}}\right), \\
k_2(T_{\mathrm{rx}}) &= A_2 \exp\left(-\frac{E_{a2}}{R T_{\mathrm{rx}}}\right),
\end{align}
where $A_i$ are pre-exponential factors, $E_{ai}$ activation energies, and $R$ the universal gas constant. We use $A_1=10^{10}$, $E_{a1}=60\,\mathrm{kJ\,mol^{-1}}$, $A_2=10^{13}$, $E_{a2}=80\,\mathrm{kJ\,mol^{-1}}$, $R=8.314\,\mathrm{J\,mol^{-1}\,K^{-1}}$, $[A]_0=1$, $T_{\mathrm{rx}}\in[300,600]\,\mathrm{K}$ and $t_{\mathrm{res}}\in[0.01,2]$.

The concentration of $B$ can be obtained analytically as:
\begin{equation}
[B](t_{\mathrm{res}}) = [A]_0 \frac{k_1}{k_2 - k_1} \left(e^{-k_1 t_{\mathrm{res}}} - e^{-k_2 t_{\mathrm{res}}}\right),
\end{equation}
which is physically bounded in $[0,1]$ when normalized by the initial reactant concentration $[A]_0$.

In addition, we use an auxiliary feedback signal corresponding to the conversion of $A$:
\begin{equation}
\text{Conversion} = 1 - e^{-k_1 t_{\mathrm{res}}},
\end{equation}
which is a standard, easily measured quantity in chemical reaction engineering. In sequential reactions, however, maximizing conversion is not aligned with maximizing $[B]$, since higher conversion drives the system toward the final product $C$. This induces a systematic mismatch between the proxy signal and the true objective.
While Bayesian optimization has been applied to reaction yield optimization \cite{shields2021bayesian}, the use of conversion as a structured auxiliary signal in this setting and its systematic bias relative to the true objective has not been explicitly studied.

\paragraph{Objective and auxiliary signal.}
Let $\x=(T_{\mathrm{rx}}, t_{\mathrm{res}})$ denote temperature and residence time. The expensive objective is the yield of the intermediate product $B$,
\begin{equation}
    f(\x) = [B](t_{\mathrm{res}},T_{\mathrm{rx}})
    =
    [A]_0
    \frac{k_1(T_{\mathrm{rx}})}{k_2(T_{\mathrm{rx}})-k_1(T_{\mathrm{rx}})}
    \left(
        e^{-k_1(T_{\mathrm{rx}})t_{\mathrm{res}}} - e^{-k_2(T_{\mathrm{rx}})t_{\mathrm{res}}}
    \right),
\end{equation}
whereas the auxiliary feedback is the conversion of reactant $A$,
\begin{equation}
    u(\x) = 1 - e^{-k_1(T_{\mathrm{rx}})t_{\mathrm{res}}}.
\end{equation}
Both $f$ and $u$ are clipped to $[0,1]$. 
The objective is standardized to zero mean and unit variance over the domain, while the conversion feedback is used unscaled and without noise. 
This proxy is cheap to measure and operationally meaningful, but only partially aligned with the objective: consumed $A$ first forms the desired intermediate $B$, which then reacts further to the undesired by-product $C$. Consequently, $u(\x)$ is monotone in $t_{\mathrm{res}}$ for fixed $T_{\mathrm{rx}}$, while $f(\x)$ typically attains an interior optimum because $B$ is eventually depleted into $C$.

\subsection{Aerodynamics}\label{sec:aero}
We consider the problem of maximizing the lift-to-drag ratio $C_L/C_D$ of a NACA 4-digit airfoil (maximum camber at 40\% chord; thickness and camber rounded to whole percent) as a function of angle of attack $\alpha \in [-2°, 15°]$, thickness $t \in [0.08, 0.20]$, and camber $c \in [0.00, 0.06]$, a standard aerodynamic shape optimization setting \cite{forrester2007multi}.

The true objective is evaluated using NeuralFoil \cite{sharpe2025neuralfoil}, a neural network surrogate trained on approximately one million XFOIL runs %
 that captures viscous boundary-layer effects, including flow separation and stall. At moderate $\alpha$, $C_D$ remains low in a laminar drag bucket; as $\alpha$ increases toward stall, boundary-layer separation causes $C_D$ to grow rapidly, so $C_L/C_D$ peaks at an interior point ($\alpha^* \approx 5$--$7°$) before declining.

As auxiliary feedback, we consider two proxy variants. The first is structurally biased: it uses lift
alone and therefore ignores the drag penalty that makes the true lift-to-drag ratio peak before stall.
The second is a lower-accuracy neural surrogate for the same lift-to-drag objective, and therefore
serves as a noisy but well-aligned control.

\paragraph{Misleading Feedback Variant (Airfoil Lift)} uses $C_L$ alone (from NeuralFoil) as the feedback signal. Since $C_L$ is the numerator of the true objective and increases monotonically with $\alpha$ until hard stall, an agent relying on it will systematically overshoot toward higher angles of attack, ignoring the drag penalty entirely.

\paragraph{Aligned Feedback Variant (Airfoil Small)} feedback is generated by a smaller NeuralFoil model (model size \texttt{small} instead of \texttt{large}), computing $C_L/C_D$ from the same viscous surrogate but at reduced accuracy. Unlike the misleading variant above, this proxy shares the same physical structure as the truth, so the optimum location should be approximately the same as the ground truth. This variant therefore represents \emph{noisy} feedback rather than \emph{structurally biased} feedback, and serves as a control.

Pairing a cheap and an accurate solver as low- and high-fidelity levels is common practice in aerodynamic optimization, for example inviscid panel codes with viscous RANS solvers \cite{forrester2007multi, priyanka2021multi}. Here, both levels are viscous NeuralFoil models of different size.
Airfoil Lift has a wrong optimum because it ignores the drag rise that causes $C_L/C_D$ to turn over. 
Meanwhile, Airfoil Small uses a smaller model, which leads to slightly worse accuracy globally.

\paragraph{Objective and auxiliary signal}
Let $\x = (\alpha,t,c)$ denote angle of attack, thickness, and camber. The expensive objective is the viscous lift-to-drag ratio
\begin{equation}
    f(\x)
    =
    \frac{C_L^{\mathrm{NF,large}}(\x)}{C_D^{\mathrm{NF,large}}(\x)},
\end{equation}
where $(C_L^{\mathrm{NF,large}}, C_D^{\mathrm{NF,large}})$ are given by the large NeuralFoil model. The auxiliary feedback is instead provided by a cheaper or structurally simplified proxy,
\begin{equation}
    u(\x) \in \{u_A(\x),u_B(\x)\},
\end{equation}
with
\begin{align}
    u_A(\x)
    &= C_L^{\mathrm{NF,large}}(\x),\\
    u_B(\x)
    &= \frac{C_L^{\mathrm{NF,small}}(\x)}{C_D^{\mathrm{NF,small}}(\x)}.
\end{align}
The key mismatch for $u_A$ is that $f(\x)$ has an interior optimum due to drag growth and stall,
whereas lift alone tends to favor larger angles of attack. In contrast, $u_B$ shares the same
lift-to-drag structure as $f$ but is evaluated with a smaller NeuralFoil model, making it a cheaper,
noisier proxy rather than a structurally biased one.

We use NeuralFoil~\cite{sharpe2025neuralfoil}, a physics-informed 
machine learning tool for airfoil aerodynamics analysis, available 
at \url{https://github.com/peterdsharpe/NeuralFoil} under the 
MIT License.

\subsection{Neural Architecture Search (NAS)}\label{sec:nas}
We work on the problem of selecting the optimal cell architecture in
NAS-Bench-201~\cite{dong2020bench}, a tabular benchmark that enumerates all neural architectures definable on a fixed directed acyclic graph with six edges, each assigned one of five candidate operations:
\texttt{none}, \texttt{skip\_connect}, \texttt{nor\_conv\_1x1},
\texttt{nor\_conv\_3x3}, and \texttt{avg\_pool\_3x3}.
We fix three edges to canonical values (edges 0, 3, 5 to
\texttt{nor\_conv\_3x3}, \texttt{skip\_connect}, and
\texttt{nor\_conv\_3x3}, respectively, a well-performing choice) to reduce the dimension, and treat the remaining three
free edges as the design variables, which results in a discrete search space of $5^3 = 125$ architectures, resulting in a reduced query set size of 123, as well as 1 target and initial context point.

\paragraph{Objective and auxiliary signal.}
Let $\x = (e_1, e_2, e_4) \in \{0,\ldots,4\}^3$ index the operation assigned to each free edge.
The expensive objective is the test accuracy on ImageNet16-120~\cite{dong2020bench}, a downsampled variant of ImageNet that is substantially harder than CIFAR-scale tasks and requires architectures with higher representational capacity:
\begin{equation}
    f(\x) = \mathrm{Acc}_{\mathrm{ImageNet16\text{-}120}}(\x).
\end{equation}
The auxiliary feedback is the test accuracy on CIFAR-10, which is cheaper to evaluate and widely used as a transferability signal in NAS~\cite{yu2019evaluating, real2019regularized}:
\begin{equation}
    u(\x) = \mathrm{Acc}_{\mathrm{CIFAR\text{-}10}}(\x).
\end{equation}
Both quantities are pre-computed and available via exact lookup in the benchmark table, so no training is performed at evaluation time.

The two signals are positively correlated as a strong architecture on CIFAR-10 tends to perform well on ImageNet16-120, but the proxy optimum and the true optimum are different. 
CIFAR-10 is a low-complexity task on which even lightweight convolutions saturate performance; ImageNet16-120 penalizes architectures that lack sufficient capacity or spatial resolution handling, so the rankings differ non-trivially between the two datasets, including at the top~\cite{yu2019evaluating}. 

We use NAS-Bench-201~\cite{dong2020bench}, available at 
\url{https://github.com/D-X-Y/NAS-Bench-201}, under the MIT License.

\subsection{Task Embedding via Feedback Error Profiles}\label{sec:umap}
To explain the relationship between ground-truth labels and feedback signals across different task configurations, we represent each task instance as a vector that summarizes the distribution of feedback errors. %

\paragraph{Task vector construction.}

Given a batch of query points with ground-truth labels
$\mathbf{y} \in \mathbb{R}^N$ and corresponding feedback signals $\mathbf{u} \in \mathbb{R}^N$ %
, we define the pointwise error as
\begin{equation}
    d_i = u_i - y_i, \qquad i = 1, \dots, N.
\end{equation}
To capture how errors vary across the output space, we sort the points by either $y_i$ or $u_i$ and partition them into $B=20$ equal-sized bins. 
For each bin $b$, we calculate both the signed mean error (capturing directional bias) and the mean absolute error (capturing the magnitude of the errors regardless of direction). This results in four statistics per bin:
\begin{align}
    \mu^{(y)}_b       &= \frac{1}{|b|}\sum_{i \in b} d_i,       &
    \mu^{(u)}_b       &= \frac{1}{|b|}\sum_{i \in b} d_i,       \\
    \bar{\mu}^{(y)}_b &= \frac{1}{|b|}\sum_{i \in b} |d_i|,     &
    \bar{\mu}^{(u)}_b &= \frac{1}{|b|}\sum_{i \in b} |d_i|,
\end{align}
where the superscripts $(y)$ and $(u)$ denote binning by $y$ and $u$,
respectively. 
We use the two statistics generated by binning by feedback ($\bar{\mu}^{(u)}_b, \mu^{(u)}_b$) and concatenate them to generate an embedding of each task that incorporates how feedback corresponds to the value of the objective.

\subsubsection*{Dimensionality Reduction and Visualisation}
All task vectors are projected to two dimensions using UMAP \citep{mcinnes2018umap} with $n_{\text{neighbors}}$ (set to 5) and $\min_{\text{dist}}$ (set to 0.1) as hyperparameters. 
Each point in the resulting embedding corresponds to a
single task instance, colored and shaped by its label (Figure~\ref{fig:umap}).

\section{Pretraining and test-time algorithms}\label{app:algs}
\subsection{Training procedure}
\label{app:training}

\paragraph{MDP formulation.}
We cast policy learning as a finite-horizon MDP $(\mathcal{S}, \mathcal{A}, \mathcal{P}, \mathcal{R}, \gamma)$. At step $t$, the state is
\begin{equation}
    s_t^\tau = (\widetilde{H}_t^\tau, U^\tau, Q_t^\tau),
\end{equation}
the action space is the current candidate pool $\mathcal{A}_t^\tau = Q_t^\tau$, and the policy selects $\x_{t}^\tau \sim \pi_\psi(\cdot \mid \widetilde{H}_t^\tau, U^\tau)$. 
The transition appends the new observation to the history and removes the selected candidate from the pool.
\begin{equation}
    \widetilde{H}_{t+1}^\tau = \widetilde{H}_t^\tau \cup \{(\x_{t}^\tau, u^\tau(\x_{t}^\tau), y_{t}^\tau)\},
    \qquad
    Q_{t+1}^\tau = Q_t^\tau \setminus \{\x_{t}^\tau\},
\end{equation}
where $u^\tau(\x)$ is the feedback of $\x$ in $U^\tau$. 
The reward is the improvement of the $t$-th query over the best value
observed strictly before round $t$,
\begin{equation}
    r_t^\tau = \max\bigl(y_t^\tau - y^*_{t},\, 0\bigr),
    \qquad
    y^*_t = \max\bigl(\max_{(\x,y)\in H_0^\tau} y,\ \max_{i<t} y_i^\tau\bigr),
\end{equation}
and
the discounted return from step $t$ is
\begin{equation}
    G_t^\tau = \sum_{t'=t}^{T_\tau-1} \gamma^{t'-t}\, r_{t'}^\tau .
\end{equation}

We use $\gamma = 0.98$ in all experiments. The policy is optimized with REINFORCE,
\begin{equation}
    \mathcal{L}_{\mathrm{pol}}(\psi)
    =
    -\mathbb{E}_{\tau,\pi_\psi}
    \left[
        \frac{1}{T_\tau}\sum_{t=0}^{T_\tau-1}
        \log \pi_\psi(\x_{t}^\tau \mid \widetilde{H}_t^\tau, U^\tau)
        \tilde{R}_t^\tau
    \right],
\end{equation}
where $\tilde{R}_t^\tau$ is the normalized return defined below.

\paragraph{Return normalization.}
To reduce variance in the REINFORCE gradient estimates, returns are normalized per timestep across the batch,
\begin{equation}
    \tilde{R}_t^\tau = \frac{G_t^\tau - \bar G_t}{s_{G,t} + \epsilon},
\end{equation}
where $\bar G_t$ and $s_{G,t}$ are the mean and standard deviation of
$\{G_t^\tau\}$ across the batch at step $t$, and $\epsilon = 10^{-8}$.
This centers and scales the gradient signal at each step regardless of the absolute magnitude of rewards, which can vary substantially across tasks and feedback conditions.

\paragraph{Warmup schedule.}
Training proceeds in two phases. During the warmup phase, lasting $\eta$ iterations, only the predictive loss $\mathcal{L}_{\mathrm{pred}}$ is active. 
Candidates are sampled from the policy head, which gets no gradient during warm-up, in a smaller query pool of size $\max T_\tau = 20$.
After $\eta$ iterations, the pool reverts to its standard size $|Q^\tau|$, the policy head is activated, and joint optimization of $(\phi, \psi)$ under $\mathcal{L} = \mathcal{L}_{\mathrm{pred}} + \lambda_{\mathrm{pol}} \mathcal{L}_{\mathrm{pol}}$ begins, with a new optimizer in which all parameters outside the policy head use one fifth of the learning rate.

\paragraph{Deployment modes.}
At test time, the model supports two deployment modes.

\emph{End-to-end amortized.} The policy head $\pi_\psi$ selects the next candidate directly,
\begin{equation}
    \x_{t}^\tau \sim \pi_\psi(\cdot \mid \widetilde{H}_t^\tau, U^\tau).
\end{equation}
During training we sample from the softmax distribution, while during evaluation the argmax is taken.

\emph{Semi-amortized.} The inference head $q_\phi$ is used as a feedback-aware surrogate and paired with a standard acquisition function $\alpha$,
\begin{equation}
    \x_{t}^\tau = \arg\max_{\x \in Q_t^\tau} \alpha\!\left(\x;\, q_\phi, \widetilde{H}_t^\tau, U^\tau\right).
\end{equation}
In our experiments, we use $q_\phi$ with expected improvement (EI) and upper confidence bound (UCB) as acquisition functions, combined with the $\pi$BO feedback reweighting (FICBO-UCB and FICBO-EI, \cref{app:baseline-details}).
Both modes share the same pretrained backbone and differ only in which head is used at deployment.

\paragraph{Implementation details.}
Log-probabilities in the REINFORCE objective are clamped to a minimum of $-10$ to prevent numerical instability from near-zero policy probabilities. Gradients are clipped to unit $\ell_2$ norm at each update step.
$\lambda_{\mathrm{pol}} = 10$ scales the normalized return in the policy loss. 

\paragraph{Algorithm.}
The full pretraining procedure is given in Algorithm~\ref{alg:feedback-pretraining}.

\begin{algorithm}[H]
\caption{Feedback-aware pretraining}
\label{alg:feedback-pretraining}
\begin{algorithmic}[1]
\Require Task prior $p(\tau)$, warmup iterations $\eta$, total iterations $N$, loss weight $\lambda_{\mathrm{pol}}$
\For{$i = 1, \dots, N$}
    \State Sample a batch of 200 tasks $\tau \sim p(\tau)$, with one horizon $T_\tau \sim \mathcal{U}\{10,\dots,20\}$ shared by the batch, each with objective $f_\tau$, feedback source $u^\tau$, query set $Q^\tau$, target set $\mathcal{T}^\tau$, horizon $T_\tau$
    \State Initialize $\widetilde{H}_0^\tau$ from initial context points, compute $U^\tau = \{u^\tau(\x) : \x \in Q^\tau\}$
    \For{$t = 0, \dots, T_\tau - 1$}
        \If{$i \leq \eta$}
        \State $\x_{t}^\tau \sim \pi_\psi(\cdot \mid \widetilde{H}_t^\tau, U^\tau)$ \Comment{warmup: policy head not trained}        
        \Else
            \State $\x_{t}^\tau \sim \pi_\psi(\cdot \mid \widetilde{H}_t^\tau, U^\tau)$ \Comment{policy selection}
        \EndIf
        \State Observe $y_{t}^\tau = f_\tau(\x_{t}^\tau) + \varepsilon_{t}^\tau$
        \State $\widetilde{H}_{t+1}^\tau \gets \widetilde{H}_t^\tau \cup \{(\x_{t}^\tau, u^\tau(\x_{t}^\tau), y_{t}^\tau)\}$
        \State $Q_{t+1}^\tau \gets Q_t^\tau \setminus \{\x_{t}^\tau\}$
        \If{$i > \eta$}
            \State Record $r_{t}^\tau \gets \max(y_{t}^\tau - y^*_t, 0)$
        \EndIf
    \EndFor
    \State Compute $\mathcal{L}_{\mathrm{pred}}(\phi)$ from $\{\widetilde{H}_t^\tau, U^\tau, \mathcal{T}^\tau\}_{t=0}^{T_\tau - 1}$ \Comment{Eq.~\eqref{eq:Lpred}}
    \If{$i \leq \eta$}
        \State Update $\phi$ by minimizing $\mathcal{L}_{\mathrm{pred}}$
    \Else
        \State Compute $\mathcal{L}_{\mathrm{pol}}(\psi)$ from $\{r_t^\tau, \log\pi_\psi(\x_{t}^\tau \mid \widetilde{H}_t^\tau, U^\tau)\}_{t=0}^{T_\tau-1}$ \Comment{Eq.~\eqref{eq:lpol}}
        \State Update $(\phi, \psi)$ by minimizing $\mathcal{L}_{\mathrm{pred}} + \lambda_{\mathrm{pol}}\mathcal{L}_{\mathrm{pol}}$ %
    \EndIf
\EndFor
\end{algorithmic}
\end{algorithm}

\section{Detailed Results}\label{app:detailed_results}
\begin{figure}[p]
    \centering
    \includegraphics[width=0.75\textwidth]{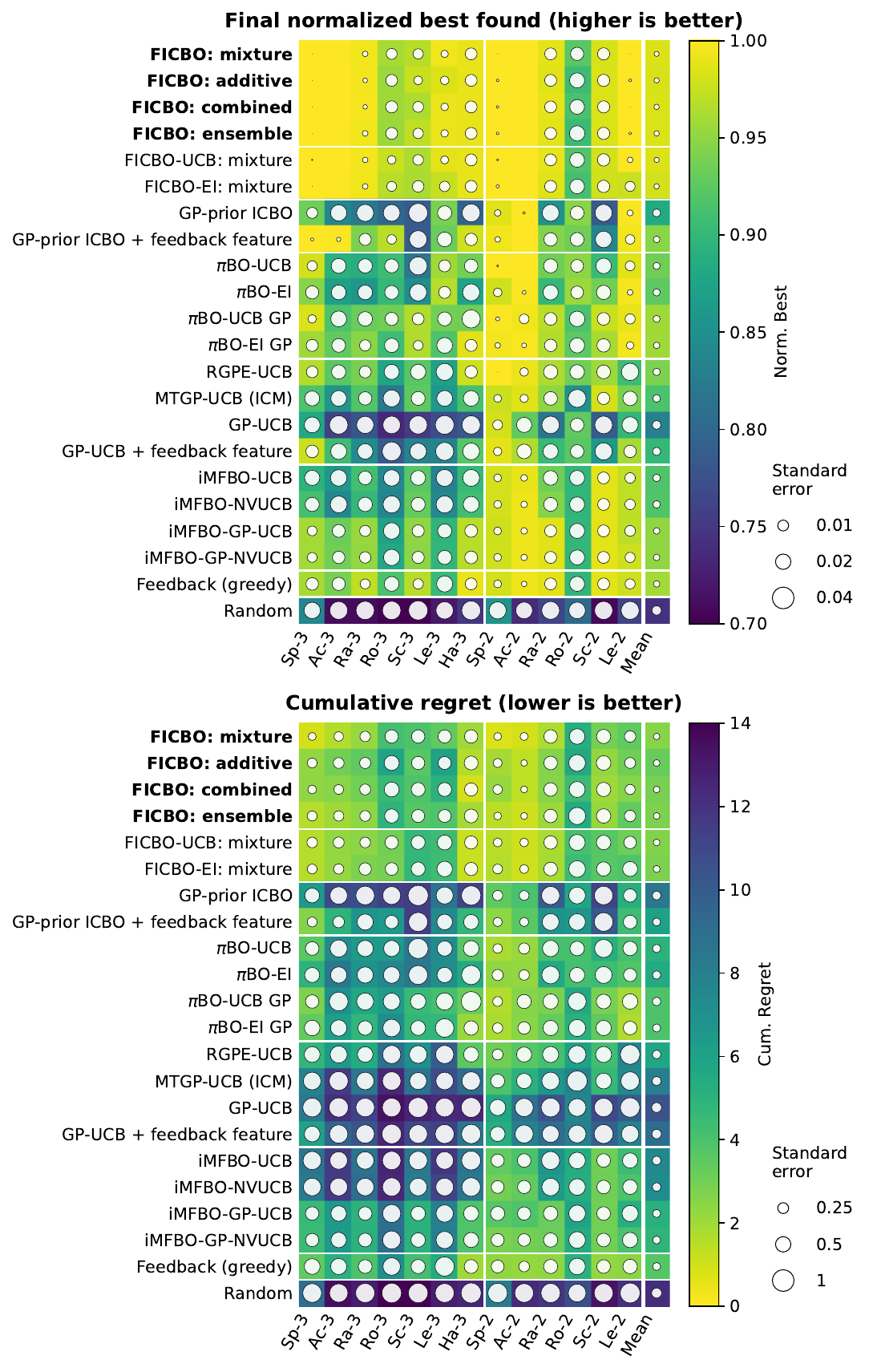}
    \caption{
        \textbf{Per-function results, marginal expert.}
        \textbf{Top:} final normalized best value found ($\uparrow$).
        \textbf{Bottom:} normalized cumulative regret ($\downarrow$).
        Columns: Sp = Sphere, Ac = Ackley, Ra = Rastrigin, Ro = Rosenbrock, Sc = Schwefel, Le = Levy, Ha = Hartmann; the suffix describes which dimension the benchmark has. Mean averages over the 13 functions.
        Color shows the mean over 100 replicas; circle size shows the standard error.
        Bold rows are FICBO with the end-to-end policy. Methods are described in \cref{app:baseline-details}.
    }
    \label{app:fig:heatmap_marginal}
\end{figure}

\begin{figure}[p]
    \centering
    \includegraphics[width=0.75\textwidth]{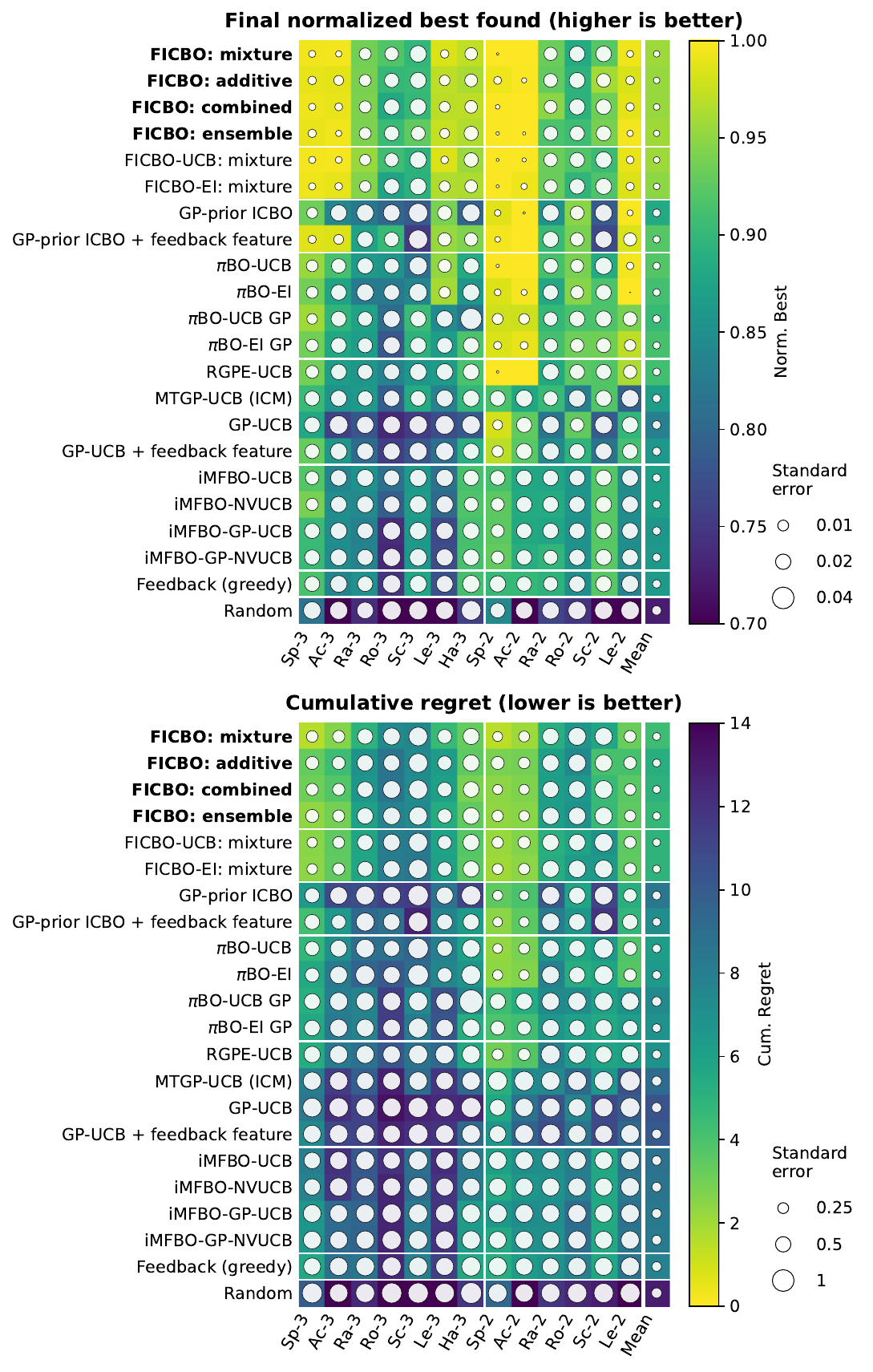}
    \caption{
        \textbf{Per-function results, XGBoost expert} (\cref{sec:benchmarkfunctions}).
        Layout and abbreviations as in \cref{app:fig:heatmap_marginal}.
    }
    \label{app:fig:heatmap_xgb}
\end{figure}

\begin{figure}[p]
    \centering
    \includegraphics[width=0.75\textwidth]{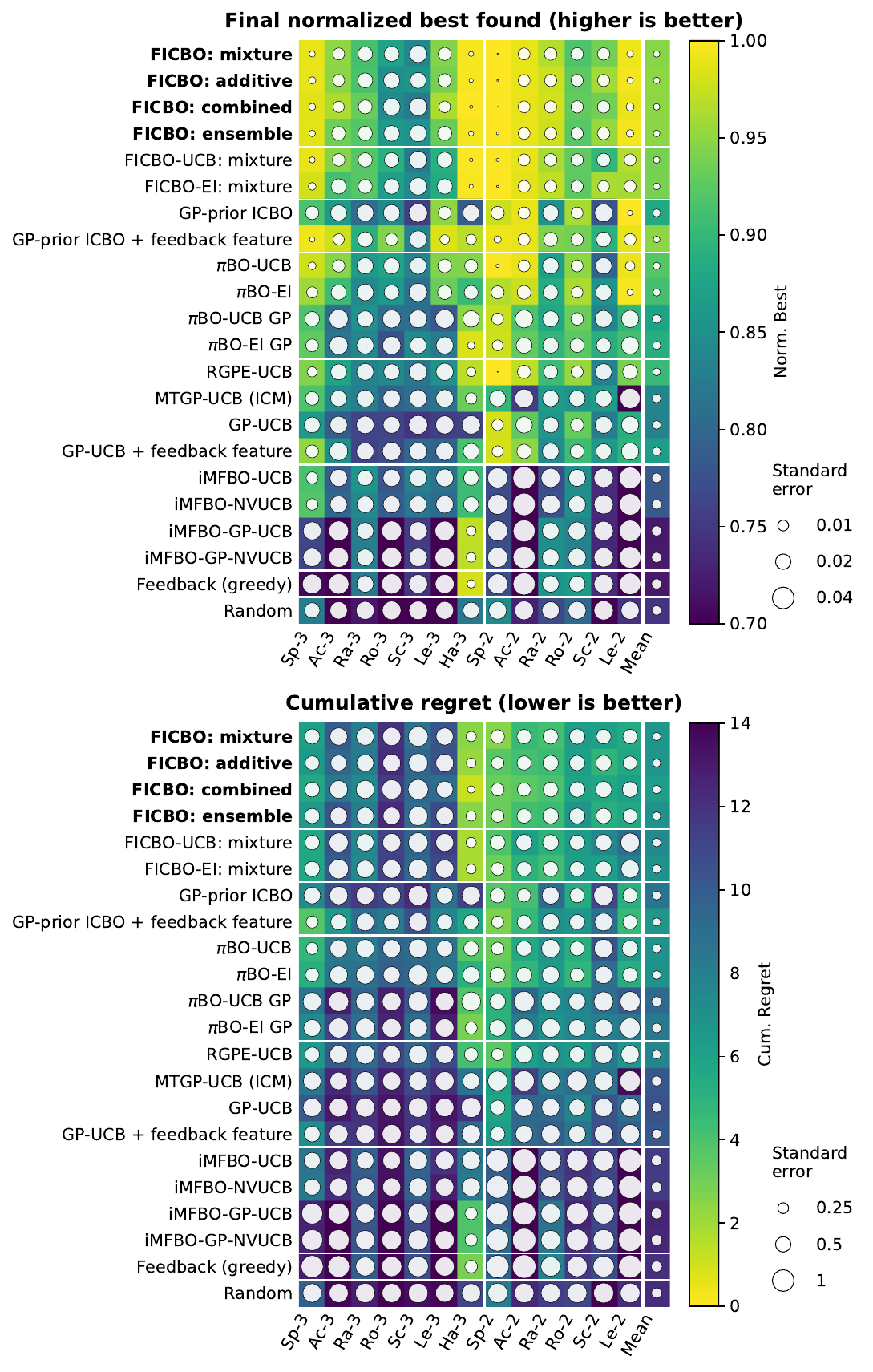}
    \caption{
        \textbf{Per-function results, marginal expert with fake peak} (\cref{app:fig:fake_peak}).
        Layout and abbreviations as in \cref{app:fig:heatmap_marginal}.
    }
    \label{app:fig:heatmap_fakepeak}
\end{figure}

The heatmaps in \cref{app:fig:heatmap_marginal,app:fig:heatmap_xgb,app:fig:heatmap_fakepeak} summarize the final normalized best value and the cumulative regret of all methods for each benchmark function.
The results in the appendix were obtained by recomputing the results, which can lead to slightly different results due to randomness.

\section{Continuous BO} \label{app:Continuous}
PFNs4BO: In-Context Learning for Bayesian Optimization \citep{muller2023pfns4bo} samples several sets of points and filters good candidates. This workflow alone does not work well with our methods since also areas with low chance to result in desirable solutions are valuable to have in the context due to the feedback.
As a consequence randomly sampled points are added to the query set, which will not be optimized and serve as background information. 
Then the x coordinates are iteratively refined via gradient updates, either starting from the acquisition head or classic acquisition functions. 
We decided to simplify the setup from PFNs4BO since our problems are low-dimensional, and sampling and discarding thousands of points is a waste in this setting. The resulting procedure is as follows: 

Each BO step:
\begin{enumerate}
    \item Draws a fresh Sobol set split into an editable candidate block (100 points) and a fixed background block (200 points), matching the 300 points the pool-based variant uses
    \item Scores the whole block in one forward pass
    \item Runs multistart Adam ascent on the acquisition to optimize the editable coordinates only with ten steps. The background points are scored alongside but get no gradient and are excluded from the ascent objective: the query block self-attends and the head softmaxes over it, so a block of only refined, high-acquisition points would be far off the IID query distribution the model was trained on; they still stay eligible to win.
    Expert feedback is re-observed from the simulator without gradients, as we cannot assume the feedback is differentiable, after every step and fed back into the score (never differentiated through), and the argmax over [refined, background] is evaluated for real and appended to the context.

\end{enumerate}

\begin{table}[t]
    \centering
    \caption{Final normalized best-so-far value and cumulative regret, averaged over 13 benchmarks with the marginal expert for continuous optimization. Best-so-far: higher is better. Cumulative regret: lower is better. Here the regret values are calculated using the global optimum, instead of the pool optimum.} %
    \label{tab:results_Cont}
    \begin{tabular}{lcc}
    \toprule
    Method & Best-so-far ($\uparrow$) & Cumulative Regret ($\downarrow$) \\
    \midrule
    FICBO (Mixture Prior) + UCB (continuous) & $\mathbf{0.8106} \pm 0.0441$ & $\mathbf{8.7665} \pm 1.6063$ \\
    FICBO (Mixture Prior) (continuous)       & $0.8036 \pm 0.0457$          & $8.8110 \pm 1.7620$ \\
    FICBO (Mixture Prior) + EI (continuous)  & $0.8014 \pm 0.0496$          & $8.7736 \pm 1.8338$ \\
    FICBO (Mixture Prior) (pool)             & $0.7775 \pm 0.0550$          & $9.0800 \pm 1.9253$ \\
    FICBO (Mixture Prior) + UCB (pool)       & $0.7775 \pm 0.0536$          & $9.2440 \pm 1.9226$ \\
    FICBO (Mixture Prior) + EI (pool)        & $0.7745 \pm 0.0545$          & $9.1886 \pm 1.9102$ \\
    GP Surrogate (pool, botorch)             & $0.7499 \pm 0.0585$          & $10.7315 \pm 2.0342$ \\
    GP Surrogate (pool)                      & $0.7458 \pm 0.0582$          & $10.6288 \pm 1.9952$ \\
    \bottomrule
    \end{tabular}
\end{table}

Table~\ref{tab:results_Cont} shows that pool-based optimization variants have been beaten by their continuous counterparts, but this comes at the cost of drastically increasing the computational cost and number of feedback oracle calls (for every optimization step, all optimized points need feedback at that point). In our implementations, the computation time jumps from 0.5 sec to about 20 seconds for the 100 rollouts of one benchmark function.

The different acquisition functions performed very similarly. The main difference comes from the surrogates.

It is worth noting that despite the improved performance for this experiment, the policy has not been trained with continuous optimization in mind, and the optimization steps may cause the distribution of x coordinates to be OOD for what the model has been trained on, as the candidates will often cluster. 
Proper continuous optimization support is left as future work.

\section{Detailed model architecture}\label{app:architecture}

\subsection{Computational Resources}\label{app:hardware}

All experiments were conducted on a high-performance computing cluster using a single 
NVIDIA V100 GPU with 32GB of memory. Training took approximately $70$ hours per run. 
All models were implemented in PyTorch~\cite{paszke2019pytorch}.

\subsection{Model Architecture}

The model follows an encode-then-decode architecture consisting of three components: 
an embedder, a Transformer encoder, and an output head. We base our implementation on \citep{huang2025aline}.
We do not claim that this is the optimal architecture to use, but it is good enough to demonstrate the benefit of training on our FICBO priors.

\paragraph{Embedder.} We provide two embedder variants. The standard embedder maps 
input locations $x$ through a two-layer MLP with ReLU activations to produce 
$d_{\mathrm{emb}}$-dimensional tokens. Observed values $y$ are embedded separately and added to 
the context tokens, so query and target tokens carry only location information.

The expert-aware embedder additionally incorporates the expert feedback signal. 
Input locations and feedback are each embedded by separate two-layer MLPs. 
In \textit{concat} mode (used in the experiments), the $x$ and feedback embeddings (each of dimension $d_{\mathrm{emb}}/2$) 
are concatenated to form a $d_{\mathrm{emb}}$-dimensional token; in \textit{add} mode they are 
summed directly. As before, observed values $y$ are embedded and added to context 
tokens only. In both variants, query and target tokens therefore carry no $y$ 
information, only location, and, in the expert-aware case, feedback.

\paragraph{Encoder.} The embeddings are processed by a Transformer encoder with $N_{\mathrm{layers}}$ layers, $N_{\mathrm{heads}}$ attention heads, embedding dimension $d_{\mathrm{emb}}$, and feed-forward dimension $d_{\mathrm{ff}}$. All tokens attend freely to context tokens. 
Optionally, all tokens can also attend to query tokens to find patterns in the feedback (turned on in experiments). 

\paragraph{Output head.} The output head splits the encoder output into query and target 
embeddings. Query embeddings are passed through the \textit{policy head}, a two-layer 
MLP that produces a scalar logit per query point; these logits are converted to a 
probability distribution over candidate points via softmax, forming the policy 
$\pi_\psi(\x \mid \widetilde H_t^\tau, U^\tau)$. Target and query embeddings are passed through the \textit{inference head}, 
which predicts a Gaussian mixture posterior with $L$ components over the target 
outputs, parameterizing means, standard deviations, and mixture weights via separate 
linear projections. An optional time token $t/T$ can be concatenated to the query embeddings before the policy head; it had negligible influence and is not used in the reported models.

\subsection{Hyperparameters}\label{app:hp}
These hyperparameters were used to train the models used in the experimental section. 
They do not represent the optimal hyperparameters for every project, but work reasonably well over a diverse set of feedback strengths and biases.
\small
\setlength{\tabcolsep}{6pt}
\renewcommand{\arraystretch}{1.12}

\begin{longtable}{>{\raggedright\arraybackslash}p{6.6cm} >{\raggedright\arraybackslash}p{7.6cm}}
\caption{Hyperparameter settings. Symbols follow the notation of
Sec.~\ref{sec:method} and \cref{app:synthgen,app:training}.}
\label{tab:hp-sketch} \\
\toprule
\textbf{Hyperparameter} & \textbf{Value / Range} \\
\midrule
\endfirsthead
\multicolumn{2}{c}{\tablename\ \thetable{} -- continued} \\
\toprule
\textbf{Hyperparameter} & \textbf{Value / Range} \\
\midrule
\endhead
\bottomrule
\endlastfoot

\multicolumn{2}{c}{\textbf{Synthetic task prior}} \\
\midrule
Candidate pool size $N_\tau$ & 200 \\
Horizon $T_\tau$ & 10--20 \\
Optimizer input dimension $d$ & varies by experiment (1--3) \\
Number of held-out targets $|\mathcal{T}^\tau|$ & $100$ \\
Kernel family & Matérn with $\nu \in \mathcal{U}(\{3/2, 5/2, \infty\})$ \\
Lengthscale prior & $\mathcal{U}([0.1\sqrt{d},\; 2.0\sqrt{d}])$, isotropic w.p.\ 0.5 \\
Output scale prior & $\mathcal{U}([0.1,\; 1.0])$ \\
Observation noise $\sigma_\tau$ & $0.01$ (evaluation: $0.1$) \\
Design space & $[-5, 5]^{d_{\mathrm{full}}^\tau}$ \\
\midrule

\multicolumn{2}{c}{\textbf{Additive prior}} \\
\midrule
Source input dimension $d_{\mathrm{src}}^\tau$ & $\mathcal{U}(\{1, 2\})$ \\
Overlap dimensions $o^\tau$ & $\mathcal{U}(\{0, 1\})$\,$^{c}$ \\
Source weight $w_\tau$ & $\mathcal{U}([0.25,\; 1.0])$ \\
Source-private observations $n_{\mathrm{src}}$ & $\mathcal{U}(\{10, \ldots, 100\})$ for $d=1$; $\mathcal{U}(\{30, \ldots, 300\})$ for $d\in\{2,3\}$ \\ %
Fraction of $\mathcal{D}_{\mathrm{src}}^\tau$ sampled uniformly & $0.25$ (remainder sampled in clusters) \\
Source cluster std & $\mathcal{U}([0.5,\; 1.5])$ \\
Number of source cluster centers & $\mathcal{U}(\{1, \ldots, 5\})$ \\
\midrule

\multicolumn{2}{c}{\textbf{Mixture prior}} \\
\midrule
Number of real components $K_{\mathrm{real}}$ & $3$ \\
Number of decoy components $K_{\mathrm{decoy}}^\tau$ & $\mathcal{U}(\{0, \ldots, 5\})$ per batch (Combined model: $\{2,\ldots,8\}$) \\
Dirichlet concentration $\alpha_{\mathrm{src}}$ & $0.7$ \\
Component inclusion probability $p_{\mathrm{src}}$ & $0.8$ \\
\midrule

\multicolumn{2}{c}{\textbf{Bias operators} $\mathcal{B}_\tau$$^{a}$} \\
\midrule
Activation probabilities (\cref{appendix:desr_of_biases}) & additive GP bias: 0.15; local distortion: 0.15; constant shift: 0.15; catastrophic failure: 0.05; knowledge-center noise: 1.0 (Mixture), off (Additive) \\
Noise: scale $\sigma_{\mathrm{fb}}$ & $\mathcal{U}([0.0,\; 0.2])$ \\
Noise: knowledge-center scale & $\mathcal{U}([0.0,\; 1.0])$ (Mixture only) \\
Additive GP bias: lengthscale & $\mathcal{U}([0.5,\; 3.0])$ \\
Additive GP bias: scale & $\mathcal{U}([0.2,\; 2.0])$ \\
Additive GP bias: max std & $\mathcal{U}([0.2,\; 1.0])$ \\
Local distortion: number of centers $n_{\mathrm{centers}}$ & $\mathcal{U}(\{1, 2\})$ \\
Local distortion: magnitude & $\mathcal{U}([0.2,\; 1.0])$, random sign \\
Local distortion: decay & $\mathcal{U}([0.5,\; 1.5])$ \\
Constant shift: $\delta$ & $\mathcal{U}([-0.5,\; 0.5])$ \\
\midrule

\multicolumn{2}{c}{\textbf{Transformer backbone}} \\
\midrule
Embedding dimension $d_{\mathrm{emb}}$ & 32 \\
Number of layers $N_{\mathrm{layers}}$ & 3 \\
Number of attention heads $N_{\mathrm{heads}}$ & 4 \\
Feed-forward dimension $d_{\mathrm{ff}}$ & 128 \\
Dropout & 0.0 \\
\midrule

\multicolumn{2}{c}{\textbf{Heads}} \\
\midrule
Number of mixture components $L$ & 10 \\
Hyperparameter targets in $\mathcal{L}_{\mathrm{pred}}$ & d+1 ($d$ lengthscales + output scale) \\
Policy head & two-layer MLP, applied to each query token \\
\midrule

\multicolumn{2}{c}{\textbf{Training}} \\
\midrule
Optimizer & AdamW \\
Learning rate & $10^{-3}$ (after warm-up: $2\cdot 10^{-4}$ for all parameters except the policy head) \\
Weight decay & $10^{-7}$ (after warm-up: $10^{-2}$, AdamW default) \\
LR schedule & cosine annealing ($T_{\max}=200{,}000$ during warm-up, reset to $180{,}000$ after warm-up) \\
Batch size (parallel tasks) & 200 \\
Total iterations $N$ & $200{,}000$ \\
Warm-up iterations $\eta$ & $20{,}000^{b}$ \\
Discount $\gamma$ & 0.98 \\
Policy loss weight $\lambda_{\mathrm{pol}}$ & $10$ \\
Return-normalization constant $\epsilon$ & $10^{-8}$ \\
Log-probability clamp & $-10$ \\
Gradient clipping & unit $\ell_2$ norm \\
\midrule

\multicolumn{2}{c}{\textbf{Evaluation}} \\
\midrule
Initial observations $|H_0^\tau|$ & 1 \\
Candidate pool size $N_\tau$ & 300 \\
Horizon $T_\tau$ & 30 \\
Number of seeds & 100 \\
\end{longtable}
\noindent\footnotesize$^{a}$ Due to computational constraints, the hyperparameters were not optimized and were set arbitrarily. \\
$^{b}$ Following ALINE~\cite{huang2025aline}, the warm-up uses only the prediction loss and a smaller query size.\\
$^{c}$ With this range, $o^\tau = d$ never occurs for the $d\in\{2,3\}$ models reported in the main paper. It does occur for the $d=1$ trained variant (see Appendix~\ref{app:synthgen:additive}), where the optimizer then has no private features, and we default to $w_\tau=1$, since there is no optimizer-only component to weigh against.

\clearpage
\newpage

\clearpage
\newpage
\section*{NeurIPS Paper Checklist}

\begin{enumerate}

\item {\bf Claims}
    \item[] Question: Do the main claims made in the abstract and introduction accurately reflect the paper's contributions and scope?
    \item[] Answer: \answerYes{} %
    \item[] Justification: 
    The claims are supported by experiments in later sections.

\item {\bf Limitations}
    \item[] Question: Does the paper discuss the limitations of the work performed by the authors?
    \item[] Answer: \answerYes{} %
    \item[] Justification: The discussion section has a paragraph about limitations addressing some core limitations.%

\item {\bf Theory assumptions and proofs}
    \item[] Question: For each theoretical result, does the paper provide the full set of assumptions and a complete (and correct) proof?
    \item[] Answer:  \answerNA{}.
    \item[] Justification: No theoretical results to prove.%

    \item {\bf Experimental result reproducibility}
    \item[] Question: Does the paper fully disclose all the information needed to reproduce the main experimental results of the paper to the extent that it affects the main claims and/or conclusions of the paper (regardless of whether the code and data are provided or not)?
    \item[] Answer: \answerYes{} %
    \item[] Justification: The appendix contains extensive descriptions of the architecture, hyperparameters (\cref{app:architecture}), benchmarks (\cref{app:details_vis}) and the prior variants (\cref{app:synthgen}).%

\item {\bf Open access to data and code}
    \item[] Question: Does the paper provide open access to the data and code, with sufficient instructions to faithfully reproduce the main experimental results, as described in supplemental material?
    \item[] Answer: \answerYes{} %
    \item[] Justification: Code is publicly available on GitHub. The repository is linked in the paper. %

\item {\bf Experimental setting/details}
    \item[] Question: Does the paper specify all the training and test details (e.g., data splits, hyperparameters, how they were chosen, type of optimizer) necessary to understand the results?
    \item[] Answer: \answerYes{} %
    \item[] Justification: The appendix contains extensive descriptions of the architecture, hyperparameters (\cref{app:architecture}), benchmarks (\cref{app:details_vis}) and the prior variants (\cref{app:synthgen}). %

\item {\bf Experiment statistical significance}
    \item[] Question: Does the paper report error bars suitably and correctly defined or other appropriate information about the statistical significance of the experiments?
    \item[] Answer: \answerYes{} %
    \item[] Justification: Our plots display error bars with explanation, usually the 95\% confidence interval is visualized. %

\item {\bf Experiments compute resources}
    \item[] Question: For each experiment, does the paper provide sufficient information on the computer resources (type of compute workers, memory, time of execution) needed to reproduce the experiments?
    \item[] Answer:\answerYes{} %
    \item[] Justification: we refer to \cref{app:hardware}. %
    
\item {\bf Code of ethics}
    \item[] Question: Does the research conducted in the paper conform, in every respect, with the NeurIPS Code of Ethics \url{https://neurips.cc/public/EthicsGuidelines}?
    \item[] Answer: \answerYes{} %
    \item[] Justification: the work is methodological. It does not involve human subjects, private data, scraped data, or high-risk deployment.

\item {\bf Broader impacts}
    \item[] Question: Does the paper discuss both potential positive societal impacts and negative societal impacts of the work performed?
    \item[] Answer: \answerNA{} %
    \item[] Justification:  Not relevant. %
    
\item {\bf Safeguards}
    \item[] Question: Does the paper describe safeguards that have been put in place for responsible release of data or models that have a high risk for misuse (e.g., pre-trained language models, image generators, or scraped datasets)?
    \item[] Answer: \answerNA{} %
    \item[] Justification:  Not relevant. %

\item {\bf Licenses for existing assets}
    \item[] Question: Are the creators or original owners of assets (e.g., code, data, models), used in the paper, properly credited and are the license and terms of use explicitly mentioned and properly respected?
    \item[] Answer: \answerYes{} %
    \item[] Justification: We use NeuralFoil~\cite{sharpe2025neuralfoil} and NAS-Bench-201~\cite{dong2020bench}. License information 
for all assets is provided in the appendix.

\item {\bf New assets}
    \item[] Question: Are new assets introduced in the paper well documented and is the documentation provided alongside the assets?
    \item[] Answer:  \answerNA{} %
    \item[] Justification: The publication does not introduce new assets. %

\item {\bf Crowdsourcing and research with human subjects}
    \item[] Question: For crowdsourcing experiments and research with human subjects, does the paper include the full text of instructions given to participants and screenshots, if applicable, as well as details about compensation (if any)? 
    \item[] Answer: \answerNA{} %
    \item[] Justification:  Not relevant. %

\item {\bf Institutional review board (IRB) approvals or equivalent for research with human subjects}
    \item[] Question: Does the paper describe potential risks incurred by study participants, whether such risks were disclosed to the subjects, and whether Institutional Review Board (IRB) approvals (or an equivalent approval/review based on the requirements of your country or institution) were obtained?
    \item[] Answer: \answerNA{} %
    \item[] Justification: Not relevant. %

\item {\bf Declaration of LLM usage}
    \item[] Question: Does the paper describe the usage of LLMs if it is an important, original, or non-standard component of the core methods in this research? Note that if the LLM is used only for writing, editing, or formatting purposes and does \emph{not} impact the core methodology, scientific rigor, or originality of the research, declaration is not required.
    \item[] Answer: \answerNA{}.
    \item[] Justification: No important, original, or non-standard LLM use.%

\end{enumerate}

\end{document}